\documentclass[10pt,journal,compsoc]{IEEEtran}
\usepackage{xcolor}
\usepackage{pifont}
\usepackage{wrapfig}
\usepackage{diagbox}
\usepackage{bm}
\usepackage{adjustbox}
\usepackage{booktabs}

\newcommand{\cmark}{\ding{51}}
\newcommand{\xmark}{\ding{55}}

\usepackage[utf8]{inputenc} 
\usepackage[T1]{fontenc}    
\usepackage{url}            
\usepackage{booktabs}       
\usepackage{amsfonts}       
\usepackage{nicefrac}       
\usepackage{microtype}      

\usepackage{color,soul}
\usepackage{pgfplots}
\usepackage{tikz}
\usepgfplotslibrary{groupplots}
\usepackage{arydshln}
\usepackage{float}
\usepackage{wrapfig,booktabs}
\usepackage{lipsum}
\usepackage{graphicx}
\usepackage{times}
\usepackage{epsfig}
\usepackage{graphicx}
\usepackage{amsmath}
\usepackage{amssymb}
\usepackage{vwcol}
\usepackage{booktabs,multirow}  
\usepackage{amsfonts}       
\usepackage{nicefrac}      
\usepackage{microtype}     

\usepackage[colorlinks=true,
            linkcolor=blue,
            citecolor=blue,
            filecolor=blue,
            urlcolor=blue]{hyperref}
\hypersetup{colorlinks = true,    linkcolor=blue,citecolor=blue,filecolor=blue,urlcolor=black}

\usepackage{amsmath}
\usepackage{amssymb}
\usepackage{mathtools}
\usepackage{amsthm}
\usepackage{multicol}
\usepackage{enumitem}

\usepackage{algorithm}
\usepackage{algorithmic}
\usepackage[misc]{ifsym}
\renewcommand{\algorithmiccomment}[1]{ 
\textcolor{black!50!white}{$\triangleright$} {\footnotesize \textcolor{black!50!white}{\texttt{#1}}}}

\theoremstyle{plain}

\theoremstyle{definition}

\theoremstyle{remark}

\usepackage{float}
\usepackage{wrapfig,booktabs}
\usepackage{amssymb}

\usepackage{textcomp,booktabs}
\definecolor{mygray}{gray}{.95}
\definecolor{definegray}{gray}{0.3}
\definecolor{definegreen}{HTML}{3FBC9D}

\newcommand{\revised}[1]{{\color{black}{#1}}}

\usepackage{amsmath,amsfonts,bm}

\def\eqref#1{equation~\ref{#1}}

\def\1{\bm{1}}

\DeclareMathAlphabet{\mathsfit}{\encodingdefault}{\sfdefault}{m}{sl}
\SetMathAlphabet{\mathsfit}{bold}{\encodingdefault}{\sfdefault}{bx}{n}

\ifCLASSOPTIONcompsoc
  \usepackage[nocompress]{cite}
\else
  \usepackage{cite}
\fi

\ifCLASSINFOpdf
\else
\fi

\begin{document}

\title{Continual Learning From a Stream of APIs}

\author{
Enneng Yang, Zhenyi Wang, Li Shen, Nan Yin, Tongliang Liu, \\
Guibing Guo, Xingwei Wang, and Dacheng Tao,~\IEEEmembership{Fellow,~IEEE}
%
\IEEEcompsocitemizethanks{
   \IEEEcompsocthanksitem Enneng Yang, Guibing Guo, and Xingwei Wang are with Northeastern University, China.  E-mail: ennengyang@stumail.neu.edu.cn, \{guogb,wangxw\}@swc.neu.edu.cn
    \IEEEcompsocthanksitem Zhenyi Wang is with University of Maryland, College Park, USA. E-mail:  wangzhenyineu@gmail.com
    \IEEEcompsocthanksitem Li Shen is with Sun Yat-sen University and JD Explore Academy, China. E-mail: mathshenli@gmail.com
    \IEEEcompsocthanksitem Nan Yin is with Mohamed bin Zayed University of Artificial Intelligence, Abu Dhabi, United Arab Emirates, E-mail: yinnan8911@gmail.com
     \IEEEcompsocthanksitem Tongliang Liu is with The University of Sydney, Australia. E-mail: tongliang.liu@sydney.edu.au
    \IEEEcompsocthanksitem Dacheng Tao is with Nanyang Technological University, Singapore. E-mail: dacheng.tao@gmail.com
}

}

\markboth{Journal of \LaTeX\ Class Files,~Vol.~xx, No.~x, September~2023}%
{Shell \MakeLowercase{\textit{et al.}}: Bare Demo of IEEEtran.cls for Computer Society Journals}

\IEEEtitleabstractindextext{%

\begin{abstract}
  Continual learning (CL) aims to learn new tasks without forgetting previous tasks. However, existing CL methods require a large amount of raw data, which is often unavailable due to copyright considerations and privacy risks. Instead, stakeholders usually release pre-trained machine learning models as a service (MLaaS), which users can access via APIs. This paper considers two practical-yet-novel CL settings: data-efficient CL (DECL-APIs) and data-free CL (DFCL-APIs), which achieve CL from a stream of APIs with partial or no raw data. Performing CL under these two new settings faces several challenges: unavailable full raw data, unknown model parameters, heterogeneous models of arbitrary architecture and scale, and catastrophic forgetting of previous APIs. To overcome these issues, we propose a novel data-free cooperative continual distillation learning framework that distills knowledge from a stream of APIs into a CL model by generating pseudo data, just by querying APIs. Specifically, our framework includes two cooperative generators and one CL model, forming their training as an adversarial game. We first use the CL model and the current API as fixed discriminators to train generators via a derivative-free method. Generators adversarially generate hard and diverse synthetic data to maximize the response gap between the CL model and the API. Next, we train the CL model by minimizing the gap between the responses of the CL model and the black-box API on synthetic data, to transfer the API's knowledge to the CL model. Furthermore, we propose a new regularization term based on network similarity to prevent catastrophic forgetting of previous APIs. Our method performs comparably to classic CL with full raw data on the MNIST and SVHN datasets in the DFCL-APIs setting. In the DECL-APIs setting, our method achieves $0.97\times$, $0.75\times$ and $0.69\times$ performance of classic CL on the more challenging CIFAR10, CIFAR100, and MiniImageNet, respectively.
\end{abstract}

\begin{IEEEkeywords}
Data-free Learning, Catastrophic Forgetting, Plasticity-Stability, Continual Learning. 
\end{IEEEkeywords}
}

\maketitle

\section{Indroduction}
\label{sec:intro}

\begin{figure}[t]
\begin{center}
\includegraphics[width=1.\linewidth]{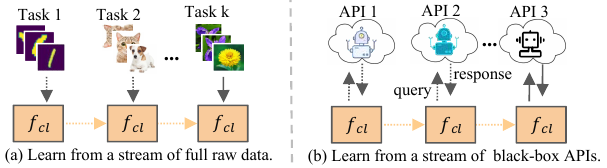}
\end{center}
\vspace{-15pt}
\caption{Illustration of (a) Classic CL and (b) Our proposed CL setting, the former learned from raw data and the latter from APIs.}
\label{fig:scenario}
\vspace{-10pt}
\end{figure}

Deep learning systems have achieved excellent performance in various tasks~\cite{AlphaGo2016,AlphaGoZero2017}. However, all these systems retrain a neural network model when a new task comes along, lacking the ability to accumulate knowledge over time as humans do. Continual learning (CL)~\cite{cl_survey2019, Forgetting_Survey_2023} expects a neural network to continuously learn new tasks without forgetting old tasks on time-evolving data streams. As shown in Fig.~\ref{fig:scenario}(a), classic CL methods~\cite{SynapticIntelligence,hat,er,TaskfreeContinualLearning_DRO_ICML2022} train a CL model on a stream of raw data. However, some valuable rare datasets, such as company sales data~\cite{MEGEX_Arxiv2021}, or hospital medical diagnostic data~\cite{DataFreeKnowledgeTransferASurvey}, cannot be accessed due to privacy concerns~\cite{dafl_iccv2019}. In addition, due to the cost invested in managing training data for machine learning models, pre-trained models become valuable intellectual property~\cite{DataFreeModelExtraction2021CVPR}. To enable these expert models to be monetized, model owners do not publicly release their pre-trained models, but release trained machine learning models as a service (MLaaS)~\cite{roman2009model_maas}, allowing users to access these powerful models through black-box APIs~\cite{sun2022bbt,sun2022bbtv2,knowledgealignment_ccs2019}, such as Amazon Rekognition~\cite{amazon_api}, and Google Explainable AI~\cite{MEGEX_Arxiv2021}. This means that existing traditional CL methods that learn from complete raw training data will not be applicable to data-free scenarios, significantly limiting the real-world application of CL. The natural question is, \emph{can CL be performed when we can only interact with MLaaS (or black-box APIs)}?

In this work, we consider two more practical-yet-novel settings for CL that we call {data-efficient CL} (DECL-APIs) and {data-free CL} (DFCL-APIs), which learn from a stream of APIs (shown in Fig.~\ref{fig:scenario}(b)), using only a small amount of raw data (such as the official API call tutorial data, or a part of the public dataset used for API training~\cite{brown2020language_gpt3}) or no arbitrary raw data, respectively. 
The main challenges of DECL-APIs and DFCL-APIs lie in four aspects:
(i) \textit{unavailable full raw data}. For each API, we have access to only a small amount of raw data~\cite{GAMIN2019,papernot2017practical_jbda}, or no raw data.
(ii) \textit{unknow model parameters}. We do not know the model's architecture and parameters of each API. We only know what task this API performs and what format its input and output are. This is a very modest and accepted assumption.
(iii) \textit{heterogeneous models of arbitrary scale}. Continuously arriving APIs are typically heterogeneous, with different architecture and scales (e.g., width and depth). For example, the architecture used by the first API is ResNet18~\cite{resnet}, the second API uses ResNet50~\cite{resnet}, and the third API uses GoogleNet~\cite{googlenet_CVPR2015}, etc. Certainly, within the context of this paper, we do not need to know what architecture it is.
(iv) \textit{catastrophic forgetting of old APIs}. 
When using a CL model to learn new APIs continuously, the CL model may catastrophically forget previously learned APIs. In other words, after learning a new API (i.e., task), the performance of the CL model on the previous task drops sharply. 
Existing work~\cite{ExModel2022} only addresses the first challenge, which is to perform CL via inverse training data from white-box pre-trained models. It requires knowledge of the pre-trained model's precise architecture and parameters, which is impossible in MLaaS. Additionally, it ignores the more severe problem of catastrophic forgetting that arises from learning on synthetic data. Therefore, how to further fill this gap of performing CL from a stream of black-box APIs is of great significance.

This paper attempts to overcome all these challenges within a unified framework. We propose a novel data-free cooperative continual distillation learning framework that distills knowledge from a stream of black-box APIs into a CL model. As shown in Fig.~\ref{fig:framework}, our framework comprises two collaborative generators, a CL model, and a stream of APIs. When learning a new API, we formulate the generators and the CL model as an adversarial game, where they compete against each other during training to improve each other's performance continuously. In other words, we alternate the following two steps to perform CL from a stream of APIs: training generative models and training the CL model. 
Specifically, (i) \textit{Training generative models}: We first use the black-box API currently being learned as a fixed discriminator and adversarially train two collaborative generators to generate `hard' images that lead to inconsistent responses from the CL model and API. We also maximize the diversity of the images generated by the two generators and constrain the number of images generated by each class to be balanced, so that the generated images fully explore the knowledge of the API. It is worth noting that since the black-box API cannot compute the generators' gradients, we further propose to use a zeroth-order gradient estimation~\cite{conn2009introduction} method to update the generators.
(ii) \textit{Training CL model}: We next train the CL model to achieve knowledge transfer from the API to the CL model by reducing the gap between the responses of the CL model and the API on generated images. We also replay a few generated old images (or a small amount of raw data, only if it is available in DECL-APIs setting) and propose a new regularization term based on network similarity measure~\cite{DistanceCorrelation_eccv2022} to mitigate catastrophic forgetting of the CL model on the old APIs.

Our framework has advantages in the following four aspects. (i) \textit{data-free}: Our framework does not require access to the original training data; instead, it trains two collaborative generators to obtain synthetic data. (ii) \textit{learning from black-box APIs}: Throughout the learning process, we only need to query the black-box API, that is, to obtain the response of the API according to the input, so we do not need to know the model architecture used by the API. (iii) \textit{arbitrary model scale}: Since we do not need to backpropagate the model behind the API to obtain the gradient value of the generators, but use zero-order gradient estimation (which relies on function values) to obtain the estimated gradient value, the model behind the API can be any scale. (iv) \textit{mitigating catastrophic forgetting}: Through the replay of a small amount of old data and the proposed new regularization term, the CL model's catastrophic forgetting of old APIs is greatly alleviated. 
Therefore, our proposed new framework significantly expands the real-world application of CL.

Extensive experiments are conducted to validate the effectiveness of the proposed method in two newly defined CL settings (i.e., DFCL-APIs and DECL-APIs) on three typical scenarios: (i) all APIs use the same architecture, (ii) different APIs use different architectures with different network types, depths, and widths, and (iii) datasets have varying numbers of classes, tasks, and image resolutions. \revised{In addition to the computer vision task, we also validated the proposed method in the natural language processing domain.}
Results from the MNIST and SVHN benchmark datasets show that in the DFCL-APIs setting, the proposed method achieves comparable results to classic CL trained on complete raw data, even without raw data. On the more challenging datasets such as CIFAR10, CIFAR100, and MiniImageNet, the proposed method achieves $\small 0.97\times$, $\small 0.75\times$, $\small 0.69\times$ the performance of classic CL in the DECL-APIs setting.
The {main contributions} in this work can be summarized as three-fold:
\begin{itemize}[noitemsep,topsep=0pt,parsep=0pt,partopsep=0pt]
    \item We define two practical-yet-novel CL settings: data-efficient CL (DECL-APIs) and data-free CL (DFCL-APIs). To the best of our knowledge, this is the first work exploring how to do CL without access to full raw data; instead, we learn from a stream of black-box APIs.
    \item We propose a novel data-free cooperative continual distillation learning framework that just queries APIs. Our framework can effectively perform CL with data-free, black-box APIs, arbitrary model scales, and mitigating catastrophic forgetting. 
    \item We conducted extensive experiments to verify the effectiveness of the proposed framework in the DECL-APIs and DFCL-APIs settings. \revised{The experiment covers computer vision and natural language processing domains, and `soft and hard labels' and `homogeneous and heterogeneous' API settings.}
\end{itemize}

This paper is organized as follows: In Sec.~\ref{sec:intro}, we introduce the research motivation of this paper. In Sec.~\ref{sec:relatedwork}, we describe related work. In Sec.~\ref{sec:setting}, we define two new continual learning settings that perform CL from a stream of APIs. In Sec.~\ref{sec:method}, we introduce the proposed framework. In Sec.~\ref{sec:experiment}, we conduct extensive experiments to verify the effectiveness of the proposed framework. Finally, we conclude this paper in Sec.~\ref{sec:conclusion}.

\section{Related Works}\label{sec:relatedwork}
Our related work mainly includes the following three aspects of research: continual learning, model inversion, and learning from a stream of pre-trained models.

\subsection{Continual Learning} 
The goal of CL is to use a machine learning model to continuously learn new tasks without forgetting old tasks. Classic CL methods can be roughly classified into the following five categories~\cite{Forgetting_Survey_2023}:
1) \emph{Memory-based approaches} store a small number of old task samples and use episodic memory to replay data from previous tasks when updating the model with a new task~\cite{er,gem_nips2017,Hindsight_aaai2021,TaskfreeContinualLearning_DRO_ICML2022,dc_plugin_neurips2023}. This type of research method mainly focuses on how to select limited and informative old task samples to prevent forgetting old tasks. 
2) \emph{Architecture-based approaches} are also called parameter isolation methods, and their core idea is that there is a subset of parameters that are not shared between tasks, thereby avoiding task interference. They can be further divided into fixed-capacity networks~\cite{Piggyback_eccv2018,hat,Packnet} and increased-capacity networks~\cite{DEN_ICLR_2018}, respectively. 
3) \emph{Regularization-based approaches} add a constraint term when updating the network with new tasks to avoid catastrophic forgetting. These methods heuristically calculate each parameter's importance to the old tasks~\cite{ewc,SynapticIntelligence,mas_eccv2018,rwalk}. When the new task updates the network, it mainly updates the parameters that are not important to the old tasks, and reduces the update of the parameters that are important to the old tasks. In addition, inspired by knowledge distillation~\cite{knowledgedistillation_survey2021}, some works introduce distillation regularization to alleviate forgetting~\cite{lwf,lwm_cvpr2019}. They use the model trained on the old task as a teacher model to guide the learning of the CL model (as a student model) on the new task.
4) \emph{Subspace-based approaches} projects old and new tasks into different subspaces, thereby alleviating conflicts or interference between tasks. According to the construction method of subspace, it can be further divided into projection in feature (or input) subspace~\cite{gpm_iclr2021,trgp_ICLR2022,fs-dgpm,dfgp_iccv2023,owm} or projection in gradient subspace~\cite{ogd,chaudhry2020continual}.
5) \emph{Bayesian approaches} alleviate forgetting by combining uncertainty estimation and regularization techniques. It can be further divided into methods such as weight space regularization~\cite{kurle2019continual, henning2021posterior}, function space regularization~\cite{pan2020continual,kapoor2021variational}, and Bayesian structural adaptation~\cite{kumar2021bayesian}.

However, all the above methods need to use the raw data to perform CL. Due to copyright considerations and the risk of sensitive information leakage, it is difficult for us to obtain the full raw data, so the existing CL methods cannot work. In this paper, we propose two more novel-yet-practical CL settings that learn from a stream of APIs instead of raw data, which will further expand the application range of CL.

\subsection{Model Attack/Extraction} 
\revised{
The goal of model inversion/attack/stealing/extraction~\cite{oliynyk2023know} is to steal the parameters~\cite{naseh2023stealing}, training hyperparameters~\cite{wang2018stealing}, or functionality~\cite{dfad_2019,DataFreeModelExtraction2021CVPR} of a given victim model. In particular, this paper focuses on how to extract the \textit{functionality} of a given victim into a clone model.} Depending on whether the parameters of a given victim model are known or not, it is further divided into white-box and black-box model inversion~\cite{DataFreeKnowledgeTransferASurvey}.
1) \emph{In a white-box setting}, the attacker has access to the architecture and parameters of the victim model~\cite{fredrikson2015model}. The basic idea of such methods is to use gradient-based optimization to generate some samples of optimal representation in each target class~\cite{PGD_iclr2018,invertinggradient_2020,dafl_iccv2019,dfad_2019,datafree_ijcai2021,fang2022up}.
\revised{For example, DAFL~\cite{dafl_iccv2019} and DFAD~\cite{dfad_2019} propose to train a generative adversarial network (GAN)~\cite{goodfellow2020generative} network to generate fake images that approximate the original training set, and then use the generated images to extract information about the victim model.}
However, due to privacy and copyright protection, the victim model cannot usually be accessed in a white-box manner; it can only be accessed through a black-box API.
2) \emph{In a black-box setting}, since the parameters of the victim model cannot be accessed~\cite{GAMIN2019}, it is impossible to generate images of the target class through gradient-based optimization directly. 
Therefore, some works choose to use auxiliary data similar to the original data to steal the knowledge of the API~\cite{StealingFunctionalityCVPR2019,barbalau2020black,dfnd_cvpr2021}. \revised{However, finding highly relevant auxiliary data is still difficult, especially when dealing with some unconventional tasks (e.g., medical datasets of various diseases)~\cite{maze_cvpr2021}. }
Some works try to optimize stochastic noise or generative adversarial networks using zeroth-order optimization~\cite{wibisono2012finite,ZOO2017,liu2020primer} to generate target class images~\cite{maze_cvpr2021,DataFreeModelExtraction2021CVPR}. \revised{For example, MAZE~\cite{maze_cvpr2021} and DFME~\cite{DataFreeModelExtraction2021CVPR} perform data-free knowledge distillation under soft labels by estimating gradients in a forward difference method. Other works~\cite{zhang2022towards,zhang2023ideal} use the output of the student model as a proxy for a black-box API to explicitly optimize the generator. Further, ZSDB3KD~\cite{wang2021zero} and DFMS-HL~\cite{sanyal2022towards} explore data-free knowledge distillation in the more challenging black-box hard-label setting. There are also some works that apply data-free knowledge distillation to data-free black-box adversarial attack task~\cite{BrendelRB18, cheng2018query,DaST_CVPR2020,wang2021delving,FE-DaST_2022}.
}

\revised{
However, these works are limited to extracting the knowledge of a single victim model into a single clone model, which lacks the ability of continual learning, that is, continuously extracting the knowledge of different victim models to enhance the clone model's ability. When one clone model is used to learn from a stream of victim models, existing cloning schemes suffer from catastrophic forgetting. That is, after learning the functionality of the new victim, the clone model forgets the other functionality it learned before. This paper proposes a novel framework to fill this research gap. Our framework allows a single clone model to continuously assimilate new functionalities from a range of black-box APIs without forgetting previously learned capabilities.
}

\subsection{Learning From a Stream of Pre-train Models}

Due to increasingly stringent privacy protection, training data for many tasks are no longer publicly available, so traditional machine learning methods that learn from raw data may be limited.
Recently, a small number of works~\cite{wang2022metauai2022,ExModel2022} have emerged to learn from a collection of pre-trained models.
Data-free meta-learning~\cite{wang2022metauai2022} is a meta-learning method that learns to initialize meta-parameters from an online pre-trained model stream without accessing their training data. However, it assumes that the pre-trained models are white-box with the same architecture. 
Ex-Model~\cite{ExModel2022} performs CL using a stream of pre-trained models and imposes strict restrictions on them, such as requiring them to be white-box and small-scale, limiting its applicability. 

However, our approach significantly expands the application of CL by only querying APIs. In general, the method proposed in this paper does not need to know API's network architecture, does not need to backpropagate a large-scale network, does not need to access the raw training data, and effectively alleviates catastrophic forgetting.

\begin{figure*}[t]
\centering
\includegraphics[width=1.0\textwidth]{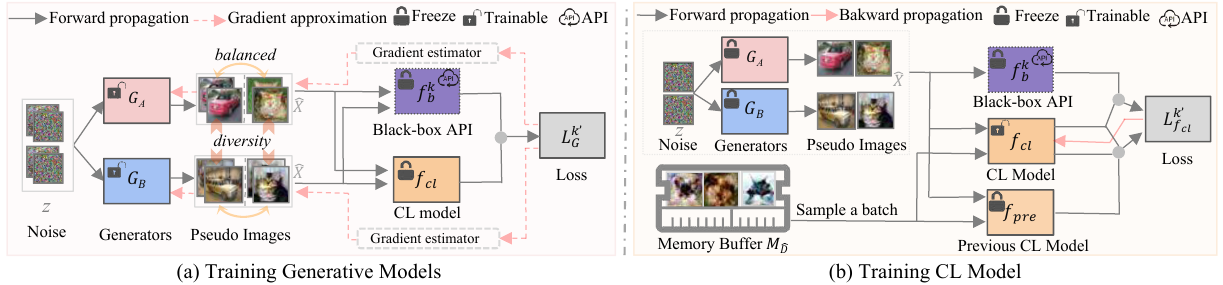}
\vspace{-20pt}
\caption{Illustration of the training process of our proposed data-free cooperative continual distillation learning framework. Our framework consists of two collaborative generative models $\small \{G_{A}, G_{B}\}$, a CL model $\small f_{cl}$, and a stream of black-box APIs $\small \{f_{b}^1, \ldots, f_{b}^k, \ldots, f_{b}^K\}$. 
}
\vspace{-15pt}
\label{fig:framework}
\end{figure*}

\section{Settings}
\label{sec:setting}

In this section, we first introduce the classic CL, which learns from a stream of raw data (see Fig.~\ref{fig:scenario}(a)), 
and then propose DFCL-APIs and DECL-APIs, which learn from a stream of APIs (see Fig.~\ref{fig:scenario}(b)).

 \subsection{Preliminary on CL: Classic Continual Learning From a Stream of Raw Data (Classic CL)}
\label{subsec:classic_settings}

The goal of CL is to enable a machine learning model to continuously learn new tasks without forgetting old tasks. In the classic CL setting, the full raw data $\small \mathcal{D}^k$ of the new task $k$ is available. Therefore, the objective of task $k$ is to minimize the empirical risk loss of $f_{cl}$ (parameters $\mathbf{\theta}_{f_{cl}}$): 
\begin{equation}
    \underset{f_{cl}}{\arg \min } \; \mathbb{E}_{(x_j^k,y_j^k) \sim \mathcal{D}^k}  \left[ \mathcal{L}^k \left(f_{cl}(\boldsymbol{x}_j^k), y_j^k\right)\right],   
\end{equation}
where $\small \mathcal{D}^k \!= \!(\mathbf{X}^k, \mathbf{Y}^k)$ is the data of task $k$, $\small k \! \in \! \{1,2,\ldots,K\}$ is the task id, and $\small (x_j^k, y_j^k)$ represents the $j$-th sample pair in the task $k$'s data $\small \mathcal{D}^k$. $\small \mathcal{L}^k(\cdot)$ represents the loss function of the $k$-th task, such as KL-divergence, cross-entropy, $L_1$ norm, etc.
To avoid catastrophic forgetting of old tasks, mainstream memory-based CL methods~\cite{er,TaskfreeContinualLearning_DRO_ICML2022} usually store a small number of raw samples from old tasks in a fixed-capacity memory $\small \mathcal{M}_{\mathcal{D}}$ for experience replay when learning a new task. Therefore, a classic CL algorithm $\small \mathcal{A}_{CL}$ can be expressed as follows:
\begin{equation}
\small
    \mathcal{A}_{CL}: \text{Input } \left\{ f_{cl}, \mathcal{M}_{\mathcal{D}}^{k-1}, \mathcal{D}^k, k \right\} \rightarrow  \text{Output } \left\{f_{cl}, \mathcal{M}_{\mathcal{D}}^k \right\},
\label{eq:classiccl_setting}
\end{equation}
where the input of the $\small \mathcal{A}_{CL}$ is the CL model $f_{cl}$, the memory $\small \mathcal{M}_{\mathcal{D}}^{k-1}$, the training data $\small \mathcal{D}^k$ and task ID $k$, and the output of the $\mathcal{A}_{CL}$ is the updated CL model $\small f_{cl}$ and memory $\small \mathcal{M}_{\mathcal{D}}^k$ on task $k$.  

However, raw data are highly sensitive, so stakeholders usually publish pre-trained machine learning models as a service (MLaaS) that users can access through online APIs~\cite{roman2009model_maas}. Therefore, existing methods cannot work, which greatly limits the application of CL.

\subsection{Continual Learning From a Stream of APIs: DFCL-APIs and DECL-APIs}
\label{subsec:ours_settings}

In this section, we propose two more novel-and-practical CL settings, called DFCL-APIs and DECL-APIs, which perform CL from a stream of black-box APIs with little or no raw data.

\noindent
\textbf{Data-Free Continual Learning From a Stream of APIs (DFCL-APIs)}.
In this setting, we do not have access to the raw training data $\small \mathcal{D}^{k}$ of task $k$. Instead, we can only access a black-box API $\small f^{k}_b$, the architecture and parameters of which are unknown, that has been trained with the corresponding data $\small \mathcal{D}^k$ of task $k$. That is, we can only input an image $\hat{\mathbf{x}}_i$ to the API $\small f^{k}_b$ to obtain the API's response, i.e., $\small \hat{y}^k_i=f^{k}_b(\hat{\mathbf{x}}_i)$, where $\small \hat{\mathbf{x}}_i \notin \mathcal{D}^k$ is the generated pseudo image. Therefore, in our DFCL-APIs setting, the objective of the CL model $f_{cl}$ for task $k$ is to match the output of the black-box API $f_b^k$ on the generated image set $\small \hat{\mathcal{D}}_{\mathcal{G}}^k$, as shown in the following formula:
\begin{equation}
\small
    \underset{f_{cl}}{\arg \min } \; \mathbb{E}_{\hat{x}_i \sim \hat{\mathcal{D}}_{\mathcal{G}}^k} \left[ \mathcal{L}^k \left(f_{cl}(\hat{\mathbf{x}}_i), f^k_b(\hat{\mathbf{x}}_i) \right) \right].
\end{equation}
Similar to the classic CL setting, to avoid catastrophic forgetting, we can store a small number of generated data in memory $\small \mathcal{M}_{\hat{\mathcal{D}}}$ for replay. The DFCL-APIs algorithm can be expressed as:
\begin{equation}
\small
    \mathcal{A}_{DFCL-APIs}: \text{Input } \left\{ f_{cl}, \mathcal{M}_{\hat{\mathcal{D}}}^{k-1}, f_b^k, k \right\} \rightarrow  \text{Output } \left\{ f_{cl}, \mathcal{M}_{\hat{\mathcal{D}}}^k \right\}.
\label{eq:dfcl_setting}
\end{equation}
DFCL-APIs setting differs from the Ex-Model~\cite{ExModel2022} in that we only require access to the pre-trained model as a black-box API, whereas Ex-Model requires the model architecture and parameters (white-box). Obtaining a white-box model in MLaaS is often not possible due to intellectual property protection.

\noindent
\textbf{Data-Efficient Continual Learning From a Stream of APIs (DECL-APIs)}. 
In some cases, we may have access to a small amount of raw data from the API~\cite{papernot2017practical_jbda}. For example, the API may use some public data for training~\cite{brown2020language_gpt3}. Therefore, we can store this data in memory $\mathcal{M}_{\mathcal{D}}$ for replay. This setting is defined as data-efficient CL (DECL-APIs), and the only difference between the DECL-APIs in Eq.~\ref{eq:decl_setting} and the DFCL-APIs in Eq.~\ref{eq:dfcl_setting} is whether a small number of raw data (DECL-APIs) or generated data (DFCL-APIs) of old tasks is stored in memory. The DECL-APIs algorithm can be formulated as:
\begin{equation}
\small
    \mathcal{A}_{DECL-APIs}: \text{Input } \left\{f_{cl},  \mathcal{M}_{\mathcal{D}}^{k-1}, f_b^k, k \right\} \rightarrow  \text{Output } \left\{f_{cl}, \mathcal{M}_{\mathcal{D}}^k \right\}.
\label{eq:decl_setting}
\end{equation}

\noindent
\textbf{{The Significance of New Settings}}:
DECL-APIs and DFCL-APIs allow us to perform CL from APIs without access to the entire raw training data. On the one hand, this greatly protects the privacy of the original training data, and on the other hand, it greatly expands the real application of CL.

\noindent
\textbf{{The Challenges of New Settings}}:
In general, DECL-APIs and DFCL-APIs are significantly more challenging than classic CL for several reasons: 
(i) The complete raw training data is not available.
(ii) Without knowledge of the architecture and parameters of each API.
(iii) The architecture of each API may be heterogeneous or of arbitrary scale.
(iv) Learning on a new API can lead to the catastrophic forgetting of old APIs.

\section{Methodology} 
\label{sec:method}

To address the challenges of DFCL-APIs and DECL-APIs settings in a unified framework, we propose a novel data-free cooperative continual distillation learning framework, as shown in Fig.~\ref{fig:framework}.
 
\subsection{End-to-End Learning Objective}
\label{sec:ourobjective}

We aim to make a CL model $\small f_{cl}$ learn from a stream of black-box APIs $\small \{f_b^k\}_{k=1}^K$. We propose to train two cooperative generators $\small \mathcal{G}=\{\mathcal{G}_A, \mathcal{G}_B\}$ to generate `hard', `diverse', and `class-balanced' pseudo samples. Specifically, for each task $k$, to optimize the CL model $\small f_{cl}$ and generators $\small \mathcal{G}$, we formulate their training as an \textit{adversarial game}.
On the one hand, generators $\small \mathcal{G}$ adversarially generate pseudo images that \textbf{maximize} the gap between the API and the CL model. 
On the other hand, the black-box API $\small f_{b}^k$ and the CL model $\small f_{cl}$ jointly play the role of the discriminator. They aim to \textbf{minimize} their output gaps on the pseudo images so that the CL model can learn from the API's knowledge. 
We alternately train $f_{cl}$ and $\small \mathcal{G}$ and the learning objective of an adversarial game can be written as:
\begin{equation}
\small
    \min _{f_{cl}}\max _{\mathcal{G}}\;\;   \mathbb{E}_{\mathbf{z}_i \sim \revised{\mathcal{N}({0},{I})}} \sum_{k=1}^K  \left[\ell^k \left(f_b^k \left(\mathcal{G}(\mathbf{z}_i) \right), f_{cl} \left(\mathcal{G}(\mathbf{z}_i)\right)\right)\right],
\end{equation}
where $\small \ell^k(\cdot,\cdot)$ is a distance function, without loss of generality, this paper uses $L_1$ loss.
$\mathbf{z}_i$ represents the noise vector sampled from a Gaussian distribution $\small \revised{\small \mathcal{N}({0},{I})}$. 
The proposed framework is shown in Fig.~\ref{fig:framework} and the algorithmic flow is summarized in Alg.~\ref{alg:our} \revised{in Appendix~\ref{subsec:AlgorithmDetails_appendix}}.
Below, we provide details on how to train generators $\mathcal{G}$ and the CL model $f_{cl}$ in Sec.~\ref{sec:trainning_generative} and Sec.~\ref{sec:trainning_cl}, respectively.

\subsection{Training Generative Models}
\label{sec:trainning_generative}

We trained generators $\small \mathcal{G}$ to generate \textit{hard, diverse, and class-balanced} pseudo images in this section, shown in Fig.~\ref{fig:framework}(a) and then used these generators in Sec.~\ref{sec:trainning_cl} to train the CL model. Specifically, we train the generators $\small \mathcal{G}\!=\!\{\mathcal{G}_A, \mathcal{G}_B\}$ by optimizing the following three objective functions: (1) adversarial generators, (2) cooperative generators, and (3) class-balance.

\noindent
\textbf{(1) Adversarial Generator Loss.}
The main goal of our generators $\small \mathcal{G}\!=\!\{\mathcal{G}_A, \mathcal{G}_B\}$  is to generate ‘hard’~\footnote{\footnotesize 
\revised{
{Definition of a `hard' sample}: Denote the pseudo images generated by generator $\mathcal{G}(\cdot)$ with random noise $\textbf{z}_i$ and random noise $\textbf{z}_j$ as $\hat{\textbf{x}}_i=\mathcal{G}(\textbf{z}_i)$ and $\hat{\textbf{x}}_j=\mathcal{G}(\textbf{z}_j)$. If $\ell^k \left(f_b^k \left(\hat{\textbf{x}}_i\right), f_{cl} \left(\hat{\textbf{x}}_i\right)\right) < \ell^k \left(f_b^k\left(\hat{\textbf{x}}_j\right), f_{cl}\left(\hat{\textbf{x}}_j\right)\right)$ is satisfied, we make $\hat{\textbf{x}}_j$ a harder sample than $\hat{\textbf{x}}_i$. Since our generators and CL model adopt the mature framework of adversarial training~\cite{gan_2014}, the generator's generation ability will be enhanced during the adversarial game process, that is, harder samples will be generated continuously.
}} 
pseudo images that maximize the output gap (such as $L_1$ norm, etc.) between the CL model $f_{cl}$ and the black-box API $f^k_b$. 
Therefore, we fixed the CL model $\small f_{cl}$ and queried the API $\small f^k_b$ to train the generators $\small \mathcal{G}$. The loss function of the generators $\small \mathcal{G}$ is defined as follows:
\begin{equation}
\small
\begin{split}
    \mathcal{L}_{\mathcal{G}}^k 
     & =   \max_{\mathcal{G}}\; \mathbb{E}_{\mathbf{z}_i \sim \revised{\mathcal{N}({0},{I})}} \ell^k \left(f_b^k \left(\mathcal{G}(\mathbf{z}_i) \right), f_{cl} \left(\mathcal{G}(\mathbf{z}_i)\right)\right) 
    \\
     &\approx     - \min_{\mathcal{G}} \frac{1}{N} \sum_{i=1}^{N} \ell^k \left(f_b^k \left(\mathcal{G}(\mathbf{z}_i) \right), f_{cl} \left(\mathcal{G}(\mathbf{z}_i)\right)\right),  
\end{split}
\label{eq:loss_generate}
\end{equation}
where \revised{$\small \mathbf{z}_i \sim \mathcal{N}({0},{I})$} and $\small N$ represents the number of generated samples. \revised{Transforming from the first row to the second row means that we use the average of $\small N$ samples to approximate the overall expected value.}
However, computing the gradient $\small \nabla_{\theta_{\mathcal{G}}} \mathcal{L}_{\mathcal{G}}^k$ w.r.t $\mathcal{G}$ in Eq.~\ref{eq:loss_generate} is not possible because $f_b^k$ is a black-box API.
To address this issue, we use zeroth-order gradient estimation~\cite{wibisono2012finite,ZOO2017,maze_cvpr2021,DataFreeModelExtraction2021CVPR} to approximate the gradient of the loss $\small \mathcal{L}_{\mathcal{G}}^k$ w.r.t the generators $\small \mathcal{G}$.
Unfortunately, generators $\mathcal{G}$ usually contain a large number of parameters, and obtaining accurate gradient estimates $\small \hat{\nabla}_{\theta_{\mathcal{G}}}$ directly through zeroth-order gradient estimation in such a large parameter space~\cite{maze_cvpr2021,DataFreeModelExtraction2021CVPR} is costly. Therefore, \revised{we decompose the gradient estimation process into two steps \revised{(Pseudocode is provided in Alg.~\ref{alg:implement_zero}.)}. We first try to estimate the gradient $\small \partial \mathcal{L}^k_{\mathcal{G}} / \partial \hat{\mathbf{x}} $ of the output image $\small  \hat{\mathbf{x}} \!=\! \mathcal{G}(\mathbf{z}) $ from $\small \mathcal{G}$ by forward differences method~\cite{wibisono2012finite} (i.e., a zeroth-order method). Then, We can use the image $\small \hat{\mathbf{x}}$ to directly backpropagate the gradient $\small {\partial \hat{\mathbf{x}} }/{\partial \theta_{\mathcal{G}}}$ of the generator $\small \mathcal{G}$}~\footnote{\footnotesize \revised{This can be easily done with an automatic differentiation tool such as PyTorch, that is, the gradient $\frac{\partial \hat{\mathbf{x}} }{\partial \theta_{\mathcal{G}}}$ is given by $\hat{\mathbf{x}}$.backward($\frac{\partial \mathcal{L}^k_{\mathcal{G}}}{\partial \hat{\mathbf{x}} }$). $\frac{\partial \mathcal{L}^k_{\mathcal{G}}}{\partial \hat{\mathbf{x}} }$ represents the gradient estimation result in the second row of Eq.~\ref{eq:gradient_generate_zerothorder}.}}:
\begin{equation}
\begin{split}
\small
     \hat{\nabla}_{\theta_{\mathcal{G}}} \mathcal{L}^k_{\mathcal{G}} &=\frac{\partial \mathcal{L}^k_{\mathcal{G}}}{\partial \theta_{\mathcal{G}}}=\frac{\partial \mathcal{L}^k_{\mathcal{G}}}{\partial \hat{\mathbf{x}} } \times \frac{\partial \hat{\mathbf{x}} }{\partial \theta_{\mathcal{G}}}, \;
     \\
    \frac{\partial \mathcal{L}^k_{\mathcal{G}}}{\partial \hat{\mathbf{x}} } &= \frac{d \cdot \left(\mathcal{L}^k_{\mathcal{G}} \left(\hat{\mathbf{x}} +\epsilon \mathbf{u}_{{i}}\right)- \mathcal{L}^k_{\mathcal{G}}(\hat{\mathbf{x}} )\right)}{\epsilon} \mathbf{u}_{i},
\end{split}
\label{eq:gradient_generate_zerothorder}
\end{equation}
where $\mathbf{u}_{i}$ is a random variable with uniform probability drawn from a $d$-dimensional unit sphere, $\epsilon$ is a positive constant called the smoothing factor.
So, we can update $\mathcal{G}$ by $\small \hat{\nabla}_{\theta_{\mathcal{G}}} \mathcal{L}^k_{\mathcal{G}}$ by Eq.~\ref{eq:gradient_generate_zerothorder}.

To comprehensively explore the API's knowledge, we further impose two constraints on generators $\mathcal{G}$: the generated images should be as diverse as possible and cover each class uniformly.

\noindent
\textbf{(2) Cooperative Generators Loss.}
Diversified pseudo data helps to improve the generalization and robustness, and enable comprehensive exploration of the knowledge contained in the API. We achieve the diversity goal by maximizing the image gap between collaborative generators $\small \mathcal{G}_A$ and $\small \mathcal{G}_B$. Specifically, for the same random noise vector $\mathbf{z}_i$, we make the pseudo images $\small \mathcal{G}_A(\mathbf{z}_i)$ and $\small \mathcal{G}_B(\mathbf{z}_i)$ generated by $\small \mathcal{G}_A$ and $\small \mathcal{G}_B$ as dissimilar as possible, that is, maximize their dissimilarity, formally expressed as:
\begin{equation}
\small
\begin{split}
    \mathcal{L}_{C}^k  &=  \max _{\mathcal{G}_A, \mathcal{G}_B}\;  \mathbb{E}_{\mathbf{z}_i \sim \revised{\mathcal{N}({0},{I})}} d^{\mathcal{G}} \left(\mathcal{G}_A \left(\mathbf{z}_i\right) \right), \left(\mathcal{G}_B \left(\mathbf{z}_i\right) \right)
    \\
    &\approx   - \min _{\mathcal{G}_A, \mathcal{G}_B}\; \frac{1}{N} \sum_{i=1}^{N} d^{\mathcal{G}}  \left(\mathcal{G}_A \left(\mathbf{z}_i \right) \right), \left(\mathcal{G}_B \left(\mathbf{z}_i \right) \right),
\end{split}
\label{eq:loss_diversity}
\end{equation}
where \revised{$\small \mathbf{z}_i \sim \mathcal{N}({0},{I})$} and $\small d^{\mathcal{G}}(\cdot,\cdot)$ is a distance function, e.g., the $\small L_1$ norm, and $N$ represents the number of samples.

\noindent
\textbf{(3) Class-balanced Loss.}
To ensure that the number of images generated by each class is as even as possible, we use information entropy to measure the balance of generated images as~\cite{dafl_iccv2019}. We use the CL model $f_{cl}$ to obtain the distribution of images for each class. The class balance loss is defined as:
\begin{equation}
\small
\begin{split}
    \mathcal{L}_{B}^k  =  &\min _{\mathcal{G}_A, \mathcal{G}_B}\; \frac{1}{C} \sum_{c=1}^C {\bar{y}_A}(c) \log ({\bar{y}_A}(c)) + \frac{1}{C} \sum_{c=1}^C {\bar{y}_B}(c) \log ({\bar{y}_B}(k)),\\
    &{\bar{y}_A}\!=\!\frac{1}{N} \sum_{i=1}^N \revised{{f}_{cl}} \left(\mathcal{G}_A(\mathbf{z}_i) \right),  \,{\bar{y}_B}\!=\!\frac{1}{N} \sum_{i=1}^N \revised{{f}_{cl}} \left(\mathcal{G}_B(\mathbf{z}_i) \right)
\end{split}
\label{eq:loss_balanced}
\end{equation}
where \revised{$\small \mathbf{z}_i \!\sim\! \mathcal{N}({0},{I})$ represents the sampled random noise}, $\bar{y}_A(c)$ is the $c$-th element of $\bar{y}_A$, $\small C$ is the number of classes. 
If $\mathcal{L}_{B}^k$ achieves its minimum value, every element within $\bar{y}_A$ and $\bar{y}_B$ will be equal to $\small 1/C$. This implies that $\small \mathcal{G}$ can generate samples for each class with an equivalent probability.

\noindent
\textbf{Total Loss of Generative Models.}
By considering the three losses above simultaneously, we can derive the total loss of generators as follows:
$
\small
    \mathcal{L}_{G}^{k^{\prime}}  \!=\! \mathcal{L}_{\mathcal{G}}^k \!+\! \lambda_{G}  \cdot (\mathcal{L}_{C}^k  \!+\!  \mathcal{L}_{B}^k),
$
where hyperparameter $\small \lambda_G$ is utilized to control the strength of diversity and balance losses. We found that simply setting it to $1$ worked, and further tuning in the future might lead to further performance improvements. Since the losses $\small \mathcal{L}_{C}^k$ and $\small \mathcal{L}_{B}^k$ do not involve the black-box API $\small f_b^k$, their gradients can be calculated directly by backpropagation. The loss $\small \mathcal{L}_{\mathcal{G}}^k$ uses Eq.~\ref{eq:gradient_generate_zerothorder} to obtain an estimated gradient. After updating generators $\mathcal{G}$ using gradient descent, we can obtain generators that are able to generate hard, diverse, and class-balanced images.

\subsection{Training CL Model}
\label{sec:trainning_cl}
In this section, we use generators $\mathcal{G}=\{\mathcal{G}_A, \mathcal{G}_B\}$ trained in Sec.~\ref{sec:trainning_generative} to generate pseudo images to query the \revised{$k$-th} black-box API $f_b^k$ and distill its knowledge into the CL model $f_{cl}$, shown in Fig.~\ref{fig:framework}(b). We also use an old data replay mechanism and propose a regularization based on network similarity to avoid the CL model forgetting old APIs. Specifically, we train the CL model $\small f_{cl}$ by optimizing the following three objective functions: (1) adversarial distillation, (2) memory replay, and (3) network similarity.

\noindent
\textbf{(1) Adversarial Distillation Loss.}
The CL model $\small f_{cl}$ is trained to imitate the black-box API $\small f_{b}^k$, and generators $\small \mathcal{G}$ remain fixed. That is, for task $k$, we aim to minimize the difference between the output of the API $\small f_{b}^k$ and the CL model $\small f_{cl}$ on the generated data $\small \mathcal{G}(\mathbf{z})$. The distillation loss of the CL model is defined as:
\begin{equation}
\setlength{\abovedisplayskip}{3pt}
\setlength{\belowdisplayskip}{3pt}
\small
\begin{split}
     \mathcal{L}_{f_{cl}}^k &=  \min_{f_{cl}}\;   \mathbb{E}_{\mathbf{z}_i \sim \revised{\mathcal{N}({0},{I})}} \ell^k \left(f_b^k \left(\mathcal{G}(\mathbf{z}_i) \right), f_{cl} \left(\mathcal{G}(\mathbf{z}_i)\right)\right)
     \\ 
     &\approx  \revised{\frac{1}{N}} \min _{f_{cl}}\; \sum_{i=1}^{N} \ell^k \left(f_b^k \left(\mathcal{G}(\mathbf{z}_i) \right), f_{cl} \left(\mathcal{G}(\mathbf{z}_i)\right)\right),
\end{split}
\label{eq:loss_cl}
\end{equation}
where \revised{$\small \mathbf{z}_i \sim \mathcal{N}({0},{I})$ is random noise} and $\small N$ is the number of generated samples. 
However, transferring the knowledge of task $k$ (i.e., API $f_{b}^k$) to the CL model $f_{cl}$ directly via Eq.~\ref{eq:loss_cl} will lead to catastrophic forgetting of previously learned APIs. 
We further adopt a memory replay mechanism and a network similarity regularization to solve this problem.

\noindent
\textbf{(2) Memory Replay Loss}. 
In the \textit{DFCL-APIs} setting, we store a small portion of generated images $\small \hat{\mathbf{X}}$ in a memory buffer $\mathcal{M}$ after learning each task $k$ as~\cite{gem_nips2017,er,TaskfreeContinualLearning_DRO_ICML2022,buzzega2020dark_der2022}, i.e.:
\begin{equation}
\setlength{\abovedisplayskip}{3pt}
\setlength{\belowdisplayskip}{3pt}
\small
\begin{split}
    \mathcal{M} \!\gets\! \mathcal{M} \!\cup\! \hat{\mathcal{D}}^{k} &, \; \hat{\mathcal{D}}^{k} \!\gets\! \left(\hat{\mathbf{X}}^k, \hat{\mathbf{Y}}^k, k \right), \\
    \hat{\mathbf{Y}}\!\gets \!f_b^{k} (\hat{\mathbf{X}}^k) &,\; \hat{\mathbf{X}}^k \!\gets \!\mathcal{G}\left(\mathbf{z}; \theta_{\mathcal{G}}\right), \mathbf{z} \sim \revised{\mathcal{N}({0},{I})}, 
\end{split}
\label{eq:dfcl_memory}
\end{equation}
where $\small \hat{\mathbf{X}}^{{\prime}}$ denotes a batch of generated samples, $\small f_b^{k} (\hat{\mathbf{X}} ) $ represents the output logits of the black-box API $\small f_b^{k}$ for the input images $\small \hat{\mathbf{X}}$.
In the \textit{DECL-APIs} setting,
since a small amount of raw data $\small {\mathbf{X}}$ is available,  we store this part of the raw images in memory $\small \mathcal{M}$ for replay, that is:
\begin{equation}
\small
\begin{split}
    & \mathcal{M}  \! \gets  \! \mathcal{M} \cup {\mathcal{D}}^k , \; {\mathcal{D}}^k \! \gets  \! \left({\mathbf{X}}^k, {\mathbf{Y}}^k, k \right), {\mathbf{Y}}^k\! \gets  \!f_b^{k} ({\mathbf{X}^k})
\end{split}
\label{eq:DECL-APIs_memory}
\end{equation}
where $\small f_b^{k} ({\mathbf{X}^k} )$ represents the output logits of the API $\small f_b^{k}$ w.r.t the input data $\small {\mathbf{X}}^k$. 
Therefore, in addition to learning the knowledge of the new API $\small f_b^k$, we must also ensure that the knowledge stored in memory $\small \mathcal{M}$ is not forgotten. We constrain the $\small f_{cl}$ model to remember the knowledge stored in memory $\small \mathcal{M}$ by the old tasks. As a result, the loss of samples in memory $\small \mathcal{M}$ is as follows:
\begin{equation}
\setlength{\abovedisplayskip}{3pt}
\setlength{\belowdisplayskip}{3pt}
\small
\begin{split}
     \mathcal{L}_{{M}}^k = \min_{f_{cl}} \frac{1}{|\mathcal{M}|} \sum_{({\mathbf{X}^{\prime}}, {\mathbf{Y}^{\prime}}, k^{\prime}) \in \mathcal{M}} \ell^{k^{\prime}} \left({Y}^{\prime}, f_{cl} ({\mathbf{X}^{\prime}} ), k^{\prime} \right),
\end{split}
\label{eq:loss_memory}
\end{equation}
where $\small k^{\prime} \!\!\in\!\! \{1,\ldots,k\!-\!1\}$ and $\small |\mathcal{M}|$ is the number of samples in $\small \mathcal{M}$.

\noindent
\textbf{(3) Network Similarity Loss}. 
We have also designed a new regularization term based on the network similarity measures~\cite{DistanceCorrelation_eccv2022} to reduce catastrophic forgetting further. Specifically, we expect that when learning a new task $k$, the training samples $\small {\mathbf{X}}$ will have similar output features at each layer or module of both the current CL model $f_{cl}$ and the CL model (denoted as $f_{pre}$) trained by previous tasks.
To achieve this goal, we first copy the CL model and freeze its network parameters after training for task $k\!-\!1$. Then, when task $k$ is being learned, we feed the training sample $\small {\mathbf{X}}$ to $f_{cl}$ and $f_{pre}$  to obtain the output features $F_{cl}^l$ and $F_{pre}^l$ for each training sample of each layer $l$, respectively.
Then, we maximize the similarity between $\small f_{cl}$ and $\small f_{pre}$ to ensure that the CL model for a new task is as similar as possible to the CL model for the old tasks to achieving unforgotten, as below:
\begin{equation}
\small
\centering
\begin{split}
  \mathcal{L}_{S}^k  &=  \min_{f_{cl}} \frac{1}{|\mathbf{X}|} \sum_l \frac{ -\langle \mathbf{A}^l, \mathbf{B}^l \rangle}{\sqrt{\langle \mathbf{A}^l, \mathbf{A}^l \rangle \langle \mathbf{B}^l, \mathbf{B}^l \rangle}} 
  \\
     \text{s.t. } A_{ij}^l &= \| F_{cl}^l[i] - F_{cl}^l[j] \|_2, F_{cl}^l=f_{cl}^l( {\mathbf{X}}) ; 
     \\
     B_{ij}^l &= \| F_{pre}^l[i] - F_{pre}^l[j] \|_2, F_{pre}^l=f_{pre}^l( {\mathbf{X}})
\end{split}
\label{eq:loss_distance_correlation}
\end{equation}
where $l$ represents the $\small l$-th layer of the CL model $\small f_{cl}$, and $\small f_{cl}^l( {\mathbf{X}})$ represents the output of samples $\small \mathbf{X}$ in the $\small l$-th layer, that is, the feature representation. 
$F_{cl}^l[i]$ represents the output feature of the model $f_{cl}$ in layer $l$ for the $i$-th sample. $\small {\mathbf{X}}$ contains a total of $\small |{\mathbf{X}}|$ samples, which can come from the generated samples of the current task and the memory $\small \mathcal{M}$ formed by the previous tasks. 

\noindent
\textbf{Total Loss of CL Model}.  
In summary, for the CL model $f_{cl}$ to learn knowledge from the new task $k$ (i.e., API $f_b^k$) without forgetting knowledge of old APIs (i.e., $f_b^1, \ldots, f_b^{k-1}$), it needs to optimize the three objectives of Eq. ~\ref{eq:loss_cl}, Eq. ~\ref{eq:loss_memory}, and Eq. ~\ref{eq:loss_distance_correlation} simultaneously: 
$
\small
   \mathcal{L}_{f_{cl}}^{k'} \!=\! \mathcal{L}_{f_{cl}}^k \!+\! \lambda_{CL} \cdot (\mathcal{L}_{{M}}^k \!+\! \mathcal{L}_{S}^k),
$
where $\small \lambda_{CL}$ is the hyperparameter used to control the strength of memory replay and network similarity losses. We found that setting $\small \lambda_{CL}$ to 1 performed well, and we believe that fine-tuning it in the future could yield even better results. We backpropagate the total loss to obtain the gradient to update the CL model $f_{cl}$.

\subsection{Discussion}
Our proposed data-free cooperative continual distillation learning framework can effectively overcome the four challenges faced in the two new CL settings (DFCL-APIs, DECL-APIs) defined in Sec.~\ref{subsec:ours_settings}. Specifically,
(i) \textit{data-free}: Our framework trains two collaborative generators $\small \mathcal{G}=\{\mathcal{G}_A,\mathcal{G}_B\}$ to invert the $k$-th API's pseudo-training data $\small \hat{\mathbf{X}}^k$, thus requiring no access to the API's raw training data $\small {\mathbf{X}}^k$.
(ii) \textit{black-box APIs}: In the training steps of the generators $\small \mathcal{G}$ and the CL model $f_{cl}$, we just input the generated pseudo samples $\small \hat{\mathbf{X}}^k$ into the API $f_b^k$ to obtain the corresponding output of the API. Therefore, our framework does not need to know the architecture and parameters of the API $f_b^k$.
(iii) \textit{arbitrary model scales}: During the training of the generators $\small \mathcal{G}$, we obtain the estimated gradients of the generators through zeroth-order gradient optimization, so there is no need to backpropagate the API $f_b^k$. At the same time, the gradient calculation of the CL model $f_{cl}$ does not involve backpropagation to the API $f_b^k$. This means that APIs can have architectures of any scale.
(iv) \textit{mitigating catastrophic forgetting}: To alleviate the CL model's forgetting of old APIs, we replay a small number of pseudo-samples of old tasks (or a small amount of raw data, if and only if the DECL-APIs setting is applicable) in memory $\small \mathcal{M}$ to strengthen the knowledge of old APIs. In addition, our network similarity loss further constrains the change of network parameters, thereby reducing forgetting.

In summary, our framework is an effective method for learning from a stream of APIs, which has the advantages of data-free, black-box APIs, arbitrary model scales, and mitigating catastrophic forgetting. We believe this framework will effectively expand the scope of CL's application.

\section{Experiment}
\label{sec:experiment}
In this section, we conducted extensive experiments to verify the effectiveness of the proposed framework under DFCL-APIs and DECL-APIs settings. Due to page limits, part of the experimental setup and results are placed in {\bf Appendix}. 

\subsection{Experimental Setup}
\label{sec:ExperimentalSetup}
This section describes the datasets, baselines, and evaluation metrics used for the experiments.

\noindent
\textbf{Datasets.} 
We verified the performance of our proposed method on five commonly used CL datasets~\cite{gpm_iclr2021,trgp_ICLR2022,dfgp_iccv2023} \revised{in \emph{computer vision (CV)} domain} with different numbers of classes, tasks, and image resolutions: \revised{Split-}MNIST~\cite{mnist}, \revised{Split-}SVHN~\cite{svhn}, \revised{Split-}CIFAR10~\cite{cafar10_cifar100}, \revised{Split-}CIFAR100~\cite{cafar10_cifar100} and \revised{Split-}MiniImageNet~\cite{miniimagenet}. The dataset statistics are shown in Tab.~\ref{tab:datasetstatic} in Appendix~\ref{sec:DatasetStatistics_Appendix}. 
\revised{In addition, following the setting of \cite{de2019episodic,huang2021continual}, we construct five text datasets: AGNews (news classification), Yelp (sentiment analysis), Amazon (sentiment analysis), DBPedia (Wikipedia article classification) and Yahoo (questions and answers categorization) from \cite{zhang2015character}, in the \emph{natural language processing (NLP)} domain as the continual learning task. Statistics for the five NLP datasets are provided in Tab.~\ref{tab:datasetstatic_nlp} in Appendix~\ref{sec:DatasetStatistics_Appendix}.
}

\begin{table*}[t]
\small
\centering
\caption{Performance comparison on \revised{Split-} MNIST and SVHN datasets (with same architectures). \textit{Raw Data} indicates the amount of raw data used. \textit{Is CL} indicates whether it is a CL setting. \textit{Model Stream} indicates whether the model is learned from continuously encountered streams of pre-trained models. \textit{Black-box API} indicates whether the pre-trained model is accessed in a black-box manner (e.g., API).
The symbols \cmark \;/ \xmark \; are for yes / no. 
}
\vspace{-8pt}
\resizebox{1.0\textwidth}{!}{
\begin{tabular}{lcccccccccccc}
\toprule
\multirow{2}{*}{Method}   & \multicolumn{1}{c}{Raw} &\multicolumn{1}{c}{Is} &  \multicolumn{1}{c}{Model} & \multicolumn{1}{c}{Black-box} & \multicolumn{2}{c}{\revised{Split-}MNIST} & \multicolumn{2}{c}{\revised{Split-}SVHN} \\
& Data  & CL & Stream & APIs
& ACC(\%) $\uparrow$        & BWT(\%) $\uparrow$        
& ACC(\%) $\uparrow$        & BWT(\%) $\uparrow$      
 \\
\midrule
\multicolumn{1}{l}{Joint} 
& All     & \xmark   & \xmark  & \xmark 
&  99.60 $\pm$ 0.08     & N/A               
&  97.16 $\pm$ 0.26     & N/A              
 \\
\multicolumn{1}{l}{Sequential} 
 & All     & \cmark   & \xmark  & \xmark 
&  57.91 $\pm$ 8.46     & -52.17 $\pm$ 10.53             
&  62.62 $\pm$ 5.40     & -43.21 $\pm$ 06.90         \\
\multicolumn{1}{l}{Models-Avg} 
& N/A      & \xmark   & \xmark  & \xmark 
&  56.71 $\pm$ 4.97     & N/A              
&  52.29 $\pm$ 5.91     & N/A     
 \\
\multicolumn{1}{l}{Classic CL} 
& All     & \cmark   & \xmark  & \xmark  
&  99.48 $\pm$ 0.05     & -00.19 $\pm$ 00.06            
&  95.84 $\pm$ 0.43     & -02.35 $\pm$ 00.56     
\\
\multicolumn{1}{l}{Ex-Model}  
 & N/A      & \cmark   & \cmark  & \xmark 
&  98.73 $\pm$ 0.10    & -00.86 $\pm$ 00.53             
&  94.79 $\pm$ 0.58    & -02.43 $\pm$ 00.63          
 \\
\multicolumn{1}{l}{\textbf{DFCL-APIs(Ours)}}  
& N/A      & \cmark   & \cmark  & \cmark 
&  98.58 $\pm$ 0.41     & -00.13 $\pm$ 00.24            
&  94.29 $\pm$ 1.57     & -02.50 $\pm$ 01.45             
\\
\midrule
\multicolumn{1}{l}{Classic CL} 
& 2\%     & \cmark   & \xmark  & \xmark 
&  91.97 $\pm$ 3.75     &  $ $ 15.76 $\pm$ 05.21              
&  55.14 $\pm$ 1.26     &  $ $ 00.09 $\pm$ 04.44          
\\
\multicolumn{1}{l}{Classic CL} 
& 5\%     & \cmark   & \xmark  & \xmark 
&  97.66 $\pm$ 0.34     &  $ $ 04.50 $\pm$ 03.91             
&  80.65 $\pm$ 4.02     &  $ $ 09.20 $\pm$ 05.13             
\\
\multicolumn{1}{l}{Classic CL} 
& 10\%     & \cmark   & \xmark  & \xmark 
&  98.17 $\pm$ 0.25     & $ $ 00.19 $\pm$  00.16           
&  87.48 $\pm$ 2.00     & $ $ 04.42 $\pm$  04.11           
\\
\multicolumn{1}{l}{\textbf{DECL-APIs(Ours)}} 
& 2\%  & \cmark   & \cmark  & \cmark 
&  98.60 $\pm$ 0.78      & $ $ 00.89 $\pm$ 00.78            
&  96.01 $\pm$ 0.48      & -00.95 $\pm$ 00.45            
\\
\multicolumn{1}{l}{\textbf{DECL-APIs(Ours)}} 
& 5\%  & \cmark   & \cmark  & \cmark 
&  99.12 $\pm$ 0.16      & $ $ 00.37 $\pm$ 00.14              
&  96.84 $\pm$ 0.24      & -00.24 $\pm$ 00.29          
\\
\multicolumn{1}{l}{\textbf{DECL-APIs(Ours)}} 
& 10\%  & \cmark   & \cmark  & \cmark 
& 99.23 $\pm$ 0.09      & $ $ 00.39 $\pm$ 00.08            
& 97.28 $\pm$ 0.17      & $ $ 00.30 $\pm$ 00.04             
\\
\bottomrule
\end{tabular}
}
\label{tab:datafreeperformance_mnist_svhn}
\vspace{-10pt}
\end{table*}

\begin{table*}[t]
\small
\centering
\caption{  Performance comparison on \revised{Split-}CIFAR10, \revised{Split-}CIFAR100 and \revised{Split-}MiniImageNet datasets (with same architectures).
}
\vspace{-8pt}
\resizebox{1.0\textwidth}{!}{
\begin{tabular}{lcccccc}
\toprule
\multirow{2}{*}{Method} & \multicolumn{2}{c}{\revised{Split-}CIFAR10}   & \multicolumn{2}{c}{\revised{Split-}CIFAR100} & \multicolumn{2}{c}{\revised{Split-}MiniImageNet} \\
& ACC(\%) $\uparrow$         & BWT(\%) $\uparrow$    
& ACC(\%) $\uparrow$         & BWT(\%) $\uparrow$        
& ACC(\%) $\uparrow$         & BWT(\%) $\uparrow$         \\
\midrule
\multicolumn{1}{l}{Joint} 
&  93.71 $\pm$ 0.27     & N/A   
&  78.53 $\pm$ 0.46     & N/A    
&  83.08 $\pm$ 0.03     & N/A   
\\
\multicolumn{1}{l}{Sequential} 
&  57.98 $\pm$ 4.04     & -41.86 $\pm$ 04.68  
&  16.20 $\pm$ 0.89     & -59.93 $\pm$ 02.10     
&  14.82 $\pm$ 1.47     & $ $ -52.17 $\pm$ 01.99  
\\
\multicolumn{1}{l}{Models-Avg} 
&  50.02 $\pm$ 0.04     & N/A    
&  10.04 $\pm$ 0.22     & N/A             
&  09.92 $\pm$ 0.15     & N/A   
\\
\multicolumn{1}{l}{Classic CL} 
&  88.45 $\pm$ 1.20     & -05.29 $\pm$ 01.25 
&  68.83 $\pm$ 0.69     & -05.46 $\pm$ 01.13       
&  74.51 $\pm$ 0.64     & $ $ 02.43 $\pm$ 01.45  
\\
\multicolumn{1}{l}{Ex-Model}  
&  69.89 $\pm$ 0.39     & -14.68 $\pm$ 01.03 
&  36.98 $\pm$ 0.37     & -25.05 $\pm$ 01.98     
&  30.50 $\pm$ 4.38     & -13.48 $\pm$ 00.54    
\\
\multicolumn{1}{l}{\textbf{DFCL-APIs(Ours)}}  
&  70.63 $\pm$ 6.03     & -09.98 $\pm$ 00.62  
&  30.52 $\pm$ 1.51     & -21.41 $\pm$ 02.32      
&  22.95 $\pm$ 2.57     & -16.51 $\pm$ 02.82    
\\
\midrule
\multicolumn{1}{l}{Classic CL: 2\%} 
&  69.91 $\pm$ 3.27     &  $ $ 01.96 $\pm$ 05.34    
&  30.10 $\pm$ 0.72     & $ $ 03.49 $\pm$ 01.49      
&  20.74 $\pm$ 2.24     & $ $ 04.79 $\pm$ 00.71   
\\
\multicolumn{1}{l}{Classic CL: 5\%} 
&  74.62 $\pm$ 1.02     &     -00.72 $\pm$  01.36   
&  34.95 $\pm$ 1.59     & $ $ 00.29 $\pm$ 02.65        
&  31.56 $\pm$ 0.98     & $ $ 04.29 $\pm$ 01.22   
\\
\multicolumn{1}{l}{Classic CL: 10\%} 
&  77.84 $\pm$ 1.47     &    -01.22 $\pm$ 01.34
&  46.67 $\pm$ 3.14     & $ $ 04.92 $\pm$ 04.68     
&  34.18 $\pm$ 1.62     & -01.77 $\pm$ 01.53   
\\
\multicolumn{1}{l}{\textbf{DECL-APIs(Ours)}: 2\%} 
&  80.23 $\pm$ 3.51     & -04.39 $\pm$ 00.54    
&  40.05 $\pm$ 2.44     & -12.42 $\pm$ 02.10
&  38.01 $\pm$ 0.40     & $ $ 06.47 $\pm$ 04.08 
\\
\multicolumn{1}{l}{\textbf{DECL-APIs(Ours)}: 5\%} 
&  83.27 $\pm$ 1.30     & $ $ 01.94 $\pm$ 03.21    
&  48.03 $\pm$ 0.66     & -02.61 $\pm$ 02.10        
&  44.46 $\pm$ 2.14     & $ $ 12.86 $\pm$ 05.17   
\\
\multicolumn{1}{l}{\textbf{DECL-APIs(Ours)}: 10\%} 
&  86.20 $\pm$ 1.12     & $ $ 07.05 $\pm$ 04.72   
&  51.90 $\pm$ 2.06     & $ $ 07.08 $\pm$ 05.31 
&  51.49 $\pm$ 0.05     & $ $ 19.42 $\pm$ 01.57  
\\
\bottomrule
\end{tabular}
}
\label{tab:cifar10_cifar100_and_mniimagenet}
\vspace{-10pt}
\end{table*}

\begin{table*}[t]
\small
\centering
\revised{
\caption{Performance comparison on natural
language processing datasets. \textit{Raw Data} indicates the amount of raw data used. \textit{Is CL} indicates whether it is a continual learning setting. \textit{Model Stream} indicates whether the model is learned from continuously encountered streams of pre-trained models. \textit{Black-box API} indicates whether the pre-trained model is accessed in a black-box manner (e.g., API).
The symbols \cmark \;/ \xmark \; are for yes / no. MR and NS denote the two loss functions of memory replay and network similarity, respectively. N/A indicates not applicable.
}
\label{tab:datafreeperformance_nlpcl}
\vspace{-8pt}
\resizebox{1.0\textwidth}{!}{
\begin{tabular}{lcccc|ccccc|ccc}
\toprule
\multirow{2}{*}{Method}   & \multicolumn{1}{c}{Raw} &\multicolumn{1}{c}{Is} &  \multicolumn{1}{c}{Model} & \multicolumn{1}{c|}{Black-box} & \multicolumn{1}{c}{AGNews} & \multicolumn{1}{c}{Yelp} & \multicolumn{1}{c}{Amazon} & \multicolumn{1}{c}{Yahoo} & \multicolumn{1}{c|}{DBpedia} & \multicolumn{2}{c}{Average} &  \\
& Data  & CL & Stream & APIs
& ACC(\%) $\uparrow$    & ACC(\%) $\uparrow$     & ACC(\%) $\uparrow$  & ACC(\%) $\uparrow$     & ACC(\%) $\uparrow$   
& ACC(\%) $\uparrow$   & BWT(\%) $\uparrow$    
 \\
\midrule  
\multicolumn{1}{l}{Joint} 
& All     & \xmark   & \xmark  & \xmark 
& 90.67   &  59.13    &  55.68  &  71.37    &  98.56    
& 75.08   &  N/A          
 \\
\multicolumn{1}{l}{Sequential} 
 & All     & \cmark   & \xmark  & \xmark 
& 81.11    &  52.33   & 48.34  &  67.20    & 98.41   
& 69.48      &  -06.37    
 \\
\multicolumn{1}{l}{Models-Avg} 
& N/A      & \xmark   & \xmark  & \xmark 
&  78.77    &  50.31    &  45.44  & 49.77    & 56.38    
&  56.13   &  N/A     
 \\
\multicolumn{1}{l}{Classic CL}  
& All      & \cmark   & \xmark  & \xmark  
&  88.63   & 57.62    & 54.22  & 69.14    & 98.64    
& 73.65    &  -02.12    
\\
 \midrule
 \multicolumn{1}{l}{\textbf{DFCL-APIs (Ours)}}  
& N/A      & \cmark   & \cmark  & \cmark 
& 85.35    &  51.74    &  38.91  &  64.51    &  90.47    
& 66.20      &  -05.68     
\\
\multicolumn{1}{l}{\textbf{\;-w/o NS}}  
& N/A      & \cmark   & \cmark  & \cmark 
& 85.95    &  45.39   & 36.82  &  65.01   & 89.08    
& 64.45    &  -12.53    
\\
\multicolumn{1}{l}{\textbf{\;-w/o MR}}  
& N/A      & \cmark   & \cmark  & \cmark 
& 78.31    & 33.36    &  21.82  & 61.33    & 97.65    
& 58.49    & -14.00   
\\
\multicolumn{1}{l}{\textbf{\;-w/o MR and NS}}  
& N/A      & \cmark   & \cmark  & \cmark 
& 76.74    &  32.45   &  22.89  & 50.02    & 91.31    
& 54.68      & -17.19   
\\
\midrule
\multicolumn{1}{l}{{Classic CL}}  
& 2\%      & \cmark   & \xmark  & \xmark
&  86.42   &  18.02   &  19.23  &  10.01    &  06.11  & 27.96 & $ $ 00.00  
\\
\multicolumn{1}{l}{{Classic CL}}  
& 10\%      & \cmark   & \xmark  & \xmark
& 86.16    & 49.96    &  49.41  & 66.55    &  97.90  & 70.00      &  $ $ 11.27  
\\
\multicolumn{1}{l}{\textbf{DECL-APIs (Ours)}}  
& 2\%      & \cmark   & \cmark  & \cmark 
&  87.50    & 56.75    & 55.08  &  70.29    &  93.51  & 72.62      &  -00.73  
\\
\multicolumn{1}{l}{\textbf{DECL-APIs (Ours)}}  
& 10\%      & \cmark   & \cmark  & \cmark 
& 90.21    &  59.09   & 54.55  & 70.64    & 92.47  & 73.39      & $ $ 00.24  
\\
\bottomrule
\end{tabular}
}
}
\vspace{-10pt}
\end{table*}

\begin{table*}[t]
\small
\centering
\caption{Performance on \revised{Split-}CIFAR10 and \revised{Split-}CIFAR-100 datasets (with heterogeneous architectures). The symbols \cmark \;/ \xmark \; are for yes / no.
}
\vspace{-8pt}
\resizebox{1.0\textwidth}{!}{
\begin{tabular}{ c cc ccccccc}
\toprule
 Datasets      & Black-box APIs   & ResNet18  & ResNet34   & ResNet50    & WideResNet     & GoogleNet  & Method         & ACC(\%) $\uparrow$ & BWT(\%) $\uparrow$ \\ 
 \midrule
 \multirow{3}{*}{\revised{Split-}CIFAR10}            
& \xmark & $T1$  & $T2$  & $T3$  & $T4$      & $T5$         & Ex-Model            & 76.68 $\pm$ 3.53   & -12.46 $\pm$ 3.75         \\  
& \cmark  & $T1$  & $T2$  & $T3$  & $T4$      & $T5$         & DFCL-APIs(Ours)     & 72.51 $\pm$ 3.23     & -06.64 $\pm$ 2.24            \\  
& \cmark  & $T1$  & $T2$  & $T3$  & $T4$      & $T5$         & DECL-APIs(Ours)     & 86.42 $\pm$ 3.01    & $ $ 06.58 $\pm$ 1.18            \\  
 \midrule
 \multirow{3}{*}{\revised{Split-}CIFAR100}        
& \xmark & $T1$ \& $T2$    & $T3$ \& $T4$     & $T5$ \& $T6$     & $T7$ \& $T8$     & $T9$ \& $T10$   & Ex-Model               & 36.57 $\pm$ 2.55      & -15.85 $\pm$ 2.56       \\  
& \cmark  & $T1$ \& $T2$    & $T3$ \& $T4$     & $T5$ \& $T6$     & $T7$ \& $T8$     & $T9$ \& $T10$   & DFCL-APIs(Ours)       & 23.48 $\pm$ 1.24      & -16.73 $\pm$ 2.33      \\ 
& \cmark  & $T1$ \& $T2$    & $T3$ \& $T4$     & $T5$ \& $T6$     & $T7$ \& $T8$     & $T9$ \& $T10$   & DECL-APIs(Ours)       & 44.16 $\pm$ 2.16        & $ $ 12.81 $\pm$ 2.26      \\  
\bottomrule
\end{tabular}
}
\label{tab:genaralization_dif}
\vspace{-12pt}
\end{table*}

\noindent
\textbf{Baselines.} 
We compare the following baselines, and their details are shown in Tab.~\ref{tab:datafreeperformance_mnist_svhn}. The architecture and implementation details are in Appendix~\ref{sec:ExperimentDetails_Appendix}. The baselines are introduced as follows:
\begin{itemize}[noitemsep,topsep=0pt,parsep=0pt,partopsep=0pt]
    \item \textbf{Joint} is a joint training method for all tasks on raw data. Since it does not involve the problem of forgetting, it is usually regarded as the upper bound of classic CL.
    \item \textbf{Sequential} represents the tasks encountered sequentially using the SGD optimizer to learn on the raw data stream. Since it does not have any strategies to alleviate forgetting, it is usually regarded as the lower bound for classic CL.
    \item \textbf{Models-Avg} directly averages the weights of all pre-trained models, requiring all APIs to have the same architecture and access to model parameters. 
    \item \textbf{Classic CL} is the traditional CL method, which learns directly from the raw data (in Fig.~\ref{fig:scenario}(a)). Without loss of generality, we exploit the classical replay-based approach (ER~\cite{er}).
    \item \textbf{Ex-Model}~\cite{ExModel2022} is the CL from a stream of pre-trained \textit{white-box} models, and it can be considered as the \textit{upper bound} of our black-box API settings. {For a fair comparison, we slightly modify the original implementation. In this paper, we invert the entire white-box model and adopt a memory replay strategy similar to Classic CL to mitigate forgetting.}
    \item \textbf{DECL-APIs (Ours)} and \textbf{DFCL-APIs (Ours)} are our proposed new settings for CL from a stream of \textit{black-box} APIs with little or no raw data (as shown in Fig.~\ref{fig:scenario}(b)). We utilize the proposed data-free cooperative continual distillation learning framework to perform CL in both settings.
\end{itemize}

\noindent
\textbf{Evaluation Metrics.}
We evaluate the performance of CL by measuring the average accuracy (ACC) and backward transfer (BWT). ACC measures the average accuracy of the CL model across all tasks, while BWT measures the degree of forgetting of the model. \revised{The accuracy of individual tasks is also shown in the Appendix.}
ACC and BWT are defined as follows:
$$
\small
    A C C \!=\!\frac{1}{K} \sum_{i=1}^{K} A_{K, i}, 
    \;\; 
    B W T \!=\!\frac{1}{K-1} \sum_{i=1}^{K-1} A_{K, i}\!-\!A_{i, i},
$$
where $\small A_{k,i}$ is the accuracy of the model tested on the task $\small i$ after the training of task $\small k$ is completed. $\small K$ denotes the number of tasks.

\subsection{Experiments on the Same Architecture}
\label{Sec:ExperimentalResults_Same}
This section assumes that the API for all tasks in each task and the CL model use the same architecture. \revised{We present results in the CV and NLP domains in Sec.~\ref{subsubsec:cv} and Sec.~\ref{subsubsec:nlp}, respectively.}

\subsubsection{Computer Vision (CV) Domain}
\label{subsubsec:cv}
\revised{In this section, we demonstrate the feasibility of performing CL from a stream of black-box APIs in the CV domain.}

\noindent
\textbf{Implementation}.
For the \revised{Split-}MNIST dataset, we use the LeNet5~\cite{mnist} architecture, and for the \revised{Split-}SVHN, \revised{Split-}CIFAR10, \revised{Split-}CIFAR100 and \revised{Split-}MiniImageNet datasets, we use the ResNet18~\cite{resnet} architecture. The network architectures of the generator, APIs, and CL model are defined in Appendix~\ref{sec:ArchitectureDetails_Appendix}. In addition, due to page limitations, detailed accuracies of all compared methods on the five datasets are provided in Appendix~\ref{sec:moreanalysis_appendix}.

\noindent
\textbf{Performance in the DFCL-APIs setting}.
As shown in upper part of Tab.~\ref{tab:datafreeperformance_mnist_svhn} and Tab.~\ref{tab:cifar10_cifar100_and_mniimagenet}, we have the following observations:
(i) The Joint and Classic CL methods achieve the best performance as they are learned from all raw data. Among them, the Joint is superior to Classic CL because it assumes that all task data are available simultaneously, avoiding the problem of catastrophic forgetting.
(ii) Sequential and Models-Avg are the worst-performing baseline methods. Sequential lacks a mechanism to prevent catastrophic forgetting. Models-Avg averages the model parameters trained on tasks with different data distributions, leading to a drift between the averaged parameters and the optimal parameters of each task, resulting in poor performance.
(iii) Ex-Model is a more practical approach that learns from a stream of pre-trained models without raw data, which achieved $\small 0.99\times$, $\small 0.98\times$, $\small 0.79\times$, $\small 0.53\times$ and $\small 0.40\times$ performance on MNIST, SVHN, CIFAR10, CIFAR100, MiniImageNet datasets, respectively, compared to Classic CL.
(iv) Our method achieves comparable performance to the upper bound Ex-Model on MNIST, SVHN, and CIFAR10 datasets in DFCL-API setting. This is because our method generates more hard and diverse samples, and catastrophic forgetting is effectively alleviated. However, our approach is slightly inferior to Ex-Model on the more challenging CIFAR100 and MiniImageNet datasets, where both the number of tasks and classes significantly increased.
This is due to the fact that Ex-Model had access to model parameters, whereas we could only use black-box API, so our settings are significantly more difficult.

\noindent
\textbf{Performance in the DECL-APIs setting}.
As shown in the lower part of Tab.~\ref{tab:datafreeperformance_mnist_svhn} and Tab.~\ref{tab:cifar10_cifar100_and_mniimagenet}, when a small amount of raw data (e.g., $\small 2\%, 5\%, 10\%$) are available, we have the following observations:
(i) Classic CL performs poorly when there is only a small amount of raw task data. For example, considering the SVHN, CIFAR10, and CIFAR100 datasets and only having access to $\small 2\%$ of their raw data, the Classic CL model can only achieve $\small 0.57\times$, $\small 0.79\times$, and $\small 0.43\times$ performance of the full data, respectively.
(ii) When our method accesses only $\small 10\%$ of the data in the DECL-APIs setting, its performance far exceeds that of the DFCL-APIs setting. It achieves $\small 1.0\times$, $\small 0.97\times$, and $\small 0.75\times$ times the performance of classic CL on the three datasets, SVHN, CIFAR10, and CIFAR100, respectively.

\revised{
\subsubsection{Natural Language Processing (NLP) Domain}
\label{subsubsec:nlp}
\revised{This section demonstrates the possibility of performing CL from a stream of black-box APIs in the NLP domain.}

\noindent
\textbf{Implementation}. For CL in the NLP domain, we follow the setting of \cite{de2019episodic,huang2021continual} to construct the CL task consisting of five text datasets. In the setting of this paper, each API is a BERT~\cite{bert} model, and the CL model is also a BERT model. For the generator, we adopt the GPT-2~\cite{gpt2} model to generate a series of pseudo-text data. Since the parameters of the GPT-2 model are large and the generation capability is strong enough, in this section, which is slightly different from the CV setting in Sec.~\ref{subsubsec:cv}, we will not further update the generator, but directly utilize the existing one.

\begin{table*}[t]
\small
\revised{
\caption{Performance comparison on \revised{Split-}MNIST and \revised{Split-}SVHN datasets (with same architectures) in the \textbf{`hard label' setting}.  \textit{Hard Label} means whether the API can only return the predicted value in one-hot form, without the predicted probability for each class.
}
\label{tab:datafreeperformance_mnist_svhn_hl}
\centering
\vspace{-8pt}
\resizebox{1.0\textwidth}{!}{
\begin{tabular}{lcccccccccccc}
\toprule
\multirow{2}{*}{Method}   & \multicolumn{1}{c}{Raw} &\multicolumn{1}{c}{Is} &  \multicolumn{1}{c}{Model} & \multicolumn{1}{c}{Black-box} & Hard  & \multicolumn{2}{c}{\revised{Split-}MNIST} & \multicolumn{2}{c}{\revised{Split-}SVHN}\\
& Data  & CL & Stream & APIs & Label 
& ACC(\%) $\uparrow$        & BWT(\%) $\uparrow$        
& ACC(\%) $\uparrow$        & BWT(\%) $\uparrow$           
 \\
 \midrule
\multicolumn{1}{l}{\textbf{DFCL-APIs}}  
& N/A      & \cmark   & \cmark  & \cmark  & \xmark 
&  98.58 $\pm$ 0.41     & -00.13 $\pm$ 00.24            
&  94.29 $\pm$ 1.57     & -02.50 $\pm$ 01.45             
\\
\multicolumn{1}{l}{\textbf{DFCL-APIs}}  
& N/A      & \cmark   & \cmark  & \cmark  & \cmark 
& 95.29 $\pm$ 2.53 &  -00.47 $\pm$ 00.42
& 91.66 $\pm$ 2.74 &  -03.90 $\pm$ 00.78
\\
\midrule
\multicolumn{1}{l}{\textbf{DECL-APIs}} 
& 2\%  & \cmark   & \cmark  & \cmark & \xmark 
&  98.60 $\pm$ 0.78      & $ $ 00.89 $\pm$ 00.78            
&  96.01 $\pm$ 0.48      & -00.95 $\pm$ 00.45    
\\
\multicolumn{1}{l}{\textbf{DECL-APIs}} 
& 2\%      & \cmark   & \cmark  & \cmark  & \cmark  
& 97.65 $\pm$ 0.92  &  $ $ 00.47 $\pm$ 00.19 
& 94.87 $\pm$ 0.30 &   -01.68 $\pm$ 00.26  
\\ 
\bottomrule
\end{tabular}
}
}
\vspace{-10pt}
\end{table*}

\begin{figure*}[t]
\centering
\includegraphics[width=.33\textwidth]{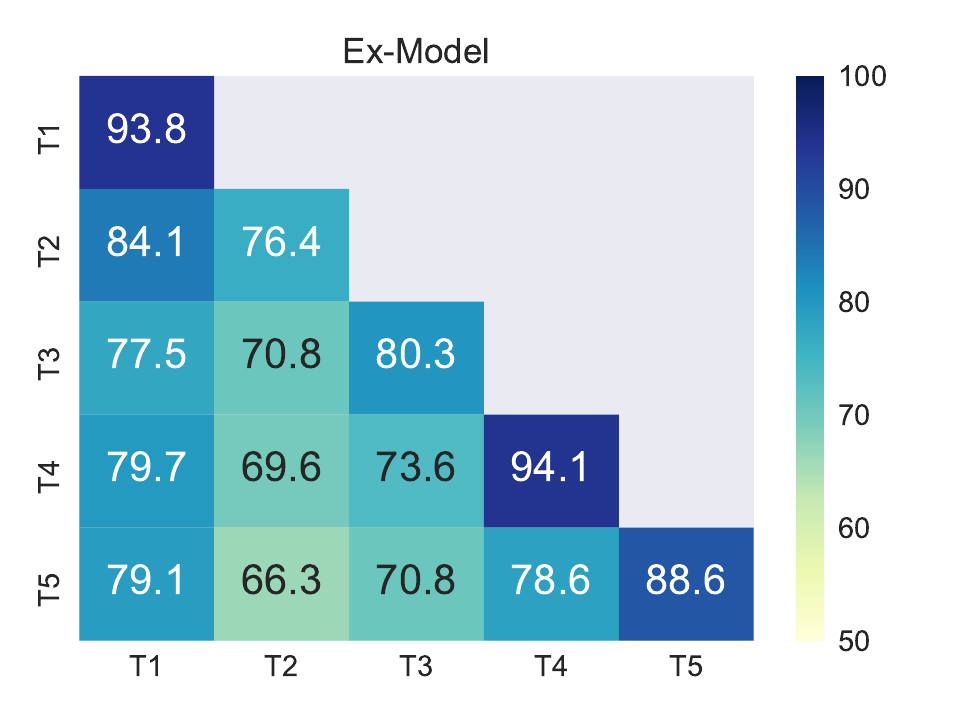}
\vspace{-8pt}
\includegraphics[width=.33\textwidth]{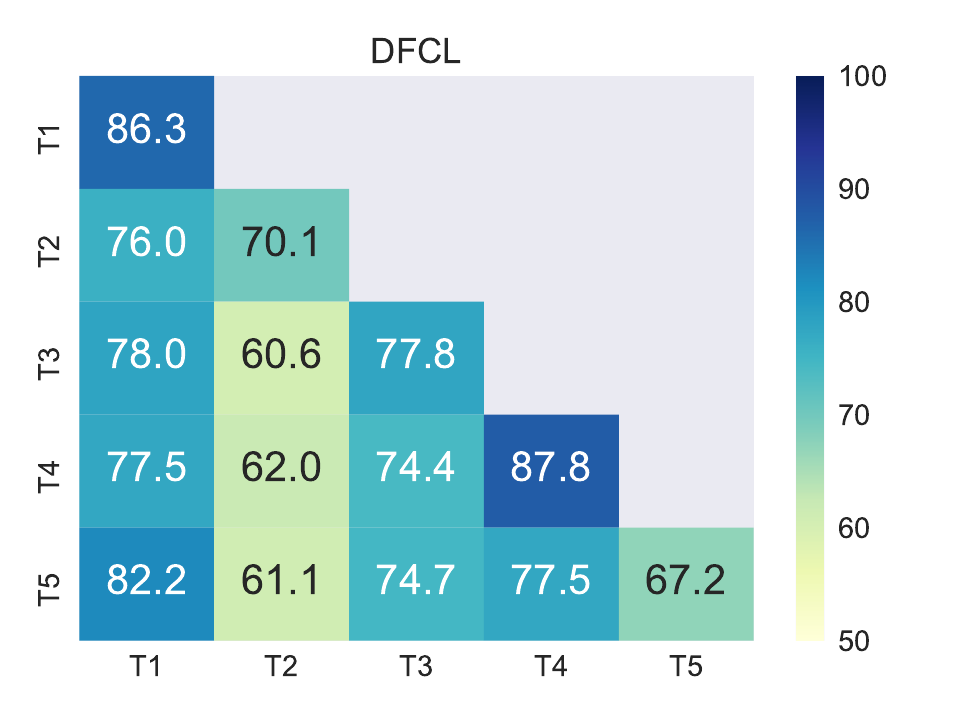}
\vspace{-8pt}
\includegraphics[width=.33\textwidth]{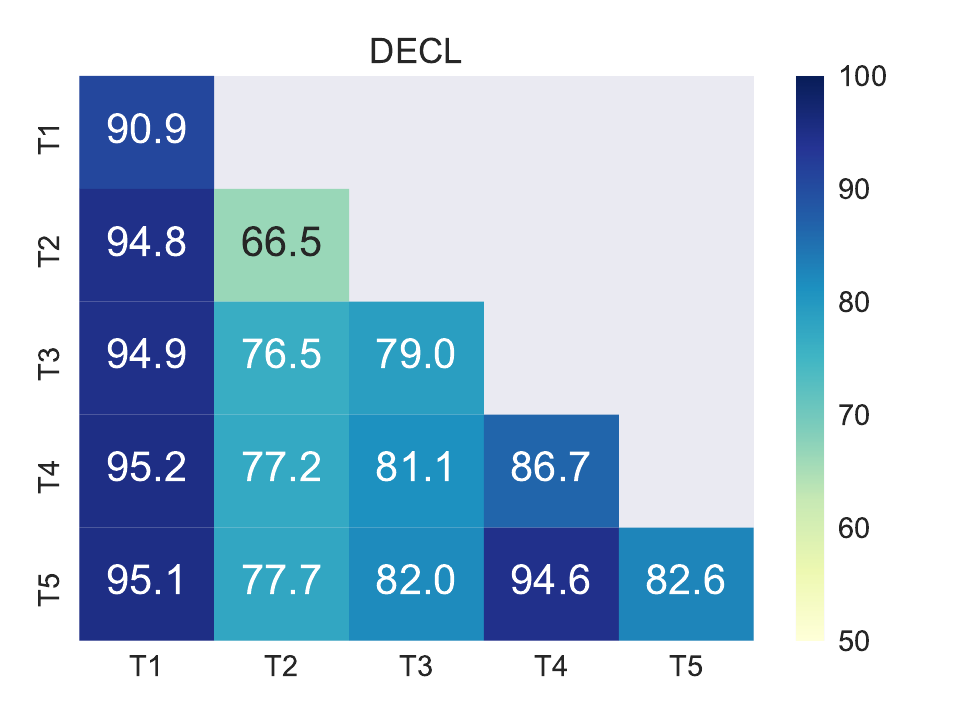}
\vspace{-8pt}
\caption{The accuracy (Higher Better) on the \revised{Split-}{CIFAR10} dataset (with heterogeneous architectures). (1) Ex-Model, (2) DFCL-APIs(ours), (3) DECL-APIs(ours). $t$-th row represents the accuracy of the network tested on tasks $1-t$ after task $t$ is learned.
}
\vspace{-10pt}
\label{fig:acc_cifar10_heterogeneous}
\end{figure*}

\begin{figure*}[t]
\centering
\includegraphics[width=.33\textwidth]{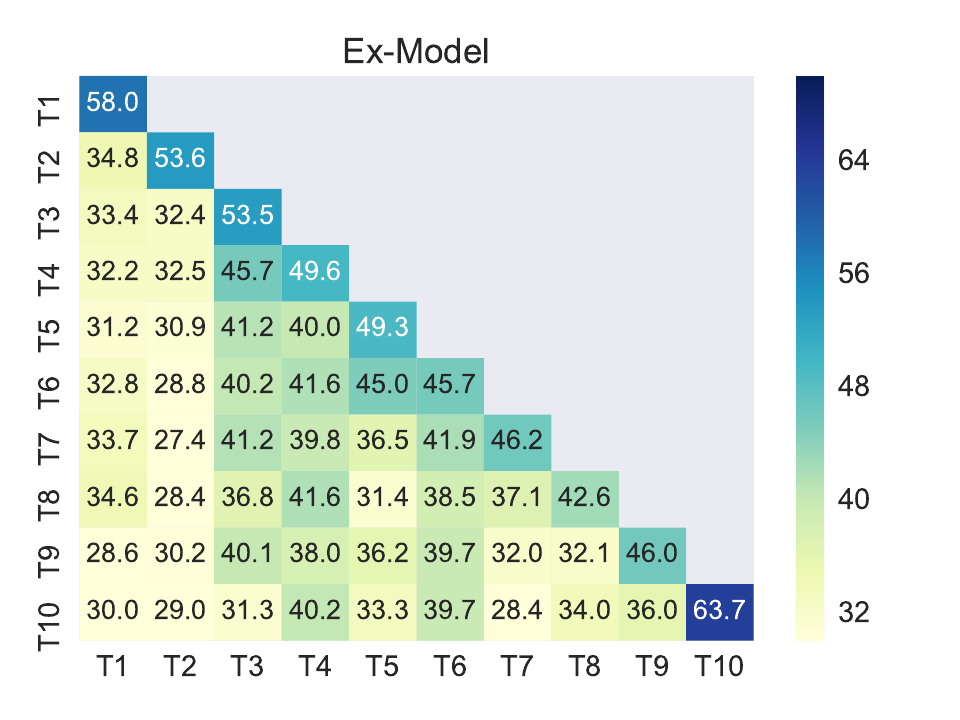}
\vspace{-8pt}
\includegraphics[width=.33\textwidth]{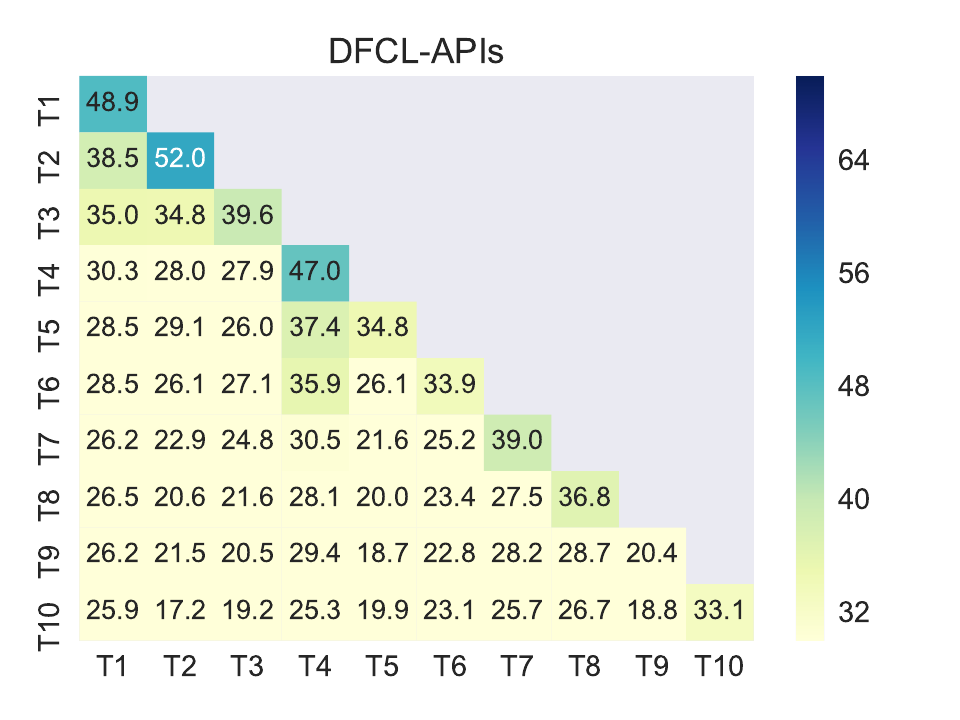}
\vspace{-8pt}
\includegraphics[width=.33\textwidth]{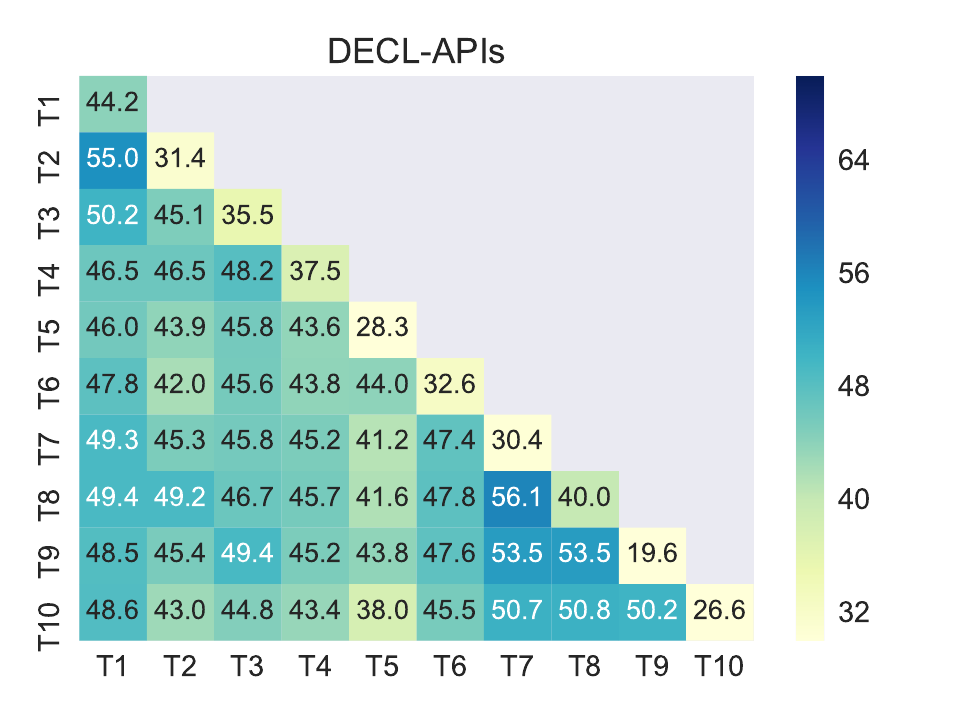}
\vspace{-8pt}
\caption{The accuracy (Higher Better) on the \revised{Split-}{CIFAR100} dataset (with heterogeneous architectures). (1) Ex-Model, (2) DFCL-APIs(ours), (3) DECL-APIs(ours). $t$-th row represents the accuracy of the network tested on tasks $1-t$ after task $t$ is learned.
}
\vspace{-10pt}
\label{fig:acc_cifar100_heterogeneous}
\end{figure*}

\noindent
\textbf{Performance in the DFCL-APIs setting}. 
As shown in Tab.~\ref{tab:datafreeperformance_nlpcl}, we have the following observations:
(i) The joint approach achieves optimal performance because it trains all task data together and therefore has no forgetting problem. On the contrary, Sequential's approach shows a significant performance degradation because it has no strategy to mitigate forgetting. Classic CL performs in between, as it takes the means of old task's data replay to mitigate forgetting.
(ii) Directly merging model parameters leads to severe performance degradation, i.e., Models-Avg only has an average accuracy of 56.13\%.
(iii) Our DFCL-APIs performs CL from the APIs corresponding to the five datasets without access to arbitrary raw training data and achieves an average accuracy of 66.20\%.  In particular, our proposed two losses, memory replay (MR) and network similarity (NS), both play a prominent role in alleviating forgetting. For example, when MR And NS are not used, the average accuracy drops to 58.49\% and 64.45\%, respectively. Without both, the average accuracy drops to 54.68\%.

\noindent
\textbf{Performance in the DECL-APIs setting}.
The results of DECL-APIs are shown at the bottom of Tab.~\ref{tab:datafreeperformance_nlpcl}. We make the following observations: (i) When only a very small amount of raw data is available, Classic CL performs poorly because it cannot be fully trained. For example, when only 2\% of the original data is accessible, the average accuracy of Classic CL is only 27.96\%.
(ii) When a small amount of data is available, DECL-APIs shows a clear performance improvement over DFCL-APIs, e.g., when 2\% of the data is available, DECL-APIs has an average accuracy of 72.62\%, which is very close to Classic CL with full raw data available, which has an average accuracy of 73.65\%.
}

\subsection{Experiments on Heterogeneous Architectures}
\label{Sec:ExperimentalResults_Het}
As the architecture of APIs for different tasks may vary in practical applications, we tested only Ex-Model and our two settings, which are learned from pre-trained models, on \revised{Split-}CIFAR10 and \revised{Split-}CIFAR100 datasets. 

\noindent
\textbf{Implementation}. 
As shown in Tab.~\ref{tab:genaralization_dif}, we use architectures with different depths (ResNet18$\rightarrow$ResNet34$\rightarrow$ResNet50), widths (ResNet$\rightarrow$WideResNet), and types (ResNet$\rightarrow$GoogleNet) for different APIs. We then learn from these APIs (For example, in Tab.~\ref{tab:genaralization_dif}, CIFAR10 uses ResNet18 for task $T1$ and ResNet34 for task $T2$.) sequentially encountered under the three settings of Ex-Model, DFCL-APIs, and DECL-APIs, and we use ResNet18 as the architecture of the CL model. 
In the DECL-APIs setting, we assume that $\small 10\%$ of the raw data is available.

\noindent
\textbf{Performance}. 
Tab.~\ref{tab:genaralization_dif} shows the experimental results in the heterogeneous network. In addition, to observe the change in accuracy and the degree of forgetting of each task in more detail during the training process, we provide the task accuracy of each stage on the CIFAR10 and CIFAR100 datasets in Fig.~\ref{fig:acc_cifar10_heterogeneous} and Fig.~\ref{fig:acc_cifar100_heterogeneous}, respectively.
We have the following observations: (i) On the CIFAR10 dataset, our method achieves performance close to that of Ex-Model under the DFCL-APIs setting, with the former being $\small 72.51\%$ and the latter being $\small 76.68\%$. Under the DECL-APIs setting, our method achieves a performance of $\small 86.42\%$. (ii) On the more challenging CIFAR100 dataset, there is a certain gap between DFCL-APIs and Ex-Model. This is because Ex-Model uses the pre-trained in a white-box manner, while DFCL-APIs access the black-box API. In DECL-APIs, our method has significantly improved performance. These results demonstrate that our method can also work with a stream of heterogeneous black-box APIs.

\subsection{\revised{Experiments on Hard Label Setting}}
\label{subsec:hardlabel}
\revised{
In this section, we validate the proposed approach in a more challenging scenario, where the black-box API only returns the predicted class (i.e., hard label) without providing the API's logits (i.e., soft label).
The experimental results are shown in Tab.~\ref{tab:datafreeperformance_mnist_svhn_hl}, showing that our approach still works in this more challenging setting. Specifically, on the Split-MNIST dataset, DFCL-APIs with soft labels achieve an accuracy of $\small 98.58\%$, while they still achieve an accuracy of $\small 95.29\%$ with hard labels. In DECL-APIs with a small amount of raw data, the performance of the hard label is closer to that of the soft label, with $\small 97.65\%$ compared to $\small 98.60\%$.
These results further enhance the possibility presented in this paper to perform continual learning from a stream of black-box APIs.
}

\begin{figure*}[h]
\centering
\includegraphics[width=1.\textwidth]{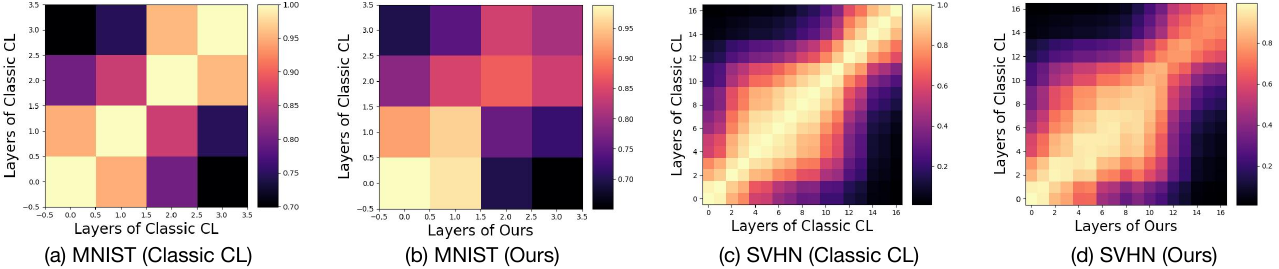}
\vspace*{-18pt} 
\caption{Layer-by-layer comparison of similarity on the trained network for the \revised{Split-}MNIST and \revised{Split-}SVHN.
}
\label{fig:cka_mnist_svhn}
\vspace{-10pt}
\end{figure*}

\begin{figure*}[h]
\centering
\includegraphics[width=1.\textwidth]{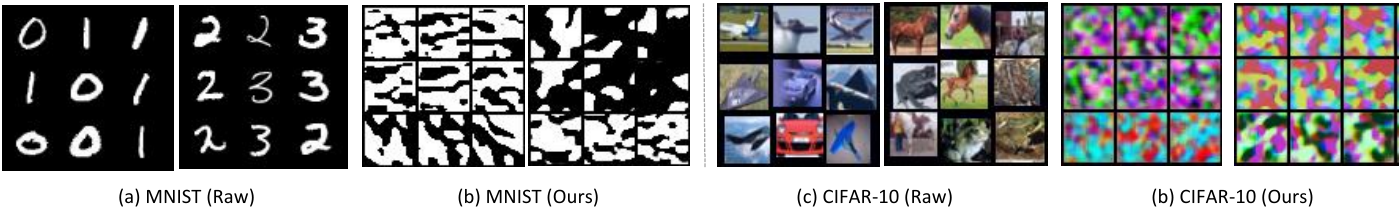}
\vspace*{-18pt} 
\caption{Visualizations of real and synthetic data on the \revised{Split-}MNIST (a-b) and \revised{Split-}CIFAR-10  (c-d) datasets.
}
\label{fig:visualization}
\vspace{-10pt}
\end{figure*}

\subsection{Analysis}
\label{sec:ablationstudy}

\noindent
\textbf{Network Similarity Analysis.}
The performance in Sec.~\ref{Sec:ExperimentalResults_Same} already illustrates the possibility of the proposed framework for CL without raw data. We further verify beyond accuracy on MNIST and SVHN datasets. We use the method of layer-by-layer similarity analysis (e.g., centered kernel alignment~\cite{cka_nips2021}) to observe whether the CL model trained by the proposed method without using raw data is highly similar to the CL model trained on raw data. As shown in Fig.~\ref{fig:cka_mnist_svhn}, in the subfigures (a) and (c), we compare the similarity between two layers of the CL model trained on raw data and use it as the ground truth. In subfigures (b) and (d), we compare the similarity of the CL model trained on raw data to the CL model trained using pseudo data in the DFCL-APIs setting. We can observe that the values on the diagonal are significantly larger, i.e., the CL model trained on the DFCL-APIs setting is similar to the CL model trained on raw data.

\noindent
\textbf{Generated Samples Visualization.}
As shown in Fig.~\ref{fig:visualization}, we provide examples of real and generated images from the MNIST and CIFAR10 datasets, respectively. Our primary goal in training generators in Sec.~\ref{sec:trainning_generative} is to create "hard" samples, rather than human-recognizable image details, so we cannot distinguish the corresponding classes of images. However, as we can see from the performance analysis, the generated images capture the main classification features of the raw task.

\section{Conclusion {and Future Work}}
\label{sec:conclusion}
In this paper, we define two new CL settings: data-free continual learning (DFCL-APIs) and data-efficient continual learning (DECL), which executes CL from a stream of APIs with little or no raw data, respectively. To address the challenges in these two new settings, we propose a data-free cooperative continual learning framework that distills the knowledge of the black-box APIs into the CL model. The proposed method (i) does not require full raw data and avoids data privacy problems to some extent, (ii) does not require model architecture and parameters, (iii) only calls API and can adapt to API with any scale, and (iv) effectively alleviates catastrophic forgetting.
Experimental results show that it is possible to conduct CL from a stream of APIs in computer vision and natural language processing domains, which greatly expands the application of CL.
In the future, we intend to further explore performing CL from a stream of APIs with limited query cost.

\IEEEdisplaynontitleabstractindextext

\IEEEpeerreviewmaketitle

{
\bibliographystyle{IEEEtran}
\bibliography{main}
}

\ifCLASSOPTIONcaptionsoff
  \newpage
\fi

\appendices
\clearpage

\revised{
\noindent
\textbf{Appendix Overview}. In the appendix, we mainly include the following content that cannot be included due to the main page limit.
\begin{itemize}[noitemsep,topsep=0pt,parsep=0pt,partopsep=0pt]
    \item We provide more experimental details in Appendix \ref{sec:ExperimentDetails_Appendix}, including dataset statistics, network architecture, implementation details, and algorithm details.
    \item We provide more experimental analysis in Appendix \ref{sec:moreanalysis_appendix}, including ablation study, query cost, hyperparameter analysis, and accuracy visualization.
\end{itemize}
}

\section{Experiment Details}
\label{sec:ExperimentDetails_Appendix}

\revised{This section describes the experimental data, architectures, and implementation details in detail.}

\noindent
\subsection{Dataset Statistics}
\label{sec:DatasetStatistics_Appendix}
We evaluate our method on \revised{Split-}MNIST\footnote{\footnotesize \revised{In this paper, we do not specifically distinguish between the two terms `Split-MNIST' and `MNIST', they both mean the same thing, that is, building a continual learning setting on the split MNIST dataset. Other computer vision datasets are similar.}}~\cite{mnist}, \revised{Split-}SVHN~\cite{svhn}, \revised{Split-}CIFAR10~\cite{cafar10_cifar100}, \revised{Split-}CIFAR100~\cite{cafar10_cifar100} and \revised{Split-}MiniImageNet~\cite{miniimagenet} datasets \revised{in the computer vision domain}. The dataset statistics are shown in Tab.~\ref{tab:datasetstatic}. Specifically, the column "\#Classes" indicates how many classes are in each dataset. The column "\#Tasks" indicates the number of continuously arriving tasks in the corresponding dataset. The columns "\#Train", "\#Valid", and "\#Test" represent the average number of training, validation, and test samples in each task, respectively. The column "\#Image Size" represents the resolution of each image in the raw dataset.
\begin{table}[h]
\centering
\caption{Statistics of the datasets in the computer vision domain.}
\vspace{-10pt}
\label{tab:datasetstatic}
\resizebox{.5\textwidth}{!}{
\begin{tabular}{l  ccrrrrrr}
\toprule
Datasets     & \#Classes    & \#Tasks& \#Train & \#Valid & \#Test & \#Image Size \\ \midrule
\revised{Split-}MNIST        & 10    & 5           & 10,800      & 1,200      &  2,000    &  1 $\times$ 28 $\times$ 28  \\ 
\revised{Split-}SVHN         & 10     & 5          & 13,919      & 732        &  5,206    &  3 $\times$ 32 $\times$ 32  \\  
\revised{Split-}CIFAR10      & 10   & 5          & 9,500       & 500        &  2,000    &  3 $\times$ 32 $\times$ 32 \\  
\revised{Split-}CIFAR100     & 100   & 10           & 4,750       & 250        &  1,000    &  3 $\times$ 32 $\times$ 32  \\ 
\revised{Split-}MiniImageNet & 100  & 10           & 4,500       & 500        &  1,000      &  3 $\times$ 64 $\times$ 64  \\
\bottomrule
\end{tabular}
}
\end{table}

\revised{
Following the setting of \cite{de2019episodic,huang2021continual}, we construct a CL task in the natural language processing domain using five text datasets AGNews, Yelp, Amazon, DBPedia, and Yahoo. These datasets are from \cite{zhang2015character}, and the statistics are in Tab.~\ref{tab:datasetstatic_nlp}.
}
\begin{table}[h]
\centering
\revised{
\caption{Statistics of the datasets in the natural language processing domain.}
\vspace{-10pt}
\label{tab:datasetstatic_nlp}
\resizebox{.5\textwidth}{!}{
\begin{tabular}{l  ccrrrrrr}
\toprule
Datasets     & Type  & \#Train & \#Train & \#Valid & \#Test  \\ \midrule
AGNews        & News Classification  & 4     & 8,000     & 8,000  &  7,600  \\ 
Yelp        & Sentiment Analysis  & 5     &  1,0000     & 1,0000  &  7,600  \\ 
Amazon        & Sentiment Analysis  & 5     & 1,0000     & 1,0000   &  7,600  \\ 
DBPedia        & Wikipedia Article Classification  & 14     & 28,000    & 28,000  &  7,600  \\ 
Yahoo        & Questions and Answers Categorization  & 10     &  20,000     & 20,000  &  7,600  \\ 
\bottomrule
\end{tabular}
}
}
\vspace{-10pt}
\end{table}

\noindent
\subsection{Architecture Details}
\label{sec:ArchitectureDetails_Appendix}
\revised{In the computer vision domain,} we refer to the generators and commonly used network architectures defined by model extraction~\cite{DataFreeModelExtraction2021CVPR,dfad_2019,maze_cvpr2021}. The architectures of generators and APIs are defined as follows:
(1) \textbf{Generators}:
In all datasets and experiments, we use the network architecture shown in Tab.~\ref{tab:generators} as generators.
(2) \textbf{APIs and CL Model}:
In the same architecture experiments in Sec.~\ref{Sec:ExperimentalResults_Same}, the APIs and CL model of the MNIST dataset use the LeNet architecture~\cite{lenet}, and the APIs and CL model on the remaining four datasets use the ResNet18 architecture~\cite{resnet}.
In the heterogeneous architecture experiments in Sec.~\ref{Sec:ExperimentalResults_Het}, the CL model uses the ResNet18 architecture, and different APIs use different network architectures (in Tab.~\ref{tab:genaralization_dif}). The architectures involved are ResNet18~\cite{resnet}, ResNet34~\cite{resnet}, ResNet50~\cite{resnet}, WideResNet~\cite{wideresnet}, and GoogleNet~\cite{googlenet}.

\begin{table}[h]
  \centering
  \vspace{-10pt}
  \caption{Generator Architecture. }
  \vspace{-8pt}
\resizebox{0.4\textwidth}{!}{
  \begin{tabular}{c|c}
  \hline
  \bf  \bf Layer/Block & \bf Configuration  \\
  \hline
  Input     & img\_size, latent\_dim, channels \\
   \hline
  FC Layer  &  Linear(latent\_dim, 128 $\times$ (img\_size // 4)$^2$)\\
   \hline
  Block 1  & BatchNorm2d(128)\\
  \hline
  Block 2 & Conv2d(128, 128, 3, stride=1, padding=1) \\
          & BatchNorm2d(128, 0.8) \\
          & LeakyReLU(0.2, inplace=True) \\
  \hline
  Block 3 & Conv2d(128, 64, 3, stride=1, padding=1) \\
          & BatchNorm2d(64, 0.8) \\
          & LeakyReLU(0.2, inplace=True) \\
          & Conv2d(64, channels, 3, stride=1, padding=1) \\
          & Tanh() \\
          & BatchNorm2d(channels, affine=False) \\
  \hline
  Output  & channels $\times$  img\_size $\times$  img\_size \\
  \hline
  \end{tabular}
  }
\label{tab:generators}
\vspace{-8pt}
\end{table}

\begin{table}[h]
\footnotesize
\centering
\caption{The hyperparameters for baselines \revised{in the computer vision domain}. The ‘MINI’ denotes the MiniImagenet dataset. 
}
\vspace{-10pt}
\label{tab:hyper}
\resizebox{0.5\textwidth}{!}{
\begin{tabular}{@{}l ll@{}}
\toprule
\textbf{Methods}&   \textbf{Hyperparameters} \\
\midrule
Joint \text{  }              
& \text{} Learning rate: 0.01 (MNIST, SVHN, CIFAR10, CIFAR100, MINI); \\
& \text{} Epoch: 10 (MNIST), 100 (SVHN, CIFAR10, MINI), 200 (CIFAR100);\\ 
& \text{} Batch size: 64 (MNIST, CIFAR100, MINI), 128 (SVHN, CIFAR) \\ \midrule
Sequential \text{  }         
& \text{} Learning rate: 0.01 (MNIST, SVHN, CIFAR10, CIFAR100, MINI); \\
& \text{} Epoch: 10 (MNIST), 100 (SVHN, CIFAR10, MINI), 200 (CIFAR100);\\ 
& \text{} Batch size: 64 (MNIST, CIFAR100, MINI), 128 (SVHN, CIFAR10) \\ \midrule
Classic CL  \text{  }  
& \text{} Learning rate: 0.01 (MNIST, SVHN, CIFAR10, CIFAR100, MINI); \\
& \text{} Epoch: 10 (MNIST), 100 (SVHN, CIFAR10, MINI), 200 (CIFAR100);\\ 
& \text{} Batch size: 64 (MNIST, CIFAR100, MINI), 128 (SVHN, CIFAR10);\\
& \text{} Maximum memory buffer size: 2500 (MNIST, SVHN, CIFAR10); 5000 (CIFAR100, MINI);\\ 
\midrule
Ex-Model  \text{  }   \text{  }  
& \text{} Learning rate: 0.01 (MNIST, SVHN, CIFAR10, CIFAR100, MINI); \\
& \text{} Epoch: 10 (MNIST), 30 (SVHN), 50 (CIFAR10, CIFAR100, MINI); \\
& \text{} Inner iteration: 50 (MNIST, SVHN, CIFAR10, CIFAR100, MINI); \\
& \text{} Batch size: 32 (MNIST, SVHN, CIFAR10, CIFAR100, MINI); \\
& \text{} Maximum memory buffer size: 2500 (MNIST, SVHN, CIFAR10); 5000 (CIFAR100, MINI);\\
& \text{} Generator step: 1 (MNIST, SVHN, CIFAR10, CIFAR100, MINI); \\
& \text{} CL step: 4 (MNIST, SVHN, CIFAR10, CIFAR100, MINI); \\
& \text{} Generators learning rate: 0.001  (MNIST, SVHN, CIFAR10, CIFAR100, MINI); \\
& \text{} Dimensions of random noise: 100 (MNIST), 256 (SVHN, CIFAR10, CIFAR100, MINI)\\ \midrule
DFCL-APIs  \text{  }    
& \text{} Learning rate: 0.01 (MNIST, SVHN, CIFAR10, CIFAR100, MINI); \\
\&
& \text{} Epoch: 10  (MNIST), 30 (SVHN), 50 (CIFAR10, CIFAR100, MINI); \\
DECL-APIs    \text{  } 
& \text{} Inner iteration: 50 (MNIST, SVHN, CIFAR10, CIFAR100, MINI);\\
& \text{} Batch size: 32 (MNIST, SVHN, CIFAR10, CIFAR100, MINI);  \\
& \text{} Maximum memory buffer size: 2500 (MNIST, SVHN, CIFAR10); 5000 (CIFAR100, MINI);\\
& \text{} Generators step: 1 (MNIST, SVHN, CIFAR10, CIFAR100, MINI); \\
& \text{} CL step: 4 (MNIST, SVHN, CIFAR10, CIFAR100, MINI); \\
& \text{} Generators learning rate: 0.001  (MNIST, SVHN, CIFAR10, CIFAR100, MINI); \\
& \text{} Dimensions of random noise: 100 (MNIST), 256 (SVHN, CIFAR10, CIFAR100, MINI); \\
& \text{} Generators regularization coefficient $\lambda_{G}$:  1 (MNIST, SVHN, CIFAR10, CIFAR100, MINI); \\
& \text{} CL regularization coefficient $\lambda_{CL}$:  1 (MNIST, SVHN, CIFAR10, CIFAR100, MINI); \\
\bottomrule
\end{tabular}
}
\vspace{-5pt}
\label{tab:hyper_config}
\end{table}

\revised{
In the natural language processing domain, we use the GPT-2~\footnote{\footnotesize \url{https://huggingface.co/openai-community/gpt2}}~\cite{gpt2} model as the generator and simply use "<|endoftext|>" as the prompt for text generation similar to \cite{CarliniTWJHLRBS21}. We believe that the performance of CL from a stream of APIs will be further improved by designing better prompts in the future. In addition, for all APIs and CL models, we use the BERT~\footnote{\footnotesize \url{https://huggingface.co/google-bert/bert-base-uncased}}~\cite{bert} architecture uniformly.
}

\subsection{Implementation Details}
\label{sec:ImplementationDetails_Appendix}
\revised{In the computer vision domain,} we use the SGD optimizer for the CL model and the Adam optimizer~\cite{adam2015} for generators. In addition, we used an early stopping strategy to avoid overfitting. We run multiple random seeds for each method and report the mean and standard deviation. The hyperparameter configurations of all experimental methods are shown in Tab.~\ref{tab:hyper_config}.

\revised{
In the natural language processing domain, we follow \cite{huang2021continual} and update the CL model using AdamW optimizer~\cite{adamw}, and building pre-trained black-box APIs. The generator uses GPT-2~\cite{gpt2} but does not update. Tab.~\ref{tab:hyper_config_nlp} gives the detailed hyperparameter configuration.
}

\begin{table}[h]
\footnotesize
\centering
\revised{
\caption{The hyperparameters for baselines \revised{in the natural language processing domain}.  
}
\vspace{-10pt}
\resizebox{0.5\textwidth}{!}{
\begin{tabular}{@{}l ll@{}}
\toprule
\textbf{Methods}&   \textbf{Hyperparameters} \\
\midrule
Joint \text{  }              
& \text{} Learning rate: $3e\!-\!5$ (AGNews, Yelp, Amazon, DBPedia, Yahoo); \\
& \text{} Epoch: 3 (AGNews, Yelp, Amazon, DBPedia, Yahoo);\\ 
& \text{} Batch size: 8 (AGNews, Yelp, Amazon, DBPedia, Yahoo) \\ \midrule
Sequential \text{  }         
& \text{} Learning rate: $3e\!-\!5$ (AGNews, Yelp, Amazon, DBPedia, Yahoo); \\
& \text{} Epoch: 4 (AGNews), 3 (Yelp, Amazon), 2 (DBPedia), 1 (Yahoo));\\ 
& \text{} Batch size: 8 (AGNews, Yelp, Amazon, DBPedia, Yahoo) \\ \midrule
Classic CL  \text{  }  
& \text{} Learning rate: $3e\!-\!5$ (AGNews, Yelp, Amazon, DBPedia, Yahoo); \\
& \text{} Epoch: 4 (AGNews), 3 (Yelp, Amazon), 2 (DBPedia), 1 (Yahoo));\\ 
& \text{} Batch size: 8 (AGNews, Yelp, Amazon, DBPedia, Yahoo) \\ 
& \text{} Memory buffer ratio: 0.02 (AGNews, Yelp, Amazon, DBPedia, Yahoo)  \\ 
\midrule
DFCL-APIs  \text{  }    
& \text{} Learning rate: $3e\!-\!5$ (AGNews, Yelp, Amazon, DBPedia, Yahoo); \\
\&
& \text{} Epoch: 4 (AGNews), 3 (Yelp, Amazon), 2 (DBPedia), 1 (Yahoo));\\ 
DECL-APIs  \text{  } 
& \text{} Batch size: 8 (AGNews, Yelp, Amazon, DBPedia, Yahoo) \\ 
& \text{} Memory buffer ratio: 0.02 (AGNews, Yelp, Amazon, DBPedia, Yahoo)  \\ 
& \text{} Regularization coefficient $\lambda_{CL}$:  1 (AGNews, Yelp, Amazon, DBPedia, Yahoo) ; \\
\bottomrule
\end{tabular}
}
\vspace{-10pt}
\label{tab:hyper_config_nlp}
}
\end{table}

\subsection{Algorithm Details}
\label{subsec:AlgorithmDetails_appendix}

The algorithmic of our data-free cooperative continual distillation learning framework is summarized in Alg.~\ref{alg:our}. Our algorithm aims to update the parameters $\theta_{\mathcal{G}}=\{\theta_{\mathcal{G}_A}, \theta_{\mathcal{G}_B}\}$ of generators $\mathcal{G}=\{\mathcal{G}_A, \mathcal{G}_B\}$ and $\theta_{f_{cl}}$ of the CL model $f_{cl}$. 
We optimize generators and CL models by alternating updates until the CL model converges. Specifically, we first update generators $\mathcal{G}$ by generating samples in lines 7-13.
Second, we update the CL model $f_{cl}$ in lines 15 to 26. 
Finally, in the \textit{DFCL-APIs} setting, we store the image $\small \hat{\mathbf{X}}^k$ generated for the task $k$ and the output logits $\small f_b^k(\hat{\mathbf{X}^k})$ of the black-box API in memory $\mathcal{M}$ (lines 32-33). In the \textit{DECL-APIs} setting, we store a small portion (e.g., 2\%, 5\%, 10\%) of the accessible raw task data $\small {\mathbf{X}^k}$ in memory $\mathcal{M}$ (line 35). 

\begin{algorithm}[t]
\centering
\caption{Data-free Cooperative Continual Distillation Learning
}
\label{alg:our}
\begin{algorithmic}[1]
        \STATE \textbf{Require}: A stream of black-box APIs $\{f^{k}_b\}_{k=1}^K$, a CL model $f_{cl}$, two cooperative generators $\mathcal{G}=\{\mathcal{G}_A, \mathcal{G}_B\}$, the generators and the CL model update steps are $N_{\mathcal{G}}$ and $N_{f_{cl}}$, hyperparameters $\lambda_G, \lambda_{CL}$, learning rate $\eta$, the number of inner and outer loops $S$ and $E$, batch size $B$
        \STATE \textbf{Initialize}: Memory buffer $\mathcal{M} \gets \{ \space \}$ 
        \FOR{task $k$ $\gets$ ${1,2,\ldots,K}$} 
        \FOR{epoch $e$ $\gets$ ${1,2,\ldots,E}$ (outer loop)} 
        \FOR{step $s$ $\gets$ ${1,2,\ldots,S}$ (inner loop)} 
                \STATE {\COMMENT{Update generators: $\mathcal{G}=\{\mathcal{G}_A, \mathcal{G}_B\}$}}
                \FOR{iter $t$ $\gets$  ${1,2,\ldots,N_{\mathcal{G}}}$}    
                    \STATE data generation $\hat{\mathcal{D}}^{k}_t$ $\gets$  $\{\mathcal{G}_A\left(\mathbf{z} \right),\mathcal{G}_B\left(\mathbf{z}\right) \}$, $\mathbf{z} \sim \revised{\mathcal{N}({0},{I})}$
                    \STATE compute losses $\mathcal{L}_{\mathcal{G}}^k$, $\mathcal{L}_{C}$, and $\mathcal{L}_{B}$ by Eq.~\ref{eq:loss_generate}, Eq.~\ref{eq:loss_diversity}, Eq.~\ref{eq:loss_balanced}
                    \STATE compute gradient $\nabla_{\theta_{\mathcal{G}}^t} 
                    \mathcal{L}_{C}$, $\nabla_{\theta_{\mathcal{G}}^t} 
                    \mathcal{L}_{B}$
                    \STATE estimated gradient $\hat{\nabla}_{\theta_{\mathcal{G}}^t} \mathcal{L}^k_{\mathcal{G}} $ by Eq.~\ref{eq:gradient_generate_zerothorder} \revised{(i.e., Alg.~\ref{alg:implement_zero})}
                    \STATE gradient update $\theta_{\mathcal{G}}^{t+1}$ $\gets$ $\theta_{\mathcal{G}}^{t}-\eta \cdot (\hat{\nabla}_{\theta_{\mathcal{G}}^t} \mathcal{L}^k_{\mathcal{G}} + \lambda_G \cdot ( \nabla_{\theta_{\mathcal{G}}^t} 
                    \mathcal{L}_{C} +  \nabla_{\theta_{\mathcal{G}}^t} 
                    \mathcal{L}_{B}))$
                \ENDFOR
                \STATE {\COMMENT{Update CL model: $f_{cl}$}}
                \FOR{step $t$ $\gets$  ${1,2,\ldots,N_{f_{cl}}}$}  
                    \STATE data generation $\hat{\mathcal{D}}^{k}_t$ $\gets$  $\{\mathcal{G}_A\left(\mathbf{z} \right),\mathcal{G}_B\left(\mathbf{z}\right) \}$, $\mathbf{z} \sim \revised{\mathcal{N}({0},{I})}$
                    \STATE compute losses $\mathcal{L}_{f_{cl}}^k$ by Eq.~\ref{eq:loss_cl} 
                    \IF{task $k > 1$}
                        \STATE memory selection $\mathcal{B}$ $\gets$ $(\mathbf{X}_s, \mathbf{Y}_s)$ $\gets$ sample($\mathcal{M}$)
                         \STATE compute losses $\mathcal{L}_{M}^k$, $\mathcal{L}_{S}^k$ by Eq.~\ref{eq:loss_memory}, Eq.~\ref{eq:loss_distance_correlation}
                        \STATE $\mathcal{L}_{f_{cl}}^{k'} \gets \mathcal{L}_{f_{cl}}^k + \lambda_{CL} \cdot( \mathcal{L}_{{M}}^k + \mathcal{L}_{S}^k )$
                    \ELSE
                        \STATE $\mathcal{L}_{f_{cl}}^{k'} \gets \mathcal{L}_{f_{cl}}^k $
                    \ENDIF
                    \STATE compute gradient $\nabla_{\theta_{cl}^t}  \mathcal{L}^{k'}_{cl} $
                    \STATE gradient update $\mathbf{\theta}_{f_{cl}}^{t+1}$ $\gets$ $\mathbf{\theta}_{f_{cl}}^{t}-\eta \cdot \nabla_{\theta_{f_{cl}}^t} \mathcal{L}^{k'}_{f_{cl}} $
                \ENDFOR
            \ENDFOR
            \ENDFOR
            \STATE {\COMMENT{Memory update: $\mathcal{M}$}}
            \IF{DFCL-APIs setting}
               \STATE $\hat{\mathbf{X}}^{k} \gets \{\mathcal{G}_A\left(\mathbf{z}\right),\mathcal{G}_B\left(\mathbf{z}\right)\}, \mathbf{z} \sim \revised{\mathcal{N}({0},{I})}$,   
               \STATE $\hat{\mathcal{D}}^{k} \gets \left(\hat{\mathbf{X}}^{k}, \hat{\mathbf{Y}}^{k}, k  \right)$, $\hat{\mathbf{Y}}^{k} \gets f_b^k (\hat{\mathbf{X}}^{k})$
            \ELSIF{DECL-APIs setting}
                \STATE ${\mathcal{D}}^{k} \gets \left({\mathbf{X}}^{k}, {\mathbf{Y}}^{k}, k \right)$, ${\mathbf{Y}}^{k} \gets f_b^k ({\mathbf{X}}^{k})$
            \ENDIF 
            \STATE $\mathcal{M} \gets   \mathcal{M} \cup {\mathcal{D}}^{k}$
        \ENDFOR
        \STATE \textbf{return} $f_{cl}$ 
\end{algorithmic}
\end{algorithm}

\begin{algorithm}
\small
\revised{
\caption{\revised{Implementation details for parameter update of generator $\mathcal{G}$ (i.e., Eq.~\ref{eq:gradient_generate_zerothorder} in main text)}}
\label{alg:implement_zero}
\begin{algorithmic}[1]
        \STATE optimizer\_$\mathcal{G}$.zero\_grad() \COMMENT{Empty the gradient of the optimizer with respect to $\mathcal{G}$}
		\STATE $\mathbf{z}$ = torch.randn(batch\_size, latent\_dim) \COMMENT{A set of random noises are sampled from a Gaussian distribution}
        \STATE $\hat{\mathbf{x}}$ = $\mathcal{G}$($\mathbf{z}$) \COMMENT{A set of pseudo images is generated based on random noise $\mathbf{z}$}
        \STATE approx\_grad\_wrt\_hat\_x = estimate\_gradient\_objective($f^k_{b}$, $f_{cl}$, $\hat{\mathbf{x}}$, $d$, $\epsilon$) \COMMENT{The gradient $\small \frac{\partial \mathcal{L}^k_{\mathcal{G}}}{\partial \hat{\mathbf{x}} }$ of $\small \hat{\mathbf{x}}$ is obtained by zeroth-order gradient estimation (e.g., a forward differences method).}
        \STATE \textbf{$\hat{\mathbf{x}}$.backward(approx\_grad\_hat\_wrt\_x)} \COMMENT{Compute the gradient $\small \frac{\partial \hat{\mathbf{x}} }{\partial \theta_{\mathcal{G}}}$ of $\hat{\mathbf{x}}$ with respect to $\small \theta_{\mathcal{G}}$ using Pytorch's automatic differentiation operation.}
        \STATE optimizer\_$\mathcal{G}$.step() \COMMENT{The gradient descent algorithm is used to update the parameters $\small \theta_{\mathcal{G}}$}
	\end{algorithmic}  
 }
\end{algorithm}

\section{More Experiment Analysis}
\label{sec:moreanalysis_appendix}

\revised{In this section, we show more experimental analysis in terms of ablation study, query cost, hyperparameter analysis, and accuracy visualization.}

\subsection{Effectiveness of Each Component} 
In Sec.~\ref{sec:trainning_generative}, we applied two regularization terms, \textit{Collaborative Generators} and \textit{Class-balanced}, to make the generated images diverse and evenly distributed in each class. 
In Sec.~\ref{sec:trainning_cl}, we use \textit{Memory Replay} and design a \textit{Network Similarity}-based regularization strategies to avoid catastrophic forgetting. We verified the effectiveness of four components under the DFCL-APIs setting.
As shown in Tab.~\ref{tab:component}, the removal of any single component leads to a decrease in accuracy. For instance, on the CIFAR-10 dataset, removing the memory buffer and network similarity in the CL model part resulted in a drop from $\small 70.63\%$ to $\small 59.92\%$ and $\small 66.73\%$, respectively. The removal of the collaborative generator part and the class-balanced part led to a drop in ACC to $\small 67.50\%$ and $\small 69.75\%$, respectively. Removing all four components can result in a sharp drop in performance, with an ACC of only $\small 52.96\%$.

\begin{table}[h]
\centering
\small
\caption{Ablation study on \revised{Split-}MNIST, \revised{Split-}CIFAR10 and \revised{Split-}CIFAR100 datasets.
}
\vspace{-8pt}
\resizebox{0.5\textwidth}{!}{
\begin{tabular}{ l c c  c}
\toprule
 Variant         & \multicolumn{1}{c}{\revised{Split-}MNIST}  & \multicolumn{1}{c}{\revised{Split-}CIFAR10}  & \multicolumn{1}{c}{\revised{Split-}CIFAR100}  \\
\midrule
DFCL-APIs (Ours)              &  98.58$\pm$0.41     & 70.63$\pm$6.03   & 30.52$\pm$1.51 \\  
\midrule
{w/o Memory Replay}       &  94.47$\pm$1.76     & 59.92$\pm$5.16   & 21.55$\pm$1.62 \\
{w/o Network Similarity}  &  98.48$\pm$0.20     & 66.73$\pm$2.72   & 26.79$\pm$2.26 \\ 
{w/o Cooperative Generators}     &  96.25$\pm$3.38     & 67.50$\pm$3.05   & 26.33$\pm$0.85 \\
{w/o Class Balance}      &  98.52$\pm$0.54     & 69.75$\pm$1.46   & 25.70$\pm$1.68 \\
{w/o Four Losses}                      &  90.13$\pm$7.54     & 52.96$\pm$3.30   & 16.47$\pm$1.02 \\
\bottomrule
\end{tabular}
}
\label{tab:component}
\vspace{-8pt}
\end{table}

\noindent
\subsection{Query Cost Analysis} 
To transfer the knowledge of the pre-trained black-box API to the CL model, we need to input the generated image into the API for a query and obtain the output logits. In this part, we study the impact of different query costs on the accuracy of the CL model under the DFCL-APIs setting. 
In our method, four API queries are required for each sample in each Generator model training step; that is, two API queries are needed for gradient estimation for each Generator, and we have two Generators. The API must be queried once per sample in each step when training the CL model. Therefore, the total API query cost for each task training is $E \times S \times B \times (4 \times N_{\mathcal{G}} + N_{f_{cl}} )$, where $E$, $S$, and $B$ represent the number of outer epochs, inner steps, and batch size, respectively, and $N_{\mathcal{G}}$ and $N_{f_{cl}}$ represent the number of times the generator models and the CL model are updated in each step.

The results are shown in Tab.~\ref{tab:querybudget}, we made the following observations: (i) As the query cost of each task increases, the final ACC gradually increases. For example, on the CIFAR100 dataset, the query cost increases from 128K to 256K and then to 640K, resulting in an increase in the final ACC from 27.75\% to 28.87\% and then to 30.52\%. This is because the higher the query cost per task, the more comprehensive the knowledge exploration of the API will be. However, (ii) increasing the query cost to 1280K leads to a decrease in the final ACC. This is due to the problem of catastrophic forgetting in CL. When the new API performs many updates to the CL model, it leads to catastrophic forgetting of old tasks, which can be seen from the BWT metric. When the query cost is 256K, 640K, and 1280K, the degree of forgetting is 20.21\%, 21.40\%, and 25.13\%, respectively. The above results suggest that in the field of CL, there is a need to balance the impact of new task plasticity and old task forgetting on the final performance.
It should be mentioned that a large number of queries increases the financial cost of calling the API~\cite{MEGEX_Arxiv2021}, so it also makes sense to explore how to reduce the cost of API queries, which is beyond the scope of this paper and will be left for future work.

\begin{table}[h]
\small
\centering
\caption{Analysis of API query cost (per task) on the \revised{Split-}MNIST, \revised{Split-}CIFAR10, and \revised{Split-}CIFAR100 datasets.
}
\vspace{-10pt}
\resizebox{0.5\textwidth}{!}{
\begin{tabular}{ c ccccccc}
\toprule
\multirow{3}{*}{\revised{Split-}MNIST}  
 & Maximum Budget & 12K     & 64K     & 128K          & 256K  \\ 
 &  ACC (\%)    & $ $ 94.86    & $ $ 96.75     & $ $ 98.58         & $ $ 96.41     \\  
 &  BWT (\%)    & $ $ 23.28    & $ $ 00.16     & -00.13        & -00.26     \\  
 \midrule
 \multirow{3}{*}{\revised{Split-}CIFAR10} 
 &  Maximum Budget          & 128K      & 256K      & 640K          & 1280K  \\
 &  ACC (\%)    & $ $ 67.72     & $ $ 68.03      & $ $ 70.63          & $ $ 67.99     \\  
 &  BWT (\%)    & -05.67    & -03.53     & -09.98         & -12.11     \\  
 \midrule
  \multirow{3}{*}{\revised{Split-}CIFAR100} 
  & Maximum Budget          & 128K      & 256K      & 640K          & 1280K  \\
 &  ACC (\%)    & $ $ 27.75      & $ $ 28.87     & $ $ 30.52          & $ $ 29.27     \\  
 &  BWT (\%)    & -13.62    & -20.21     & -21.41         & -22.57     \\  
\bottomrule
\end{tabular}
}
\label{tab:querybudget}
\vspace{-10pt}
\end{table}

\noindent
\subsection{Hyperparameter Analysis}
\begin{figure}[h]
\centering
\vspace{-12pt}
\includegraphics[width=.23\textwidth]{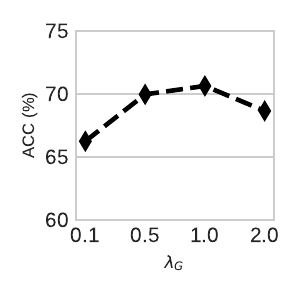}
\vspace{-10pt}
\includegraphics[width=.23\textwidth]{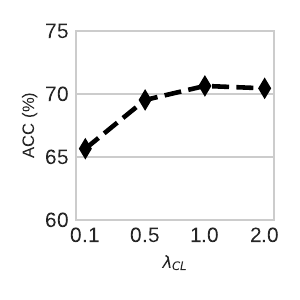}
\vspace{-10pt}
\caption{Hyperparameter sensitivity analysis on \revised{Split-}CIFAR10 dataset.}
\label{fig:hyperparameter_sensitivity}
\vspace{-5pt}
\end{figure}

As shown in Fig.~\ref{fig:hyperparameter_sensitivity}, we evaluate the performance sensitivity of our framework for different values of $\lambda_G$ and $\lambda_{CL}$ in Sec.~\ref{sec:trainning_generative} and Sec.~\ref{sec:trainning_cl} on the CIFAR10 dataset, respectively. We observed that the two hyperparameters have similar trends. When the regularization strength is low, i.e., the hyperparameter value is small, the generated images may lack diversity or have a class imbalance, and the CL model may suffer from catastrophic forgetting, resulting in poor performance. A better result can be obtained when the values of $\small \lambda_G$ and $\small \lambda_{CL}$ are set close to 1. We believe that further performance improvements can be achieved through finer tuning of the hyperparameters in the future. 

\noindent
\subsection{Accuracy Visualization}
\begin{figure*}[h]
\centering
\includegraphics[width=.33\textwidth]{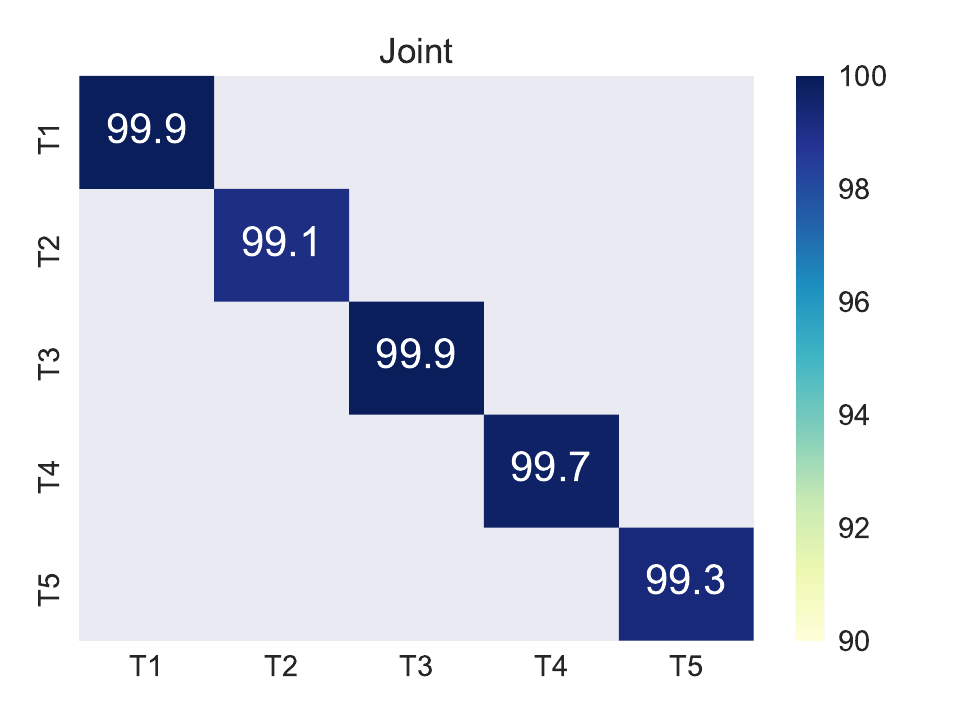}
\includegraphics[width=.33\textwidth]{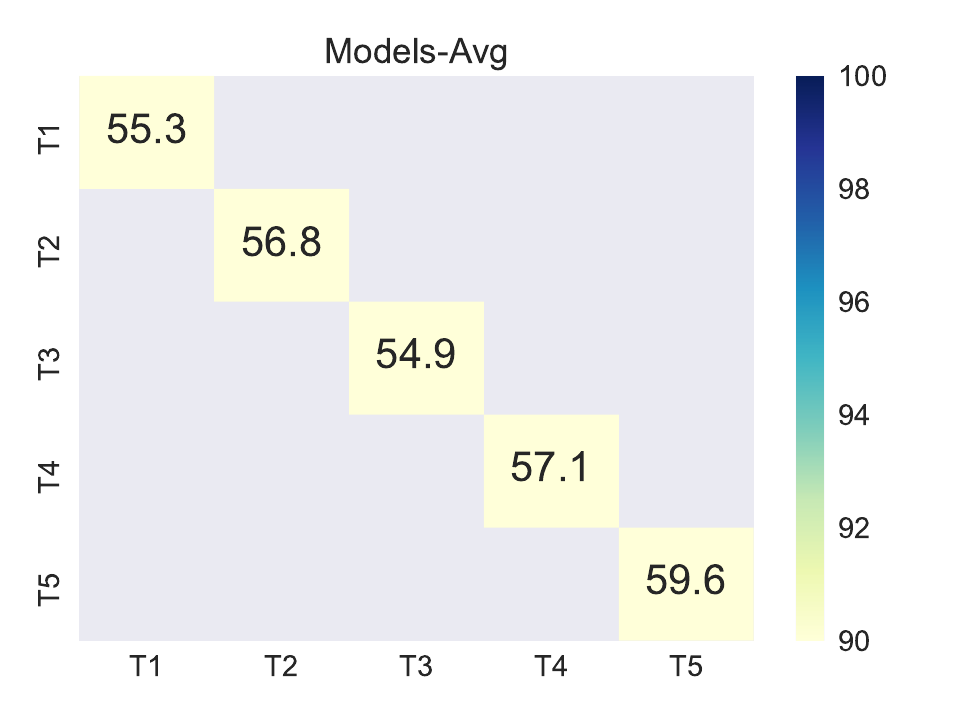}
\includegraphics[width=.33\textwidth]{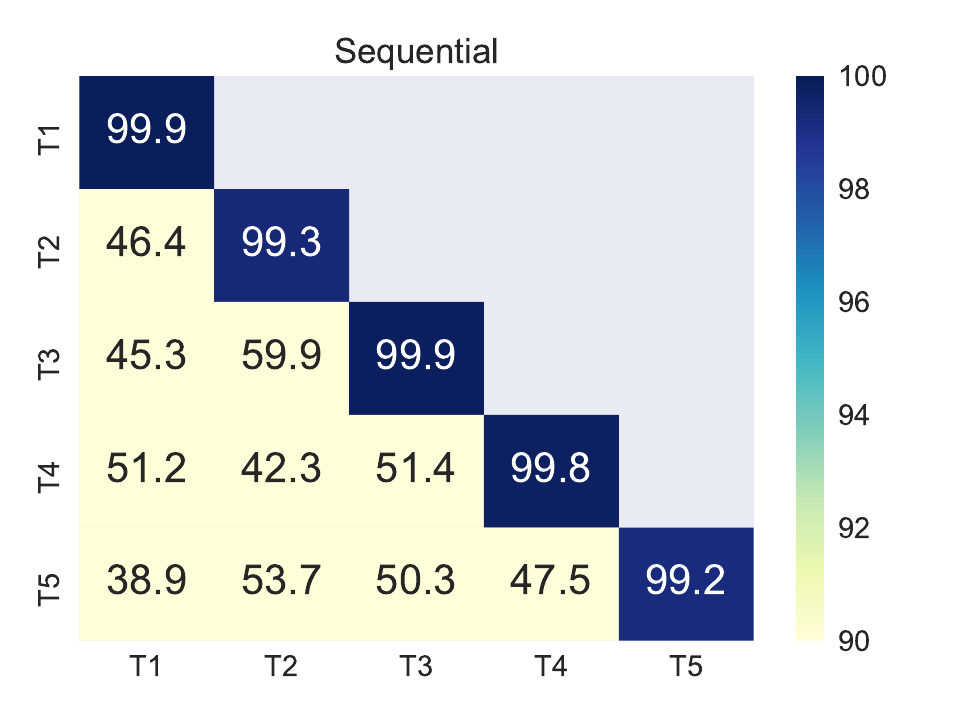}
\includegraphics[width=.33\textwidth]{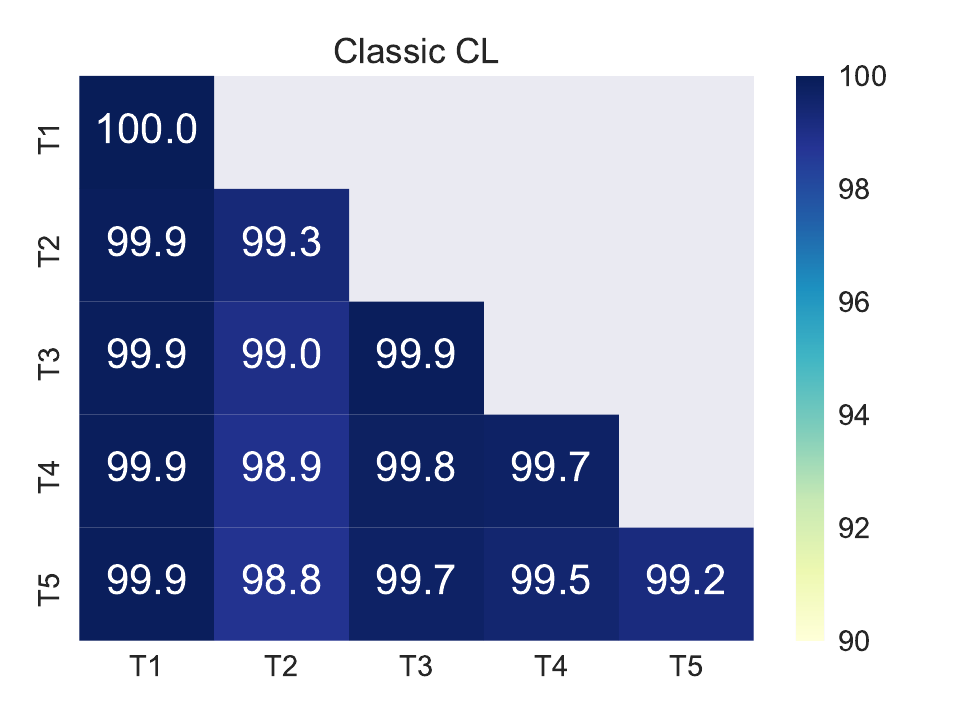}
\includegraphics[width=.33\textwidth]{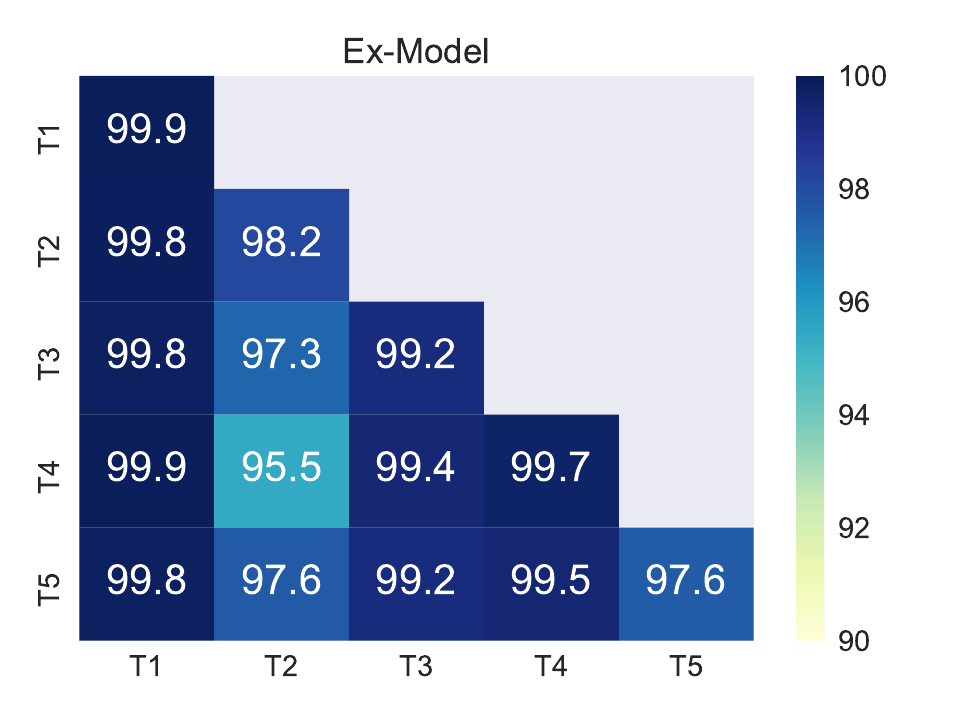}
\includegraphics[width=.33\textwidth]{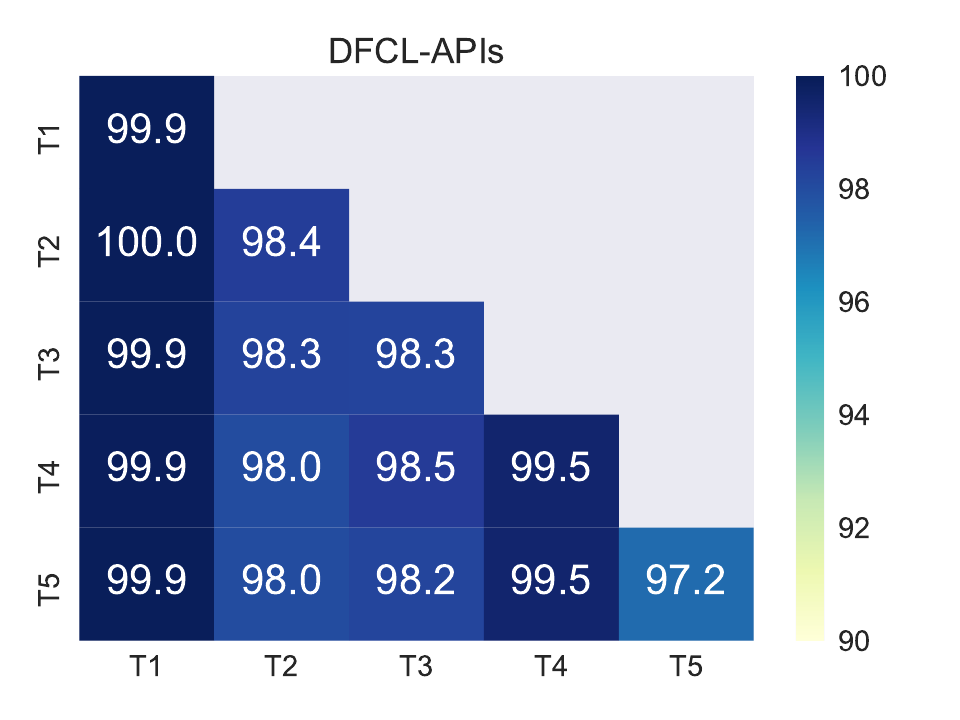}
\includegraphics[width=.33\textwidth]{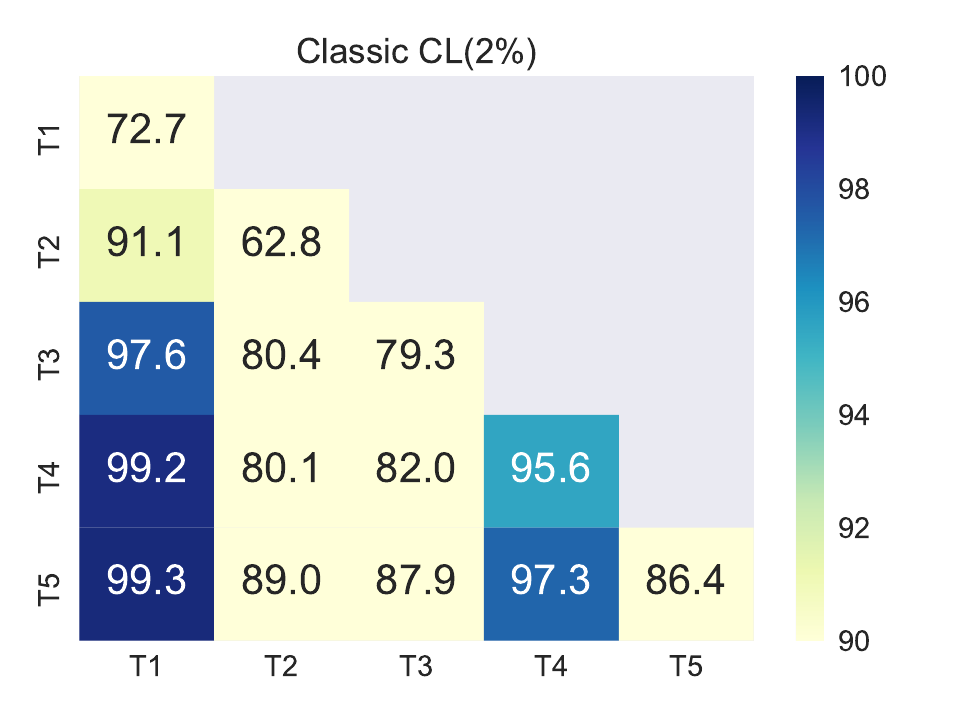}
\includegraphics[width=.33\textwidth]{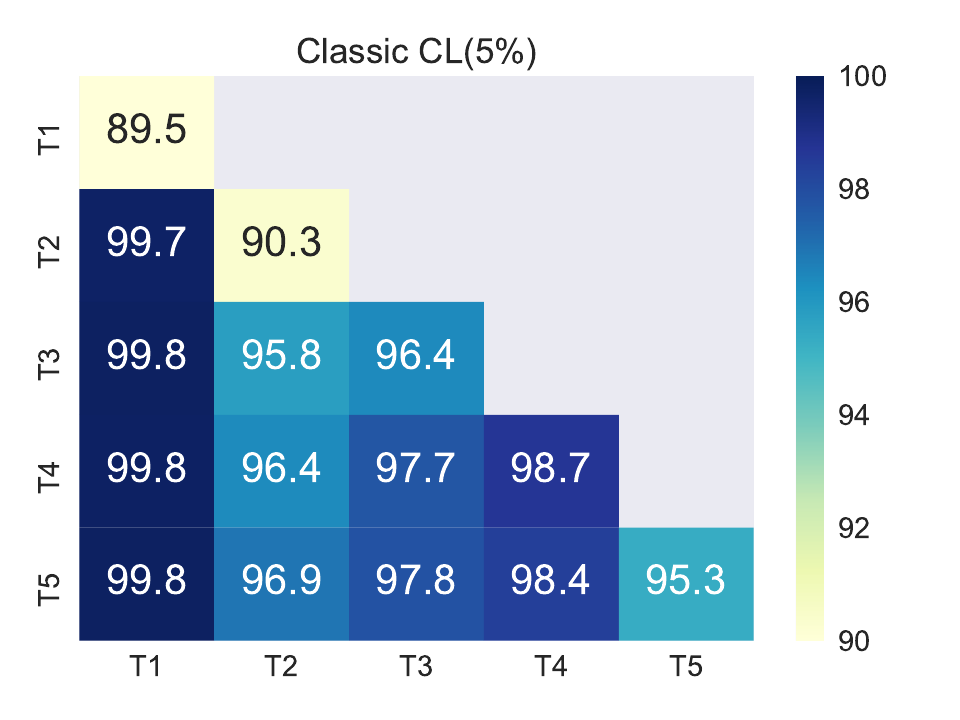}
\includegraphics[width=.33\textwidth]{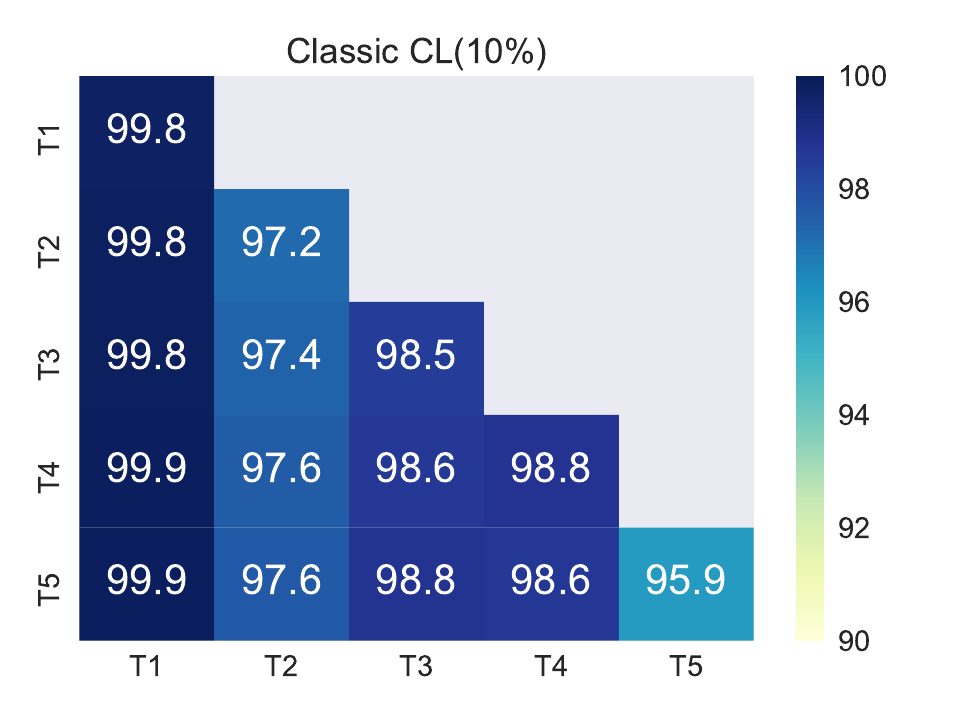}
\includegraphics[width=.33\textwidth]{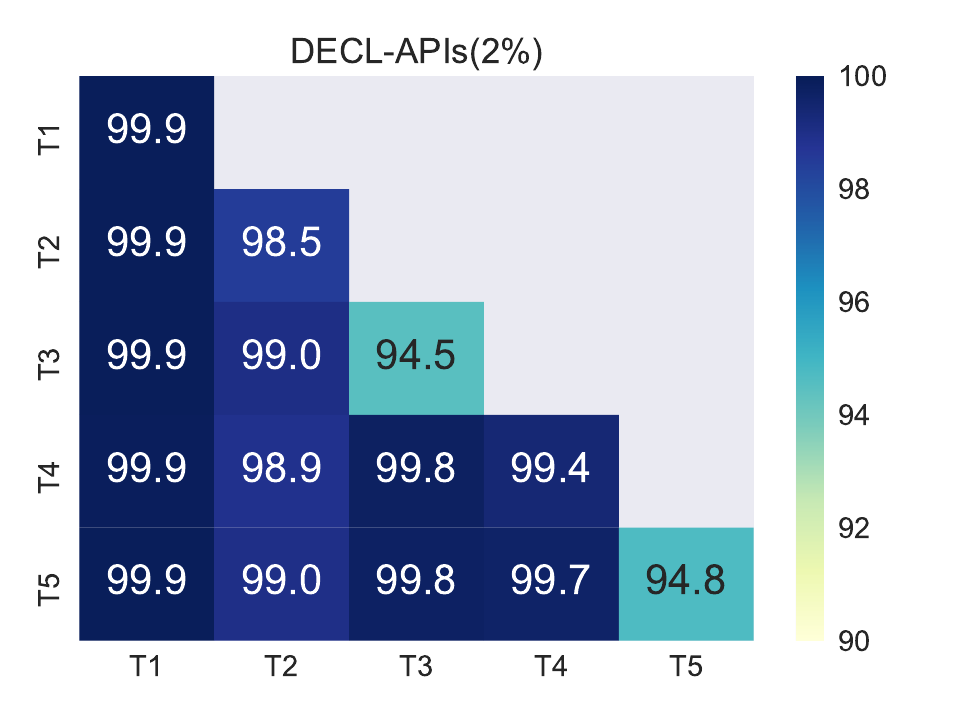}
\includegraphics[width=.33\textwidth]{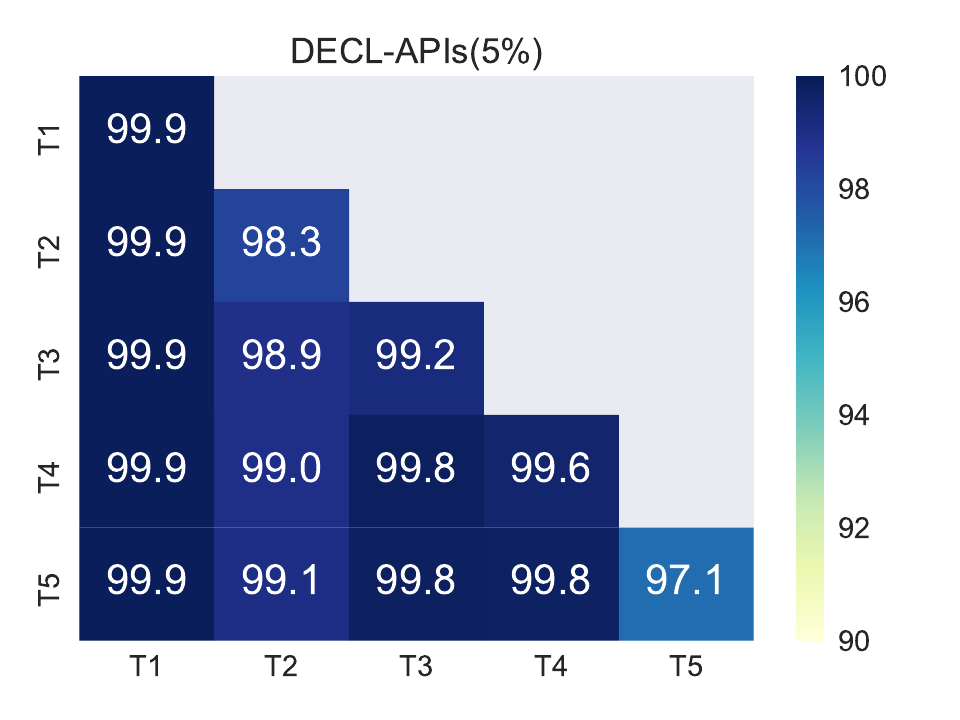}
\includegraphics[width=.33\textwidth]{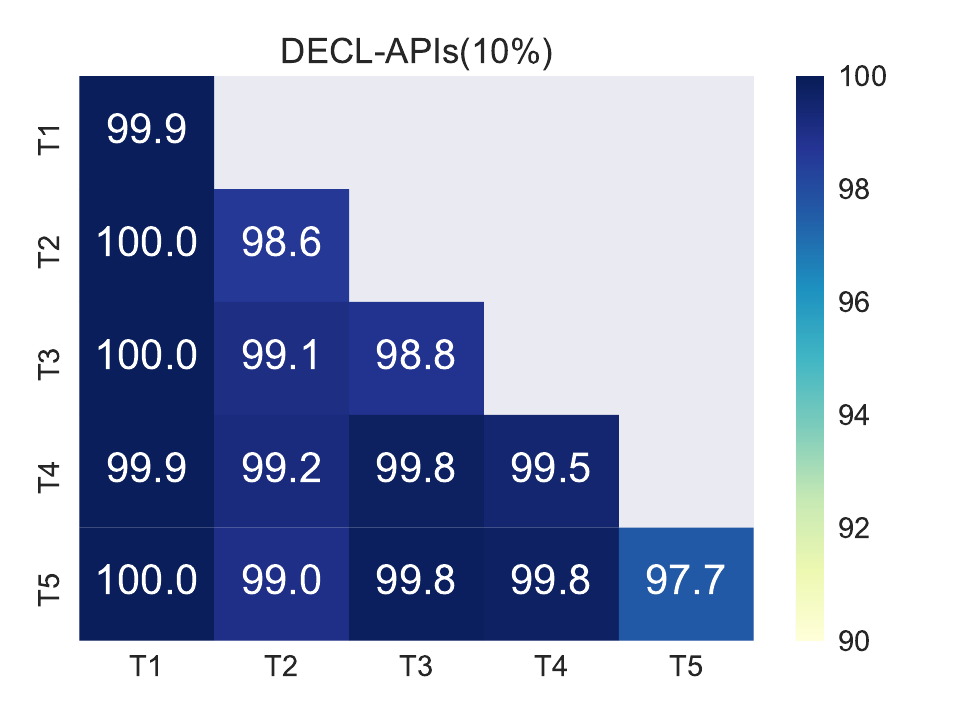}
\caption{The accuracy (Higher Better) on the \textbf{\revised{Split-}MNIST} dataset. (1) Joint, (2) Models-Avg, (3) Sequential, (4) Classic CL, (5) Ex-Model, (6) DFCL-APIs(ours), (7) Classic CL-2\%, (8) Classic CL-5\%, (9) Classic CL-10\%, (10) DECL-APIs-2\%(ours), (11) DECL-APIs-5\%(ours), (12) DECL-APIs-10\%(ours). $t$-th row represents the accuracy of the network tested on tasks $1-t$ after task $t$ is learned.
}
\label{fig:acc_mnist}
\end{figure*}

\begin{figure*}[h]
\centering
\includegraphics[width=.33\textwidth]{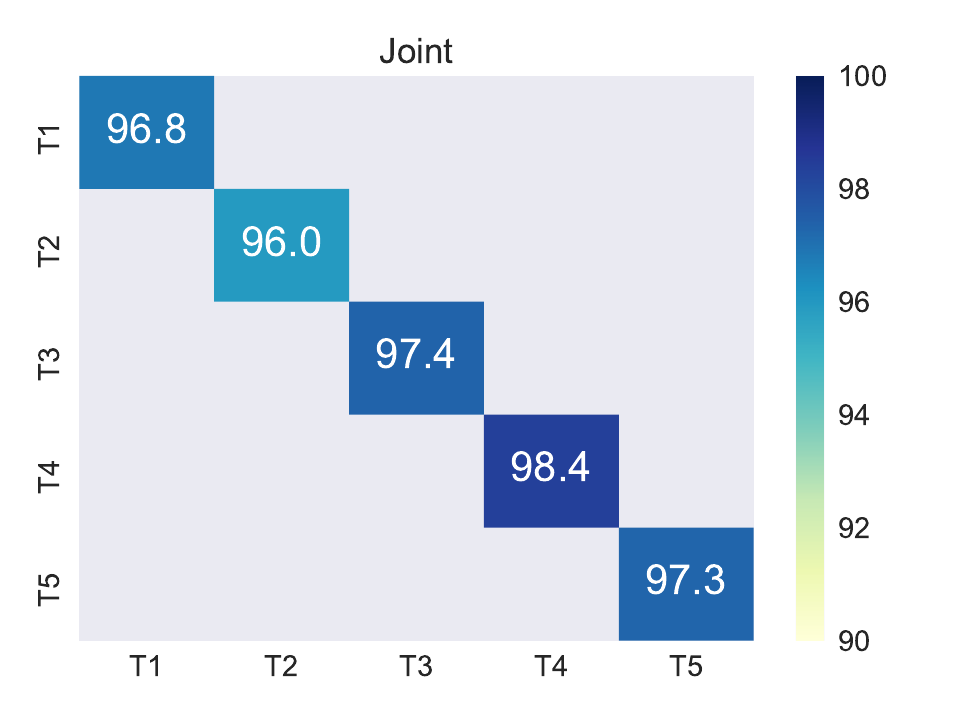}
\includegraphics[width=.33\textwidth]{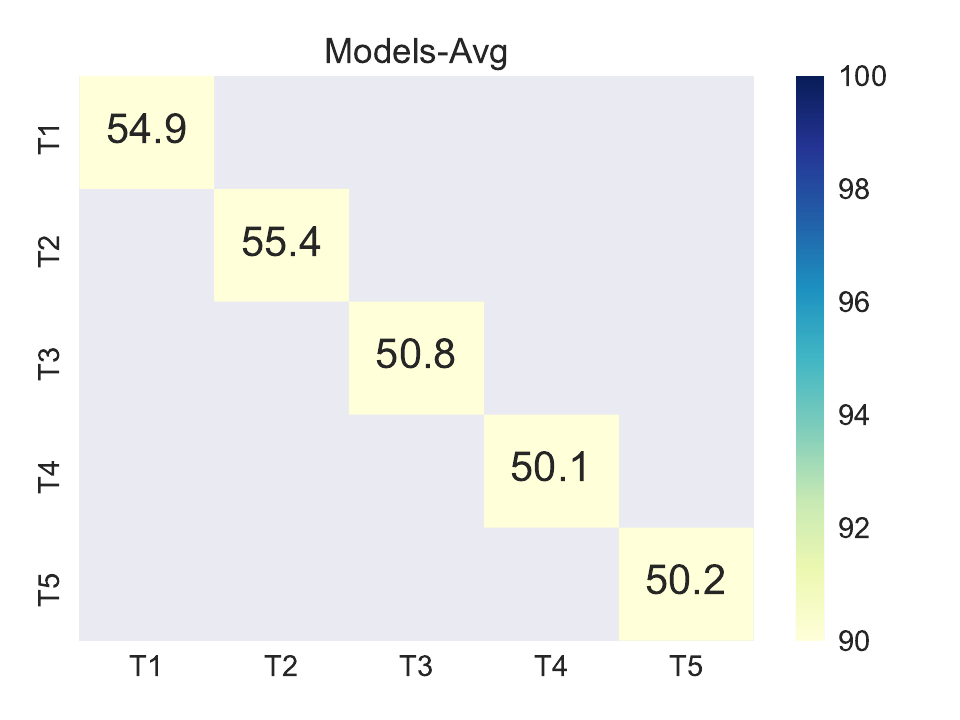}
\includegraphics[width=.33\textwidth]{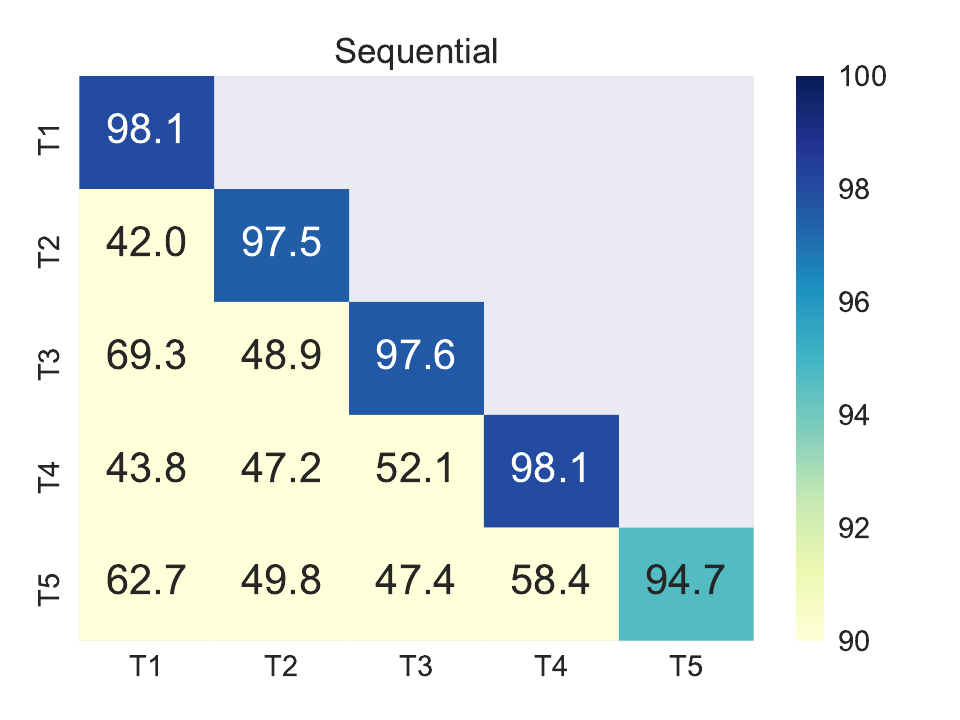}
\includegraphics[width=.33\textwidth]{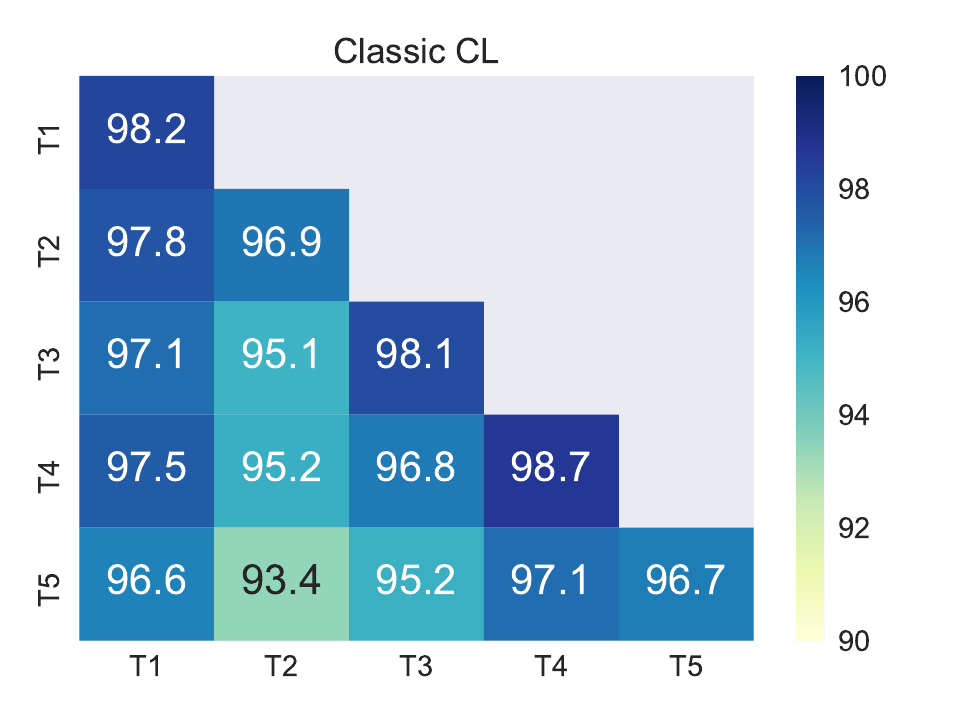}
\includegraphics[width=.33\textwidth]{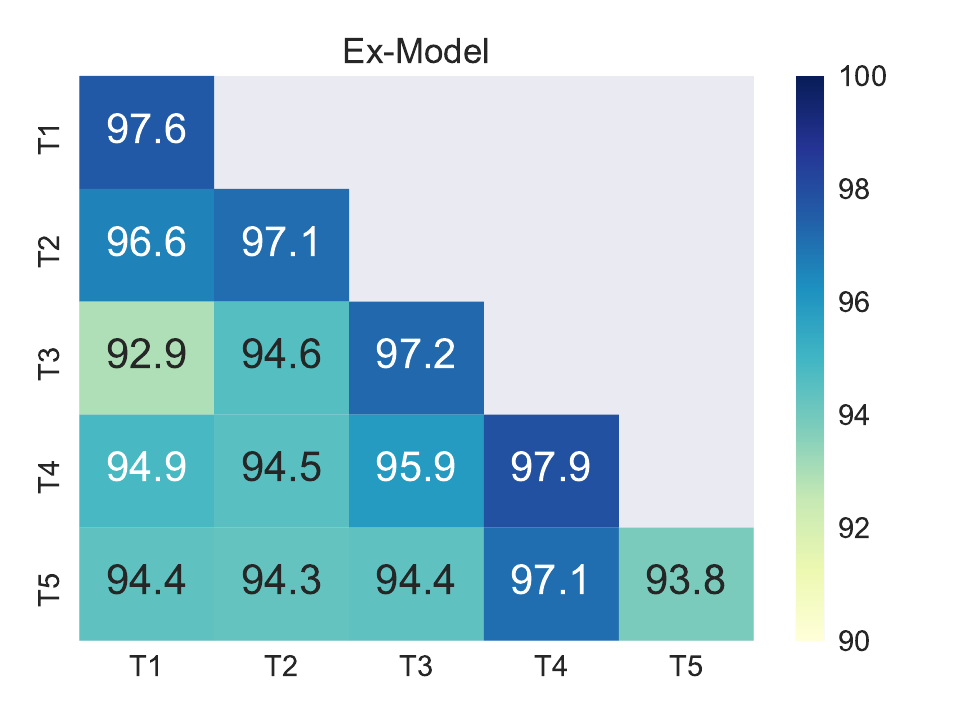}
\includegraphics[width=.33\textwidth]{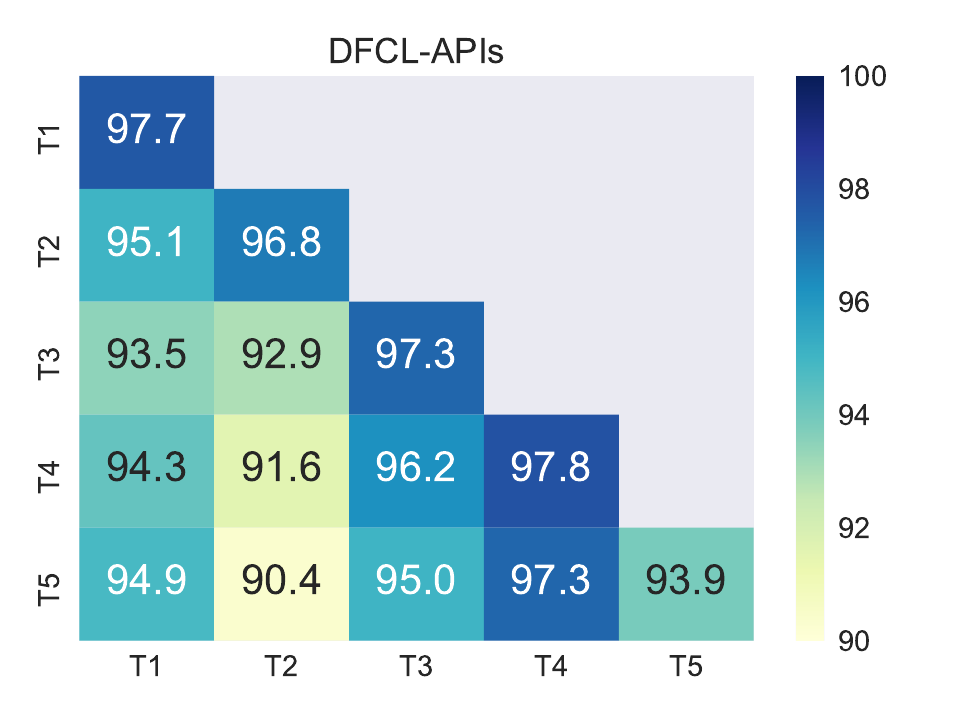}
\includegraphics[width=.33\textwidth]{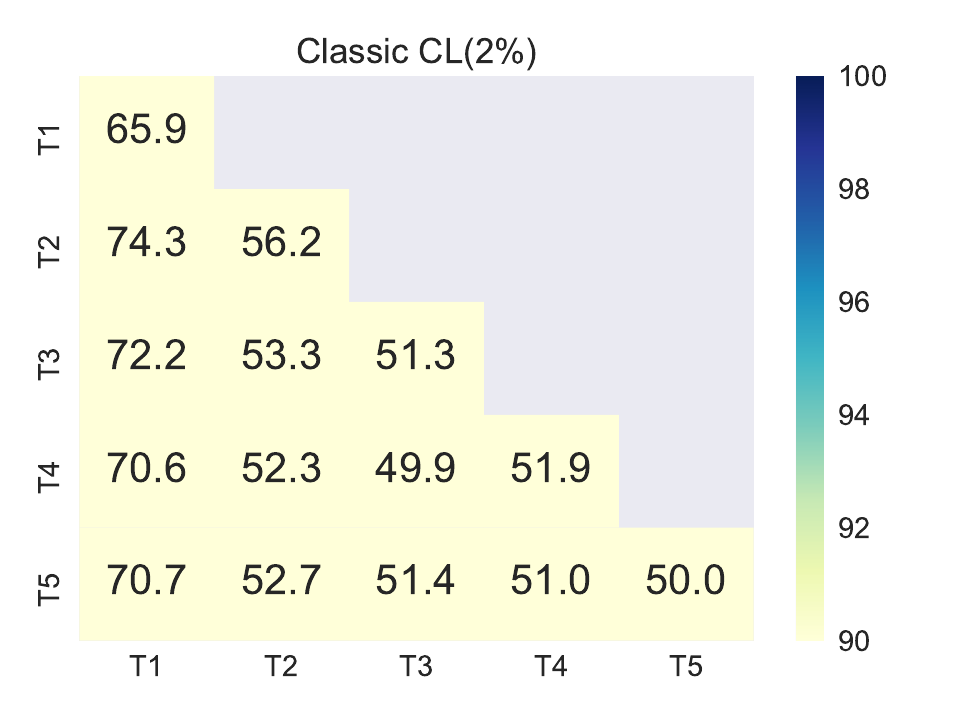}
\includegraphics[width=.33\textwidth]{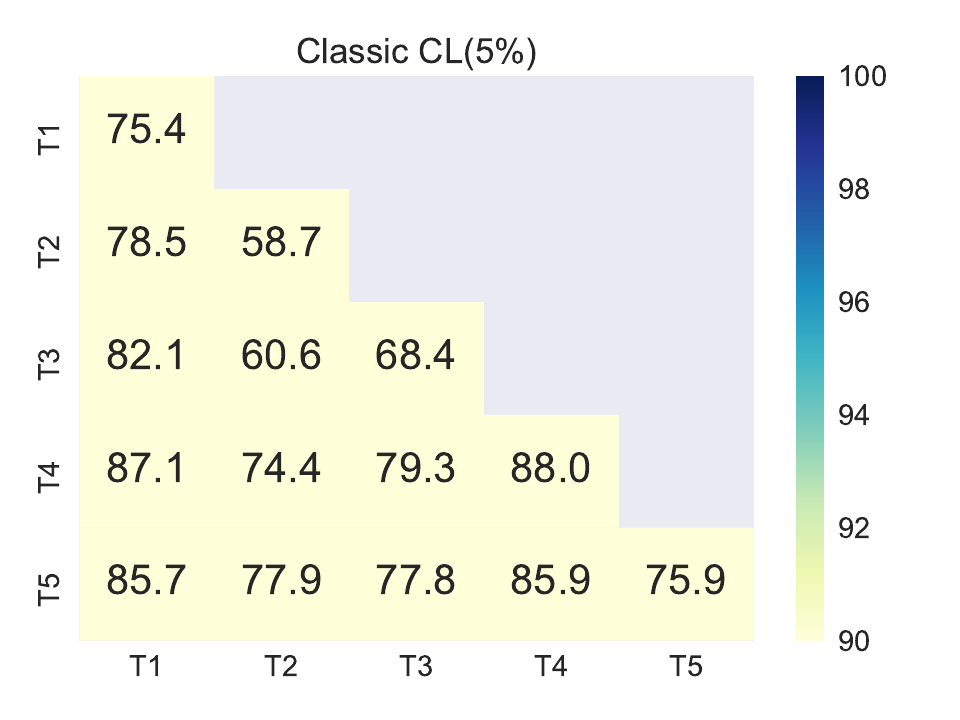}
\includegraphics[width=.33\textwidth]{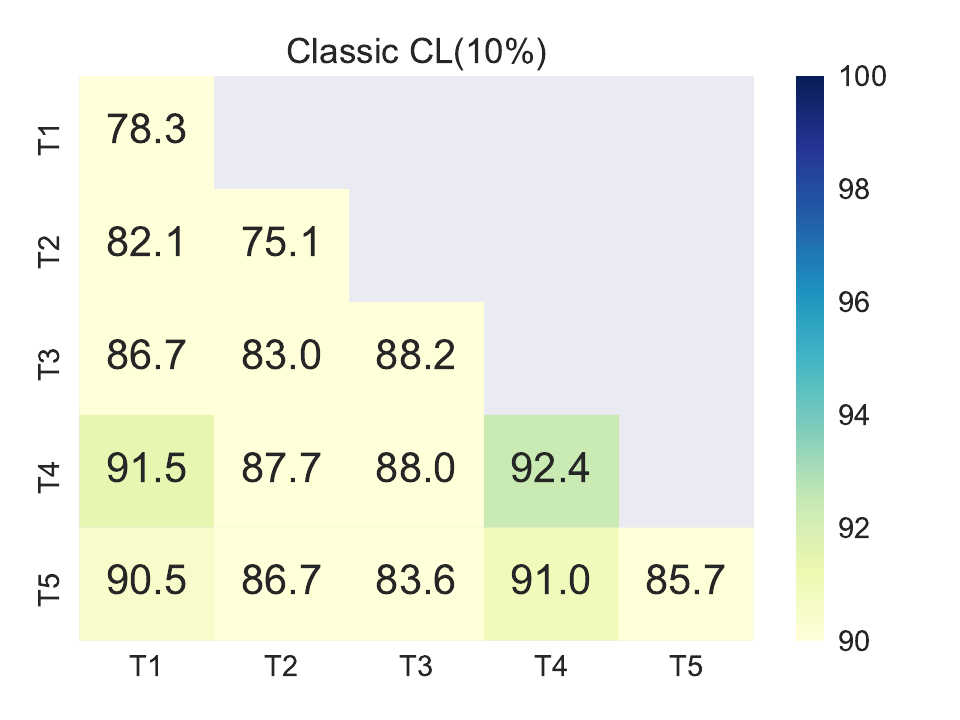}
\includegraphics[width=.33\textwidth]{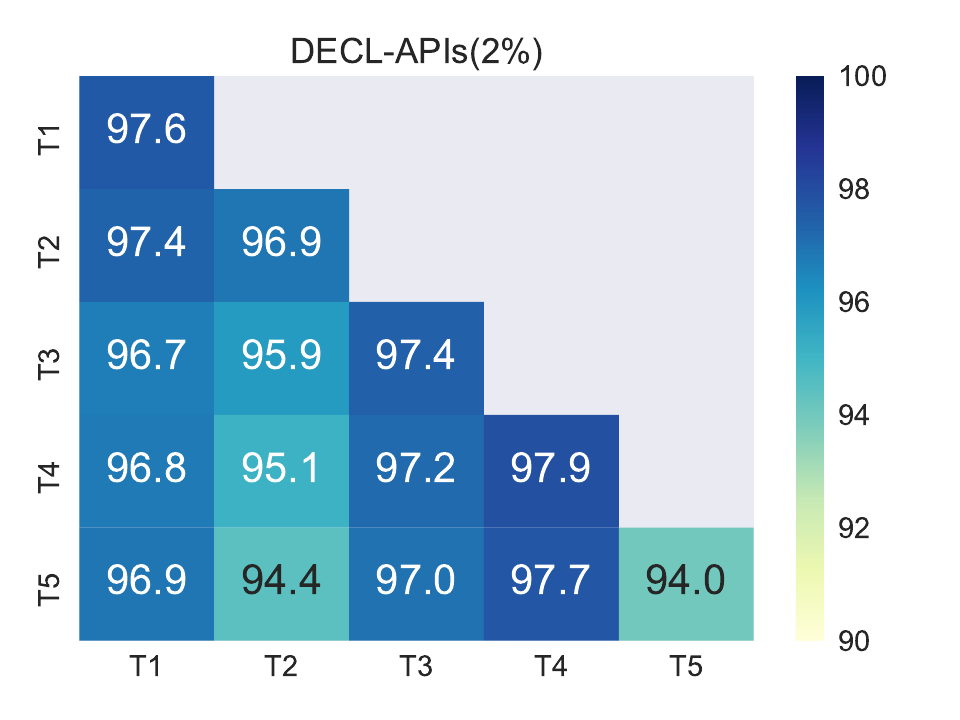}
\includegraphics[width=.33\textwidth]{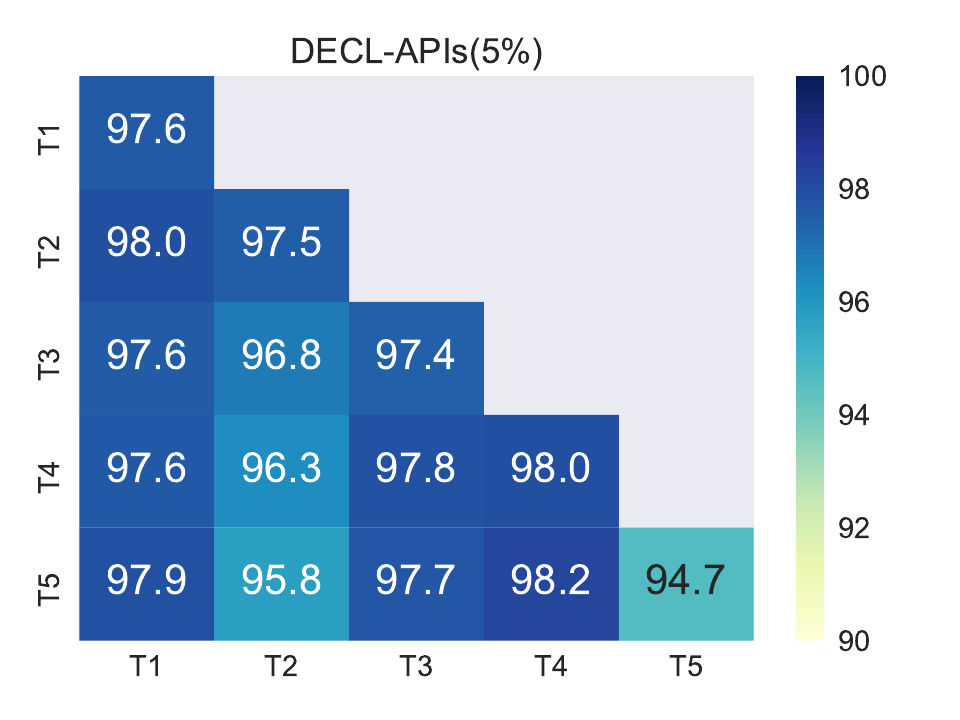}
\includegraphics[width=.33\textwidth]{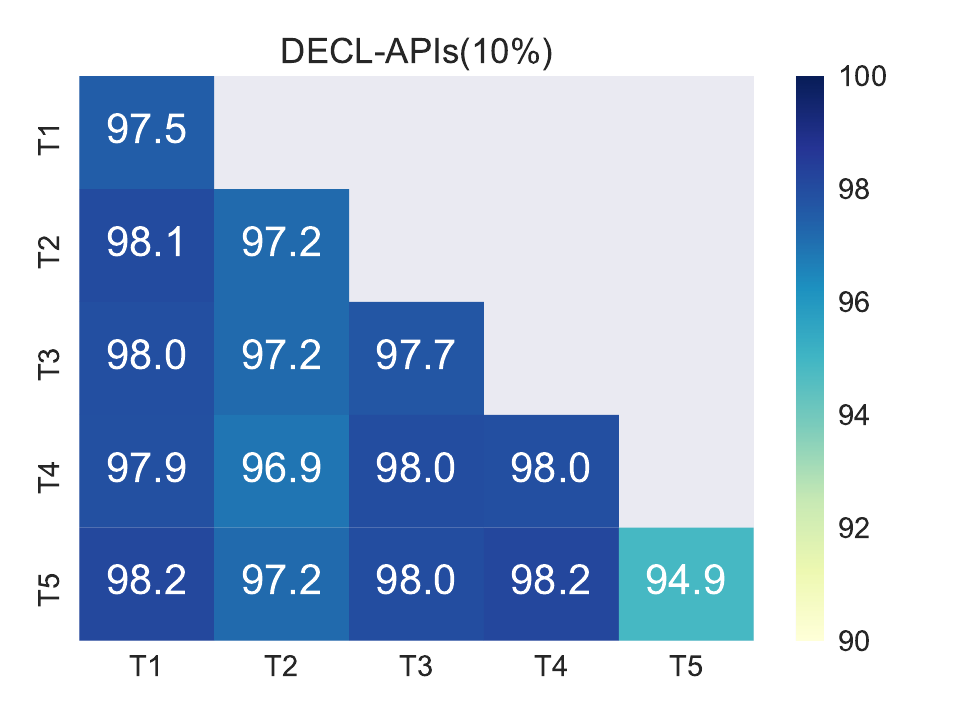}
    \caption{The accuracy (Higher Better) on the \textbf{\revised{Split-}SVHN} dataset. (1) Joint, (2) Models-Avg, (3) Sequential, (4) Classic CL, (5) Ex-Model, (6) DFCL-APIs(ours), (7) Classic CL-2\%, (8) Classic CL-5\%, (9) Classic CL-10\%, (10) DECL-APIs-2\%(ours), (11) DECL-APIs-5\%(ours), (12) DECL-APIs-10\%(ours). $t$-th row represents the accuracy of the network tested on tasks $1-t$ after task $t$ is learned.
}
\label{fig:acc_svhn}
\end{figure*}

\begin{figure*}[h]
\centering
\includegraphics[width=.33\textwidth]{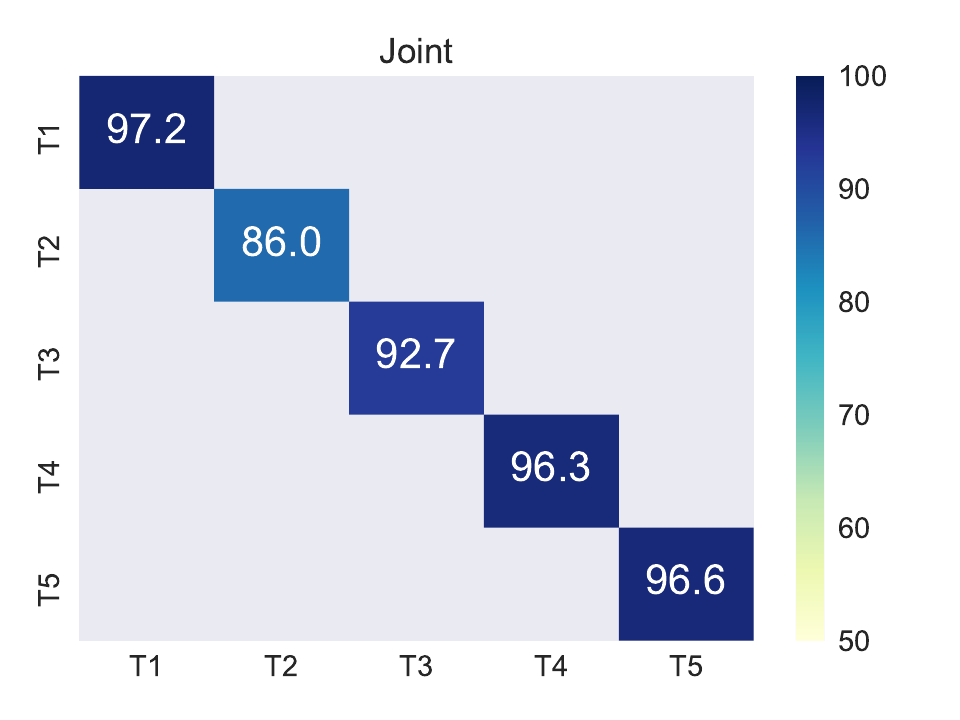}
\includegraphics[width=.33\textwidth]{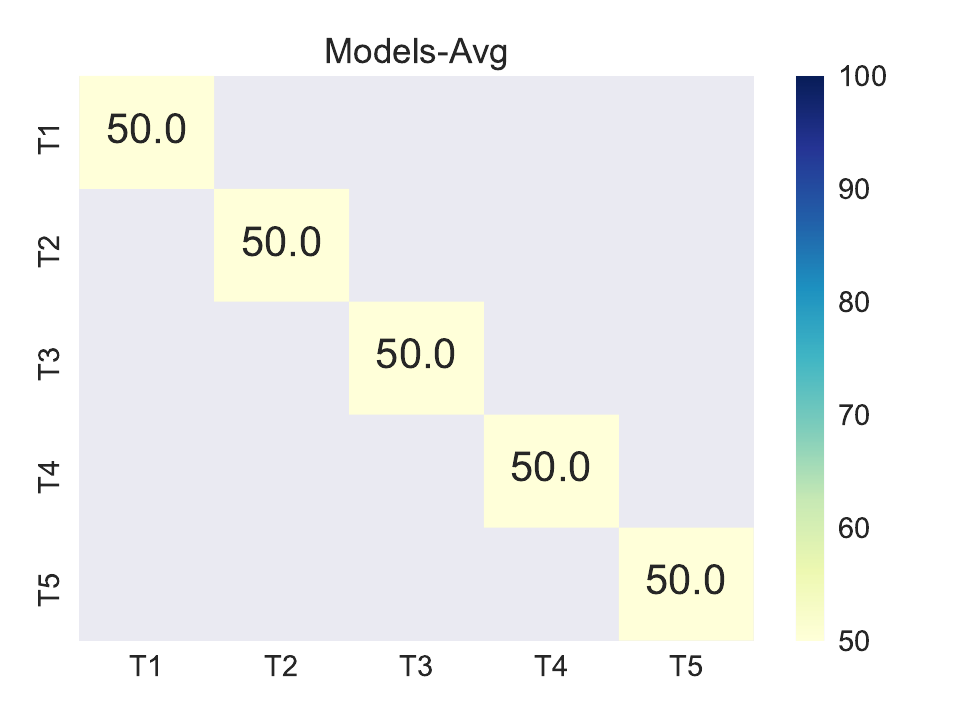}
\includegraphics[width=.33\textwidth]{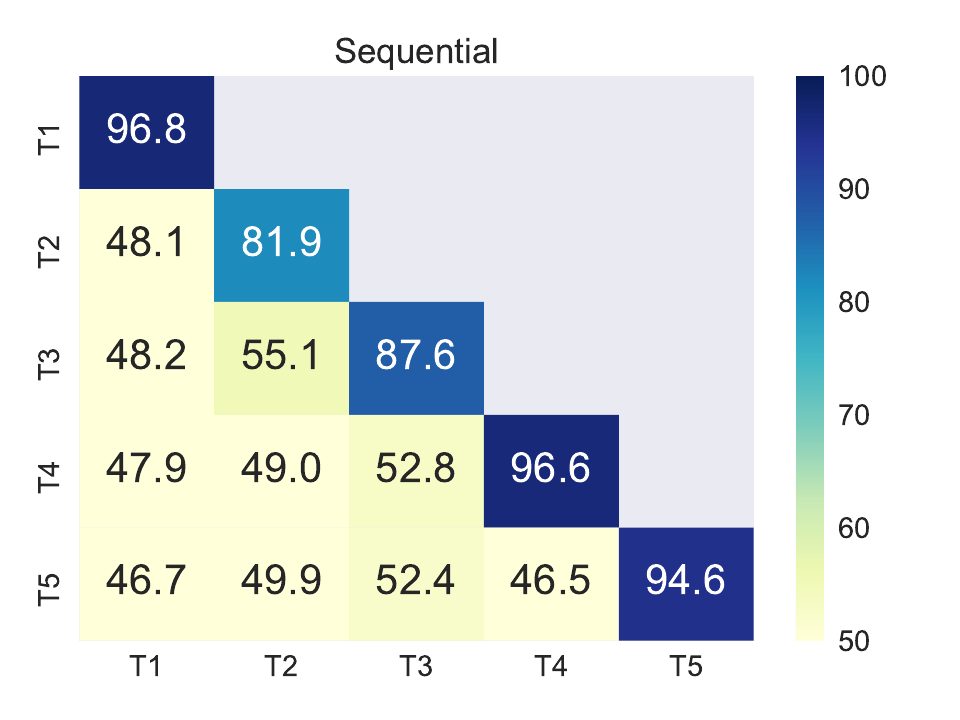}
\includegraphics[width=.33\textwidth]{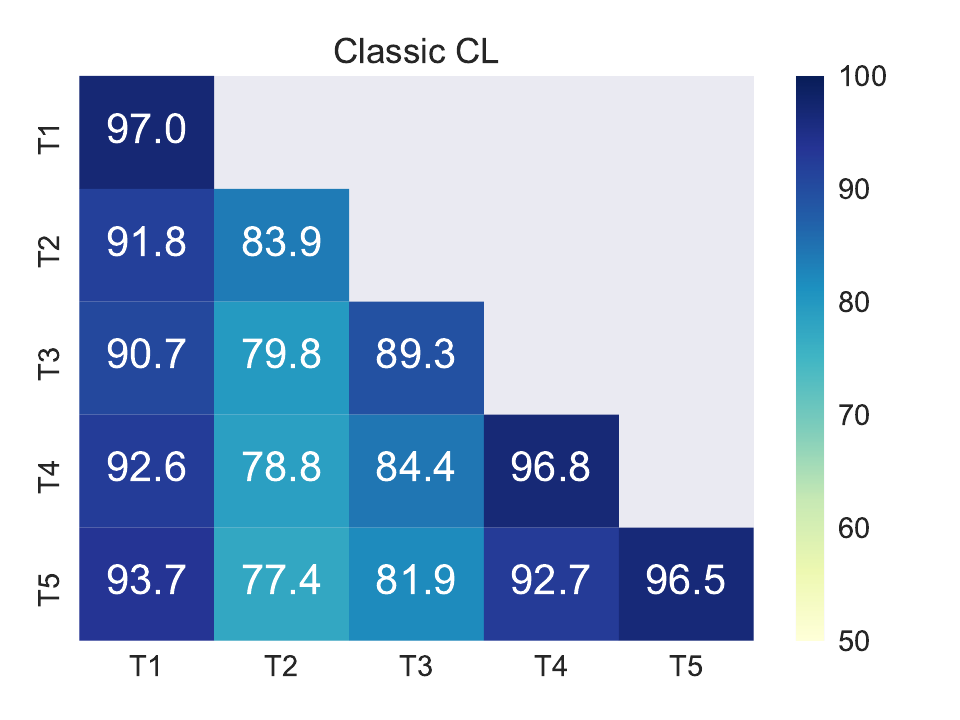}
\includegraphics[width=.33\textwidth]{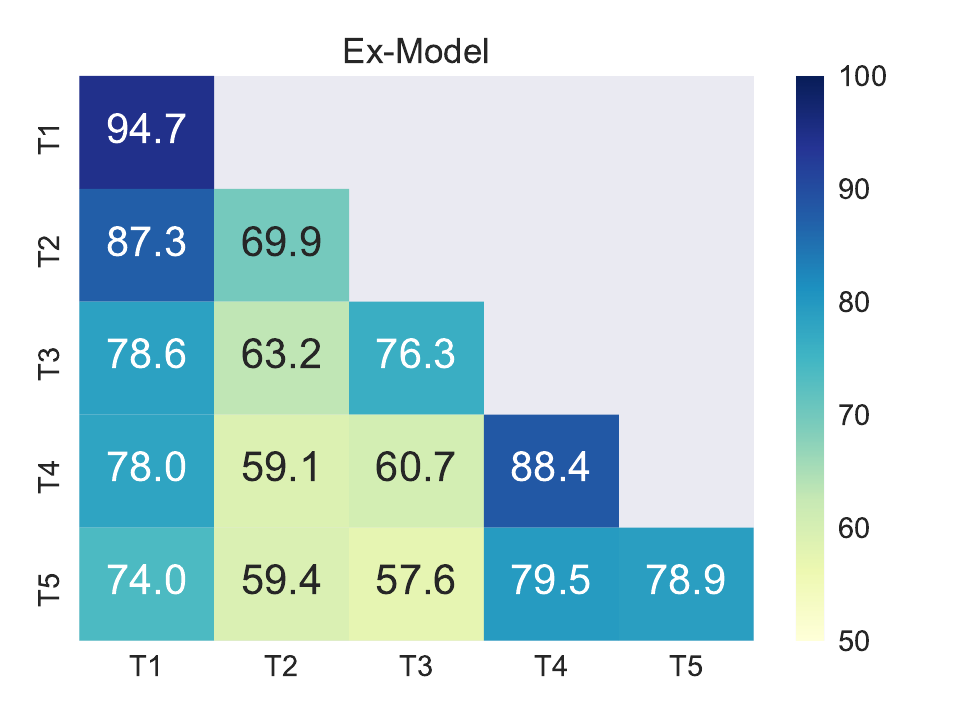}
\includegraphics[width=.33\textwidth]{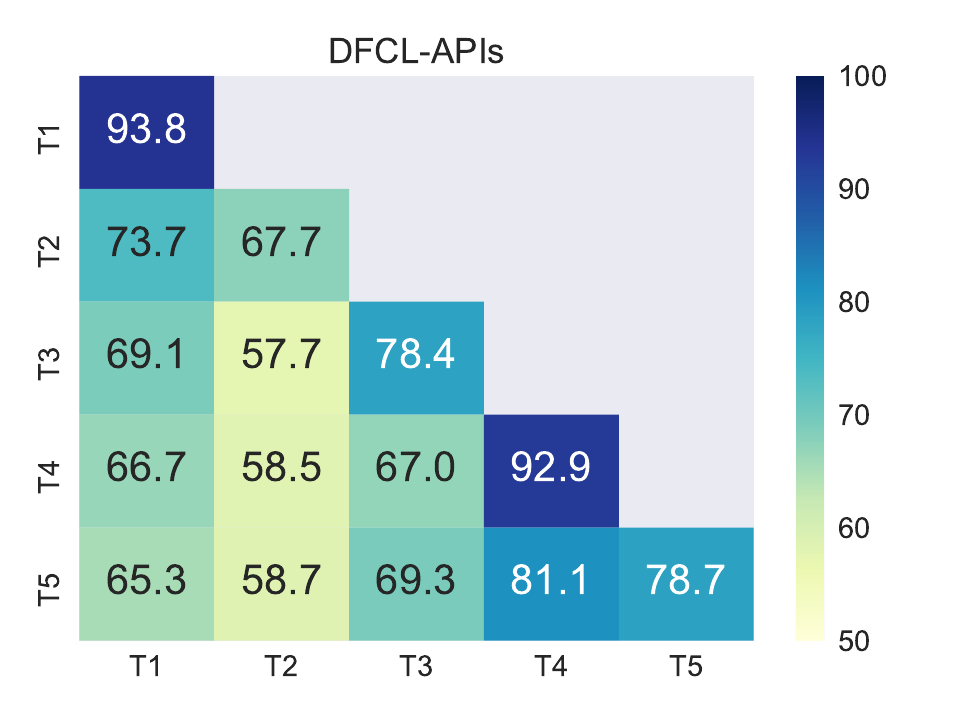}
\includegraphics[width=.33\textwidth]{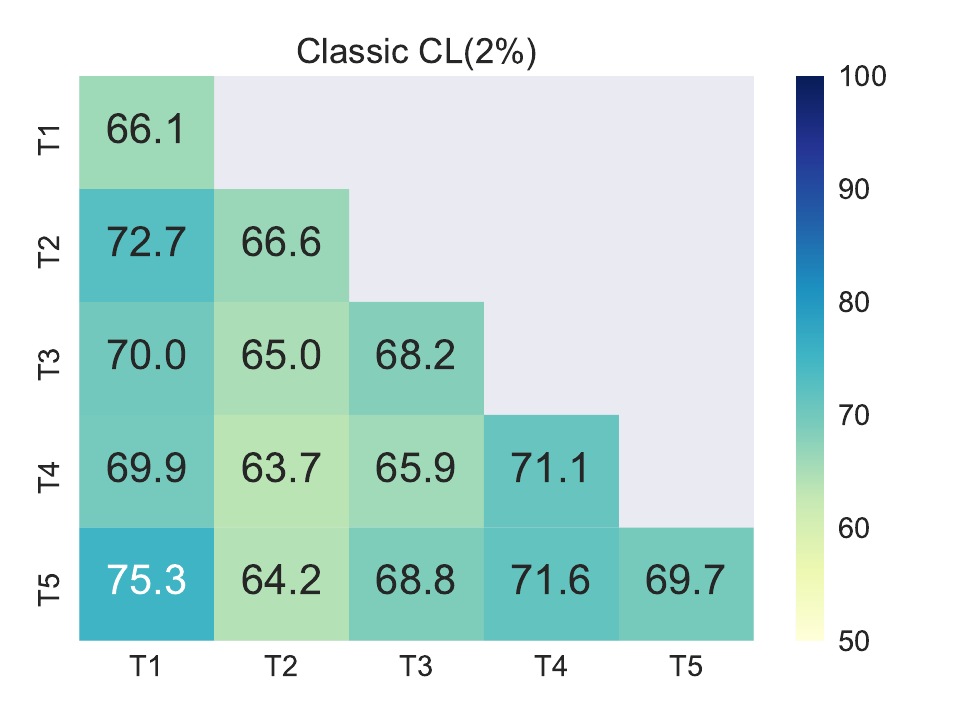}
\includegraphics[width=.33\textwidth]{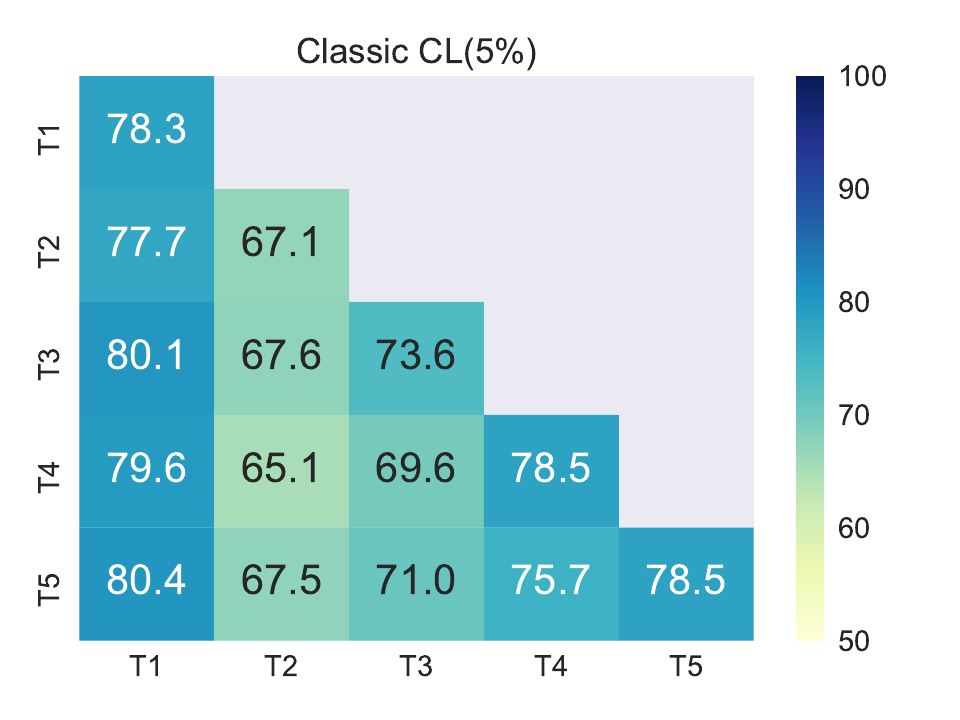}
\includegraphics[width=.33\textwidth]{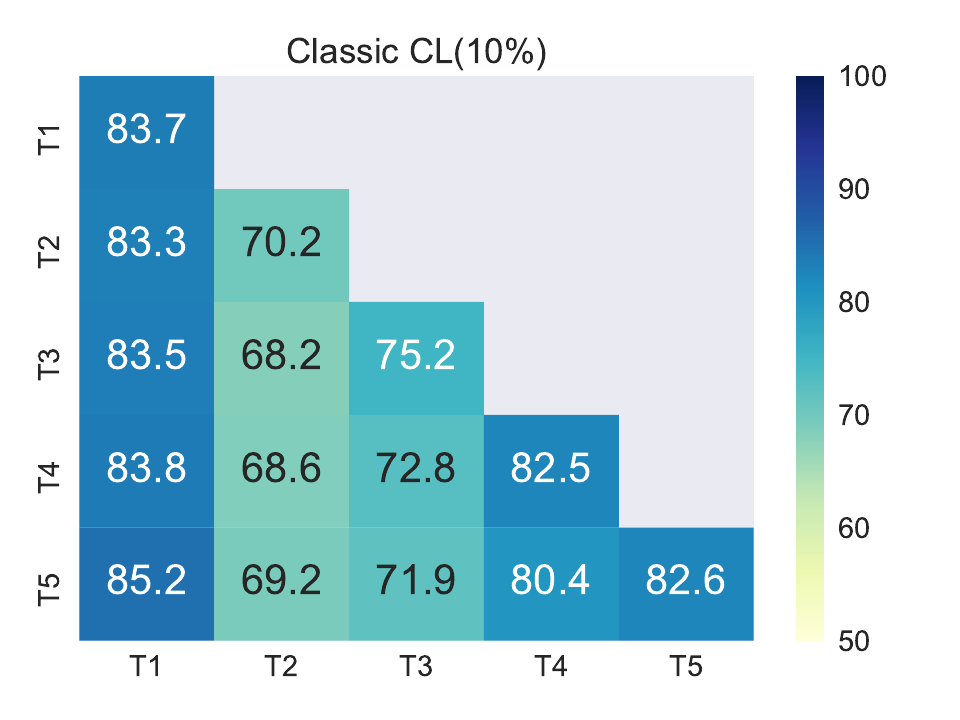}
\includegraphics[width=.33\textwidth]{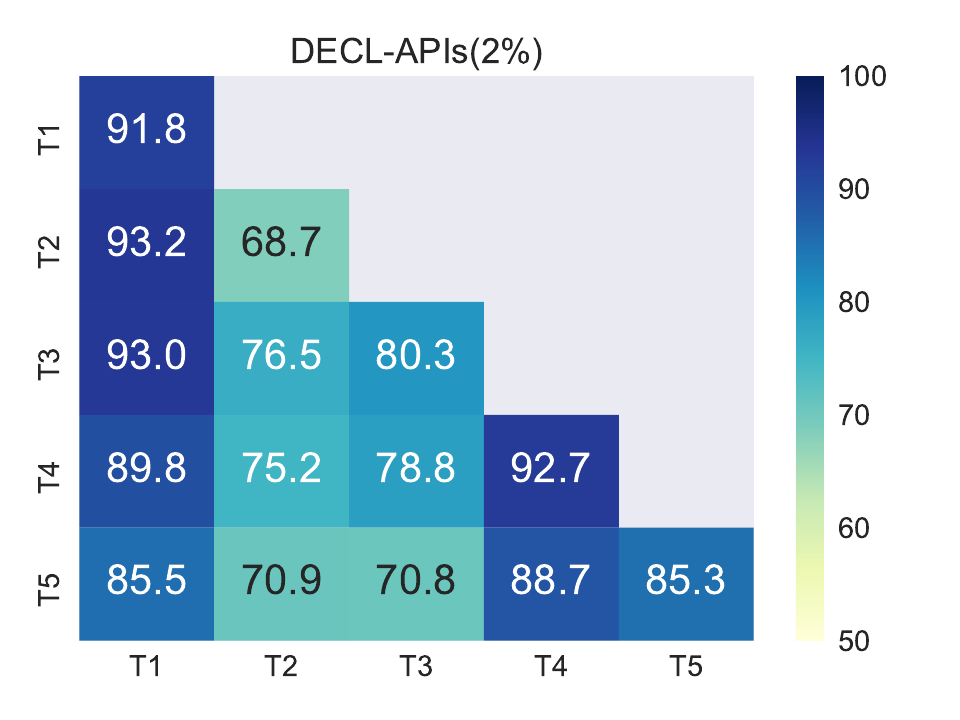}
\includegraphics[width=.33\textwidth]{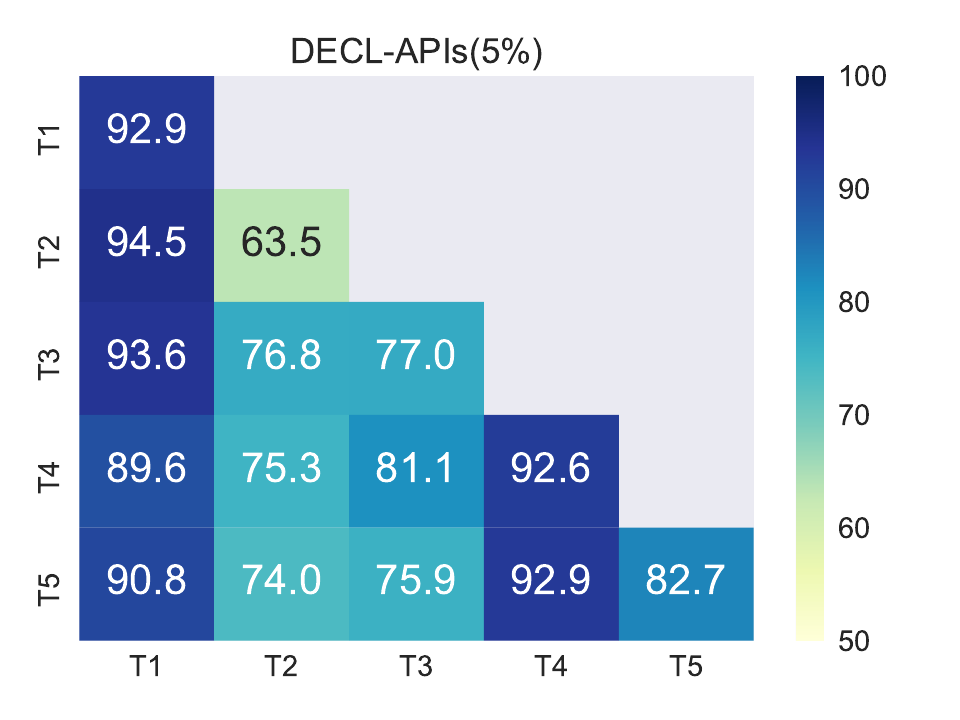}
\includegraphics[width=.33\textwidth]{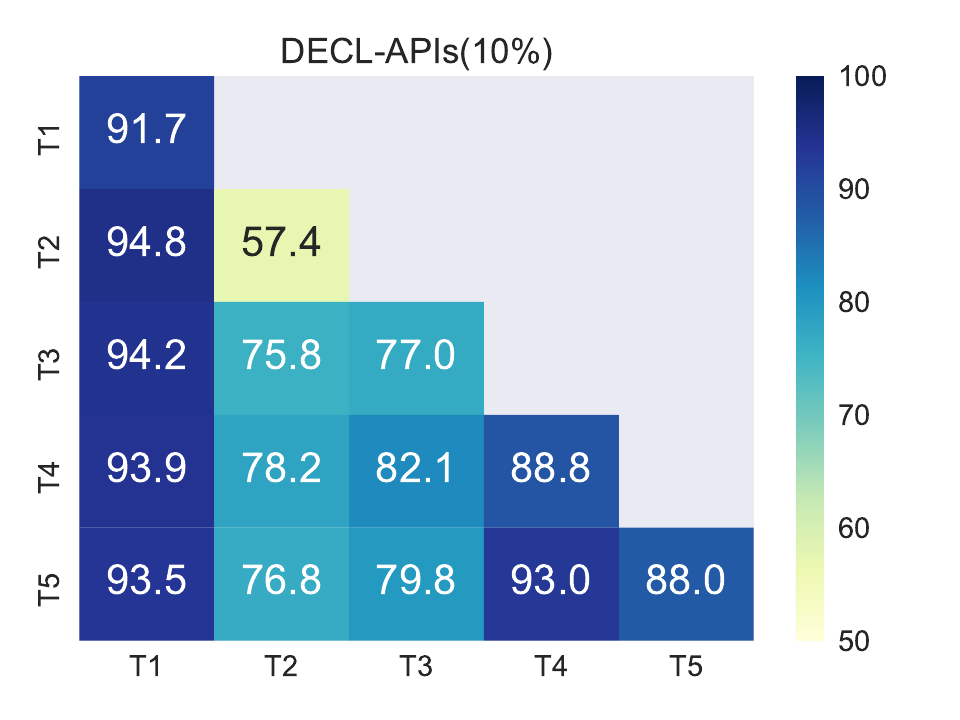}
\caption{The accuracy (Higher Better) on the \textbf{\revised{Split-}CIFAR10} dataset. (1) Joint, (2) Models-Avg, (3) Sequential, (4) Classic CL, (5) Ex-Model, (6) DFCL-APIs(ours), (7) Classic CL-2\%, (8) Classic CL-5\%, (9) Classic CL-10\%, (10) DECL-APIs-2\%(ours), (11) DECL-APIs-5\%(ours), (12) DECL-APIs-10\%(ours). $t$-th row represents the accuracy of the network tested on tasks $1-t$ after task $t$ is learned.
}
\label{fig:acc_cifar10}
\end{figure*}

\begin{figure*}[h]
\centering
\includegraphics[width=.33\textwidth]{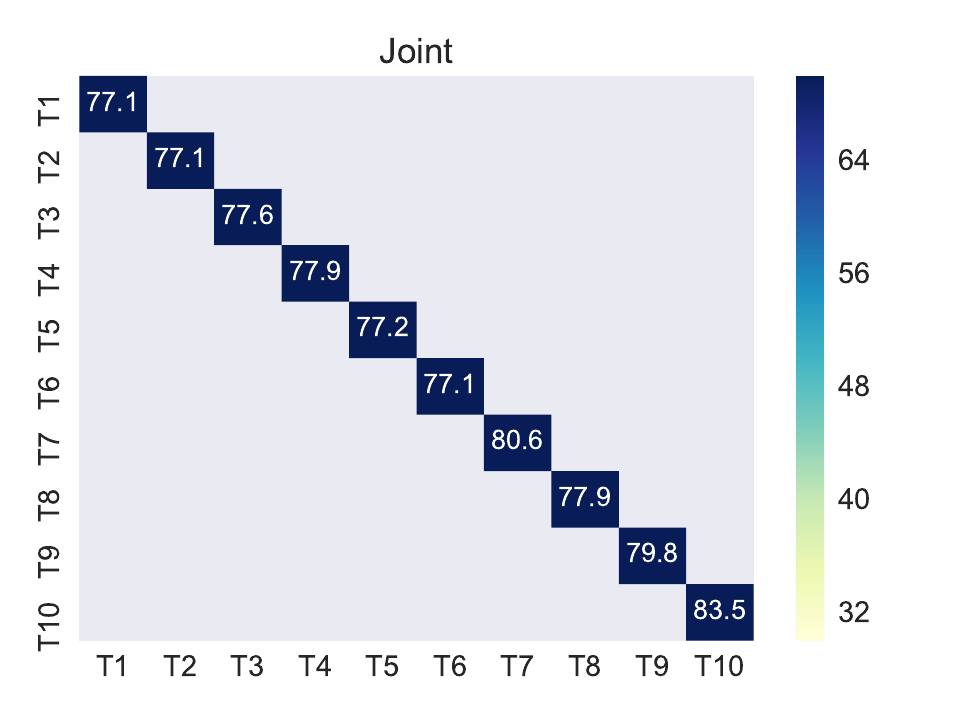}
\includegraphics[width=.33\textwidth]{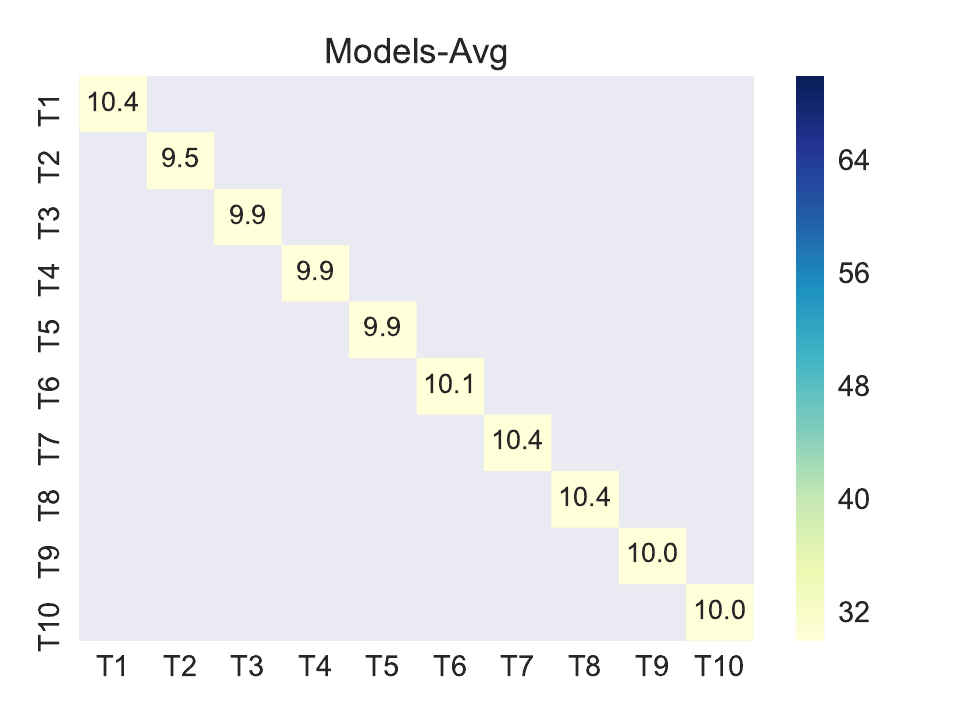}
\includegraphics[width=.33\textwidth]{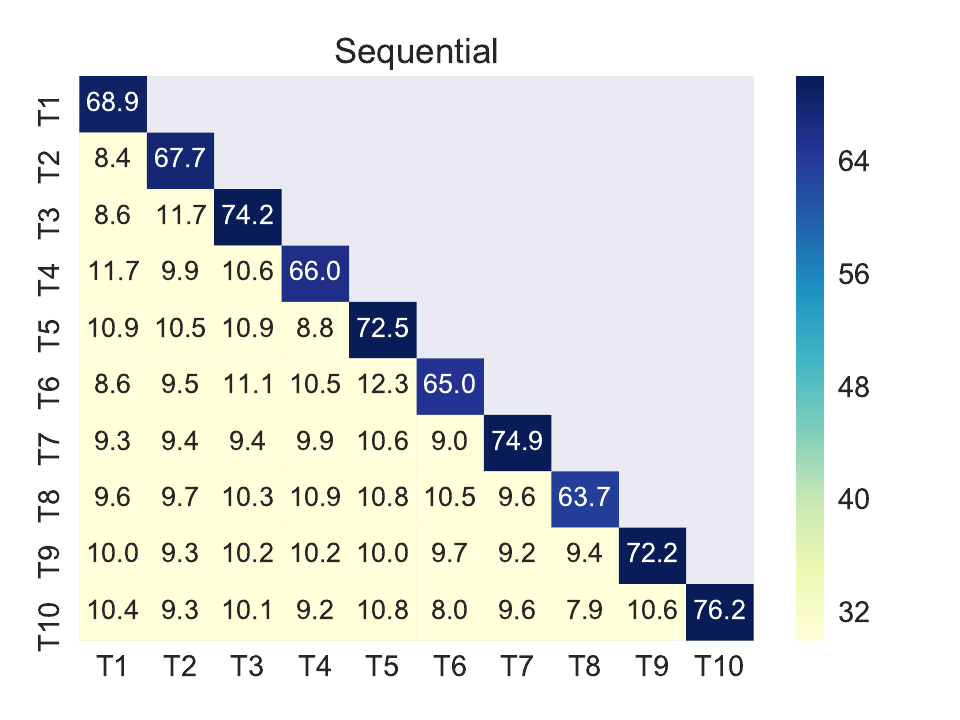}
\includegraphics[width=.33\textwidth]{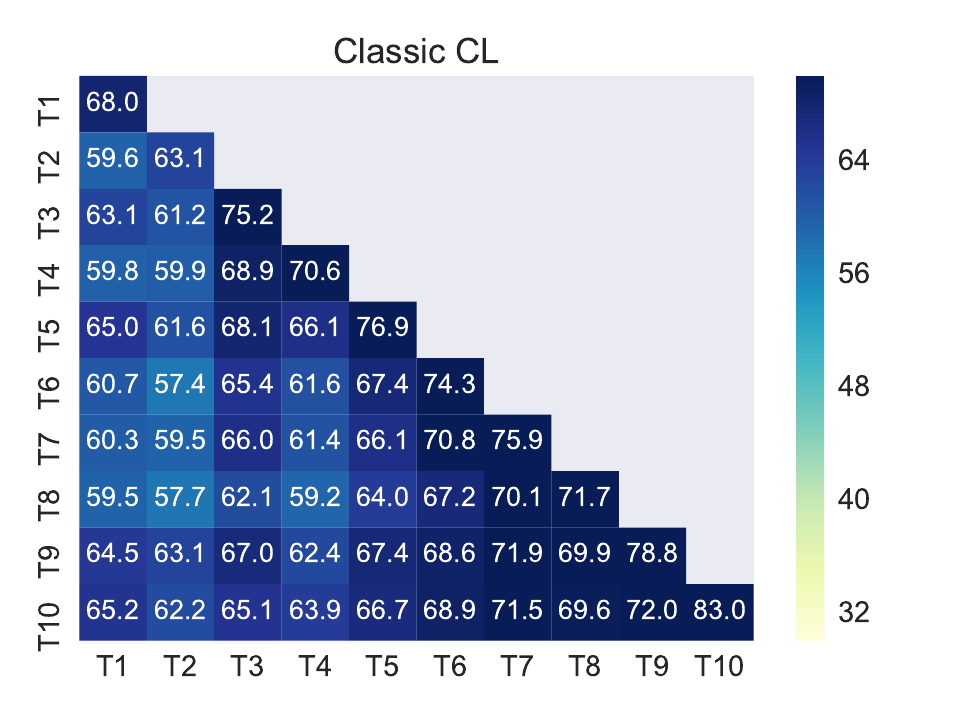}
\includegraphics[width=.33\textwidth]{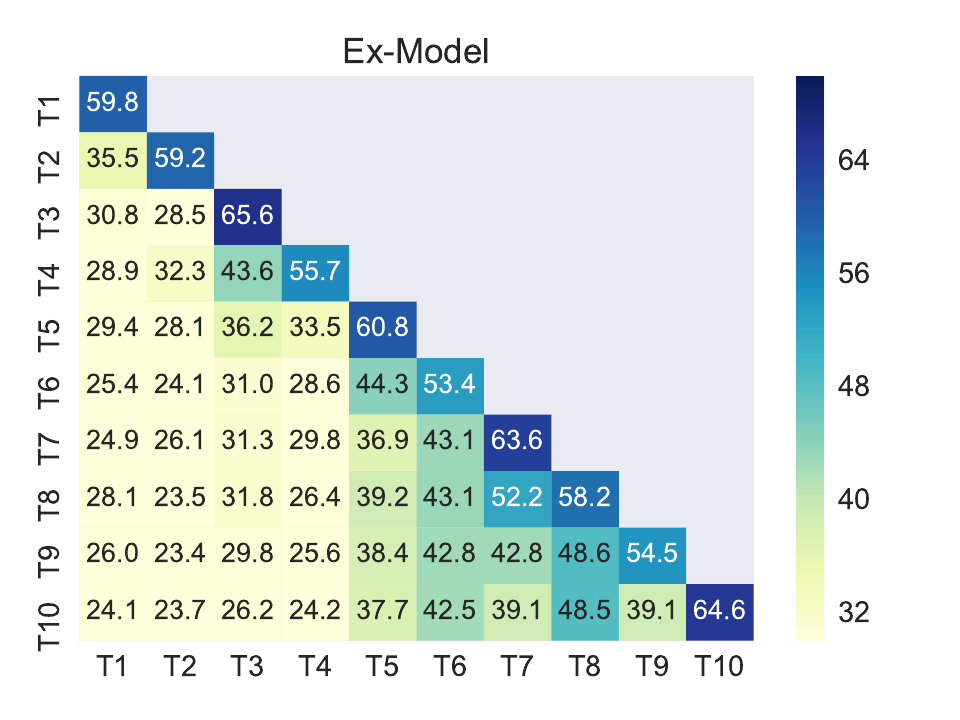}
\includegraphics[width=.33\textwidth]{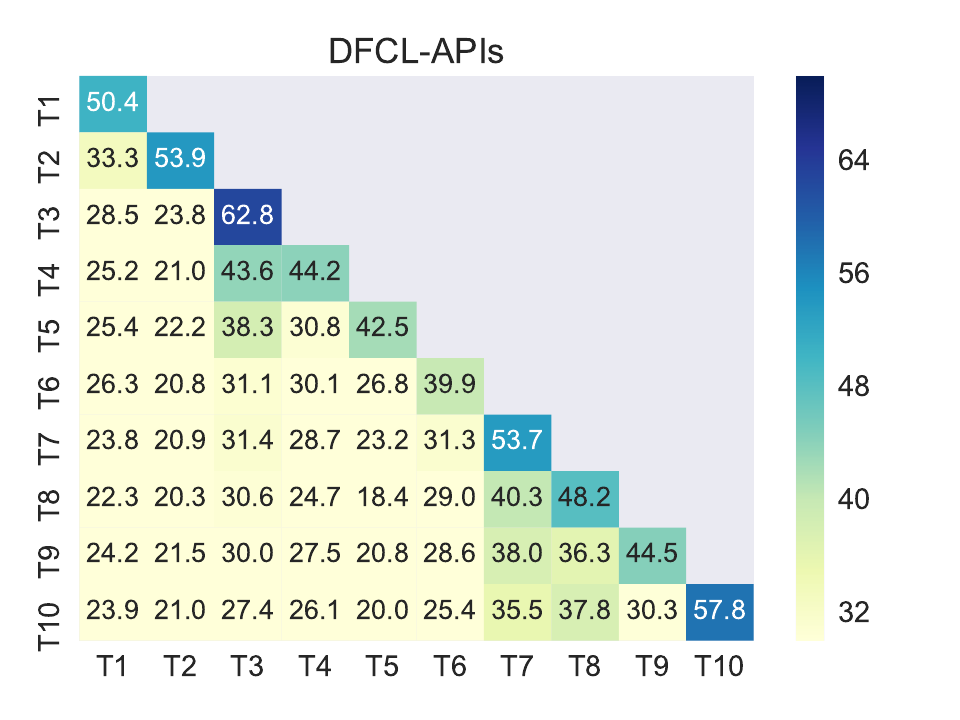}
\includegraphics[width=.33\textwidth]{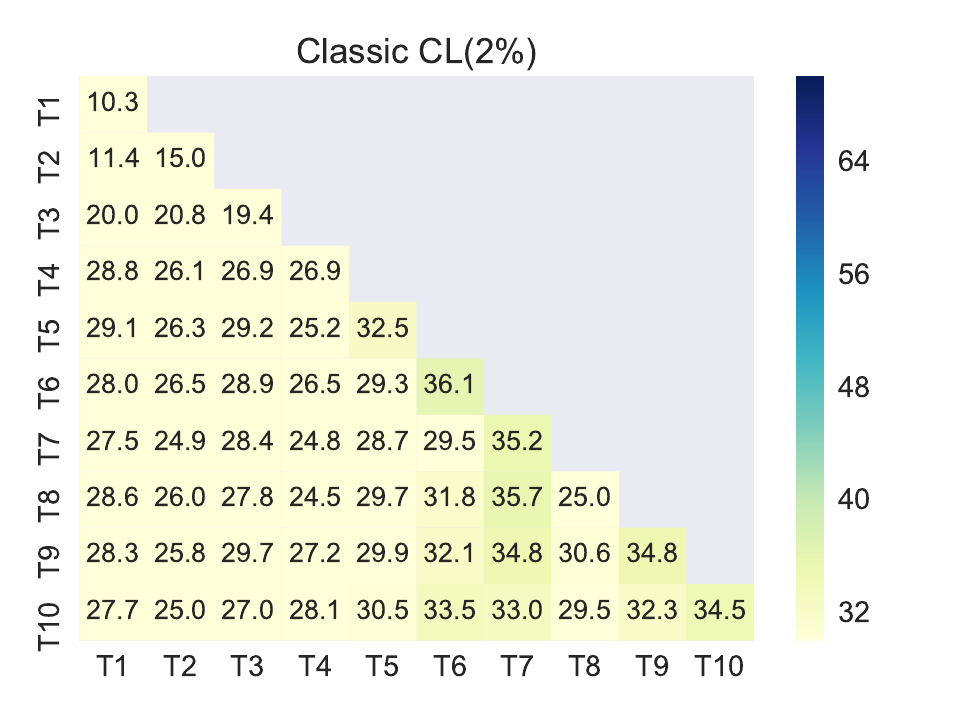}
\includegraphics[width=.33\textwidth]{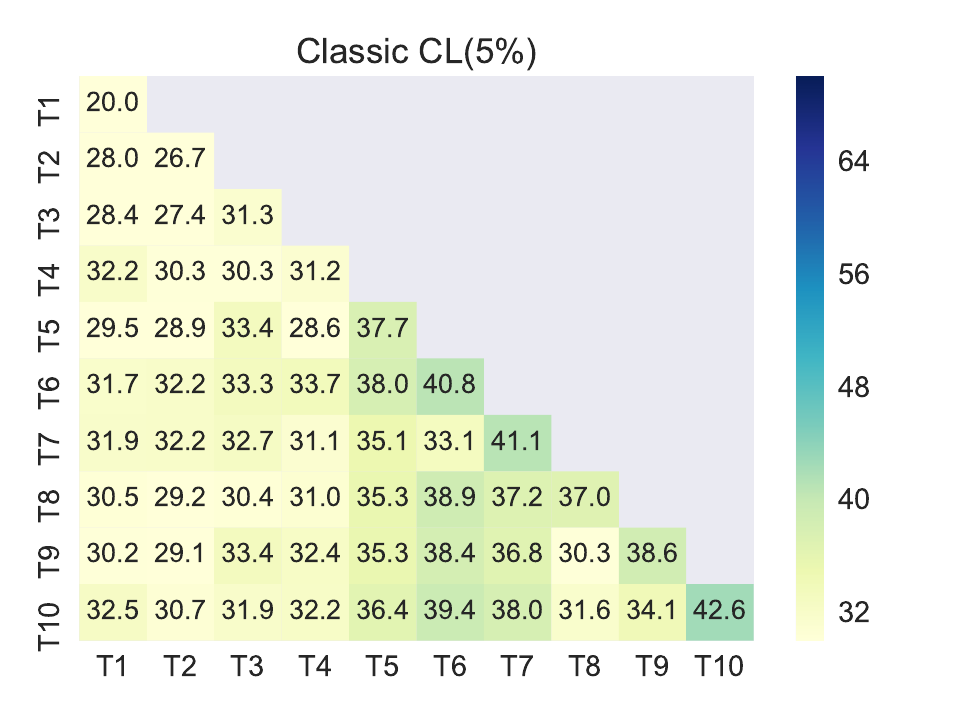}
\includegraphics[width=.33\textwidth]{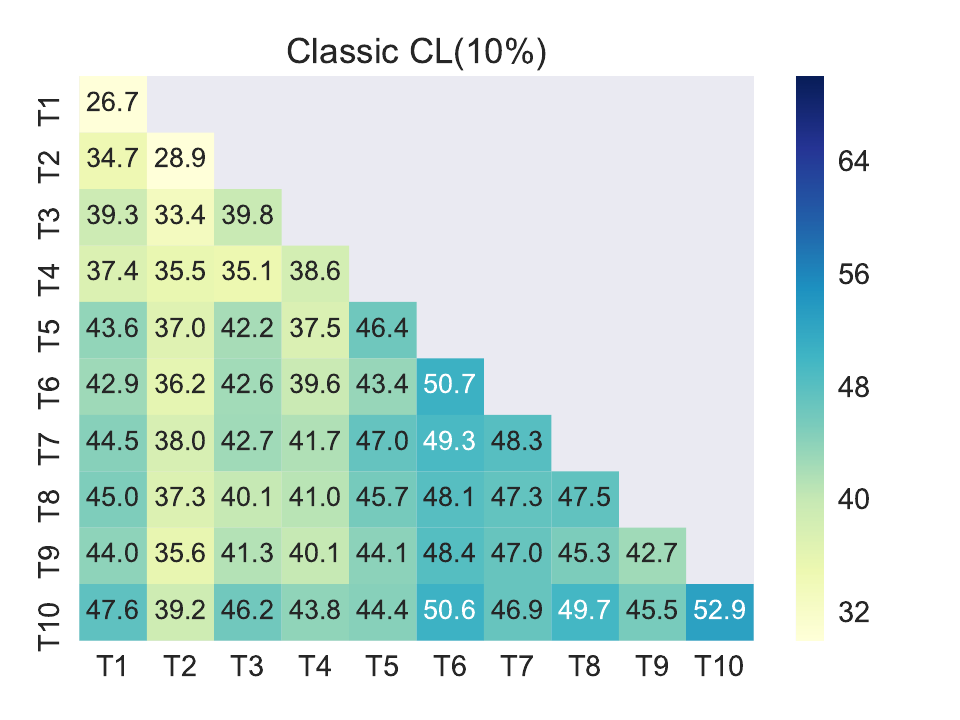}
\includegraphics[width=.33\textwidth]{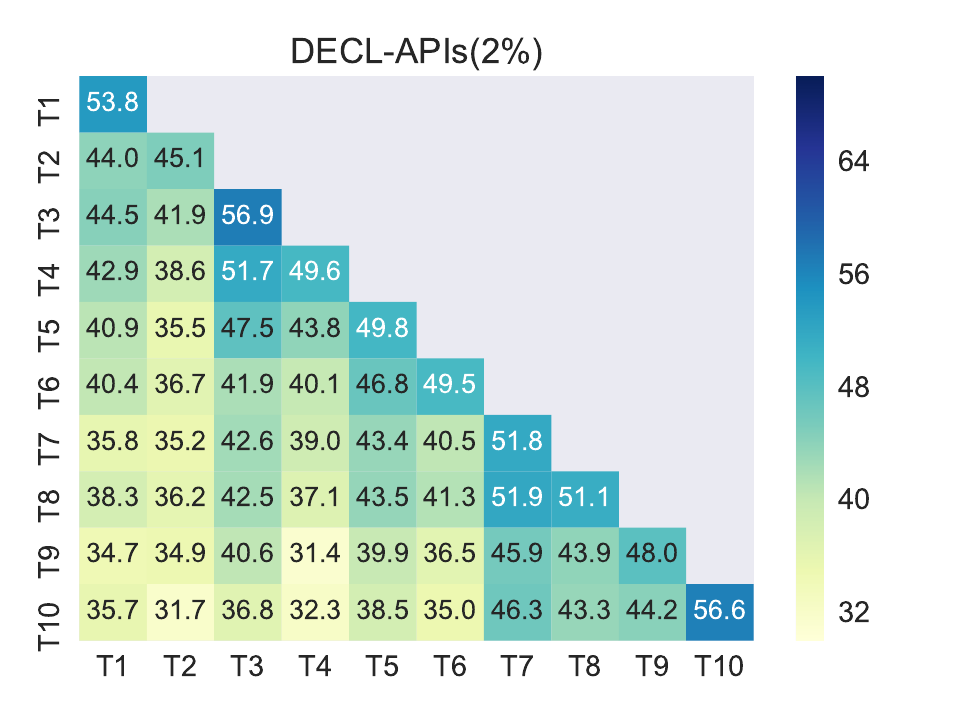}
\includegraphics[width=.33\textwidth]{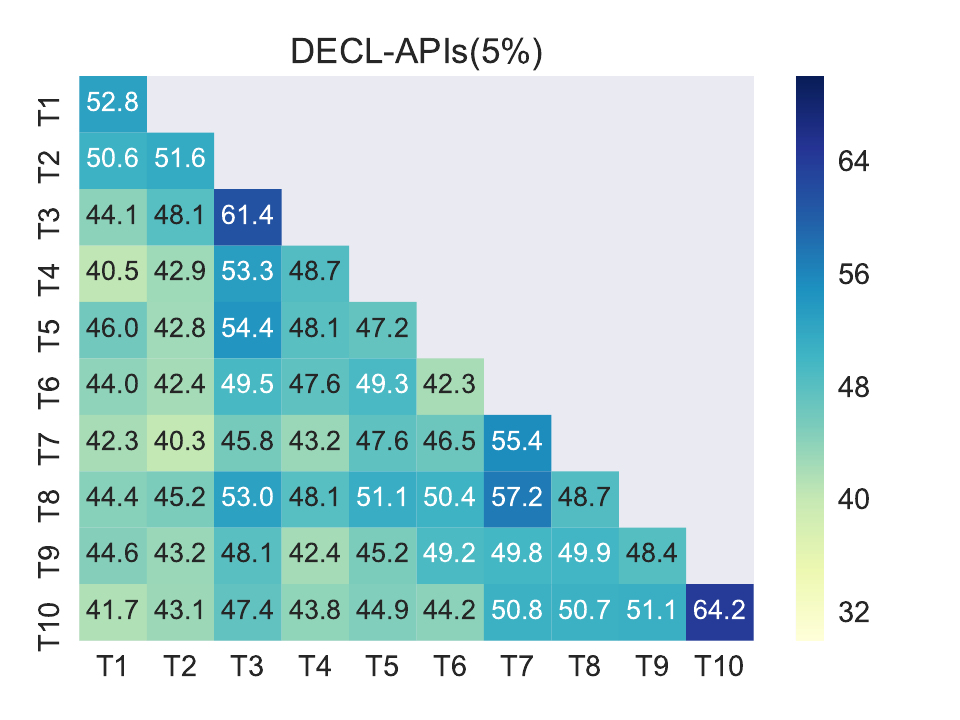}
\includegraphics[width=.33\textwidth]{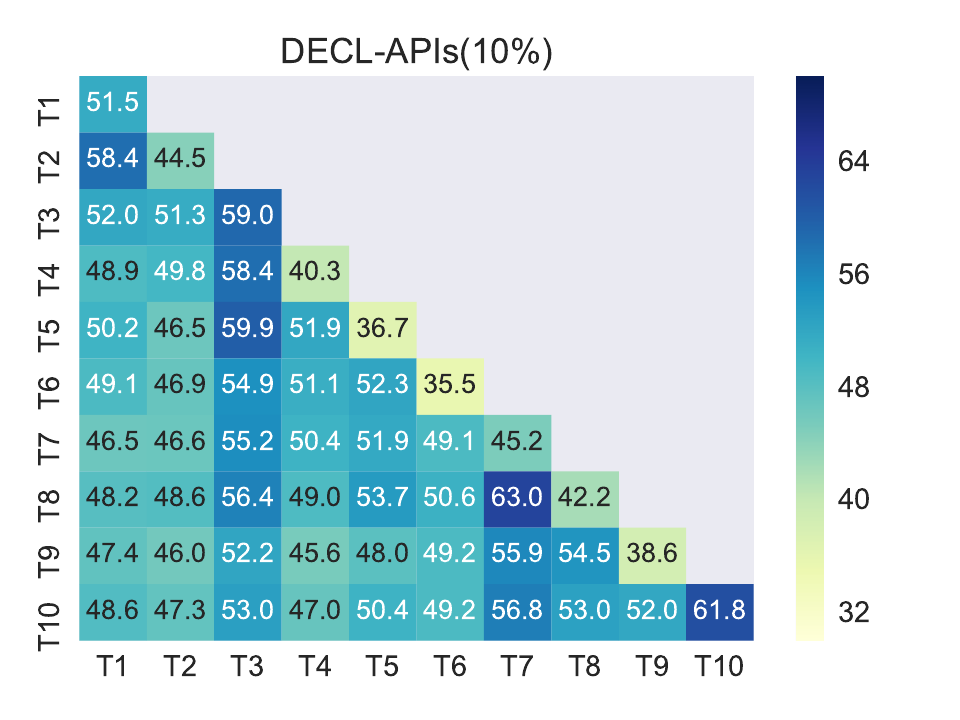}
\caption{The accuracy (Higher Better) on the \textbf{\revised{Split-}CIFAR100} dataset. (1) Joint, (2) Models-Avg, (3) Sequential, (4) Classic CL, (5) Ex-Model, (6) DFCL-APIs(ours), (7) Classic CL-2\%, (8) Classic CL-5\%, (9) Classic CL-10\%, (10) DECL-APIs-2\%(ours), (11) DECL-APIs-5\%(ours), (12) DECL-APIs-10\%(ours). $t$-th row represents the accuracy of the network tested on tasks $1-t$ after task $t$ is learned.
}
\label{fig:acc_cifar100}
\end{figure*}

\begin{figure*}[h]
\centering
\includegraphics[width=.33\textwidth]{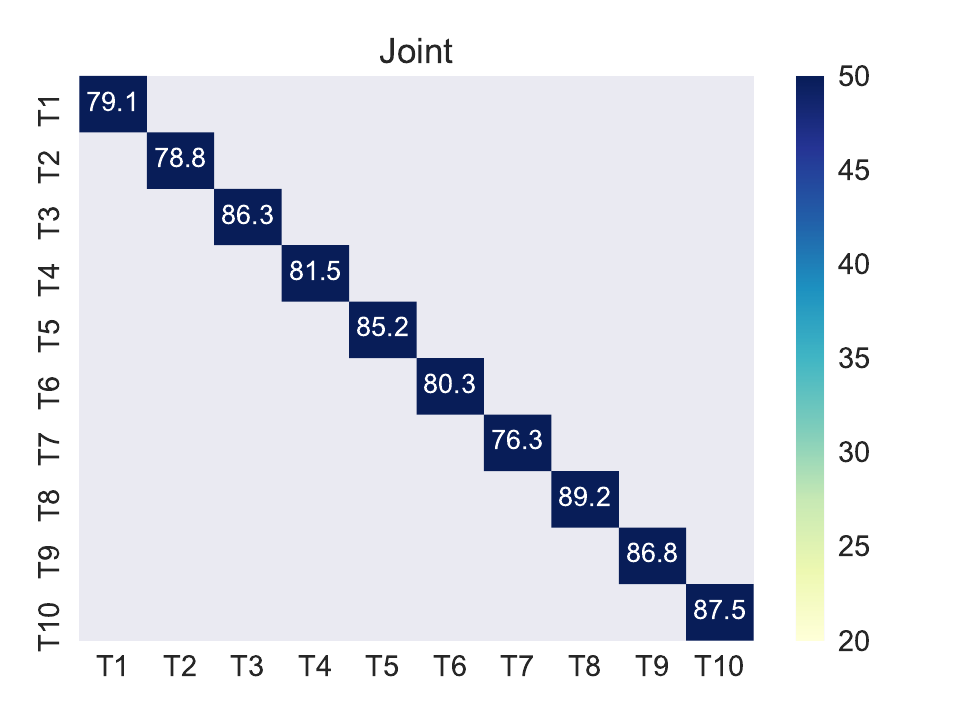}
\includegraphics[width=.33\textwidth]{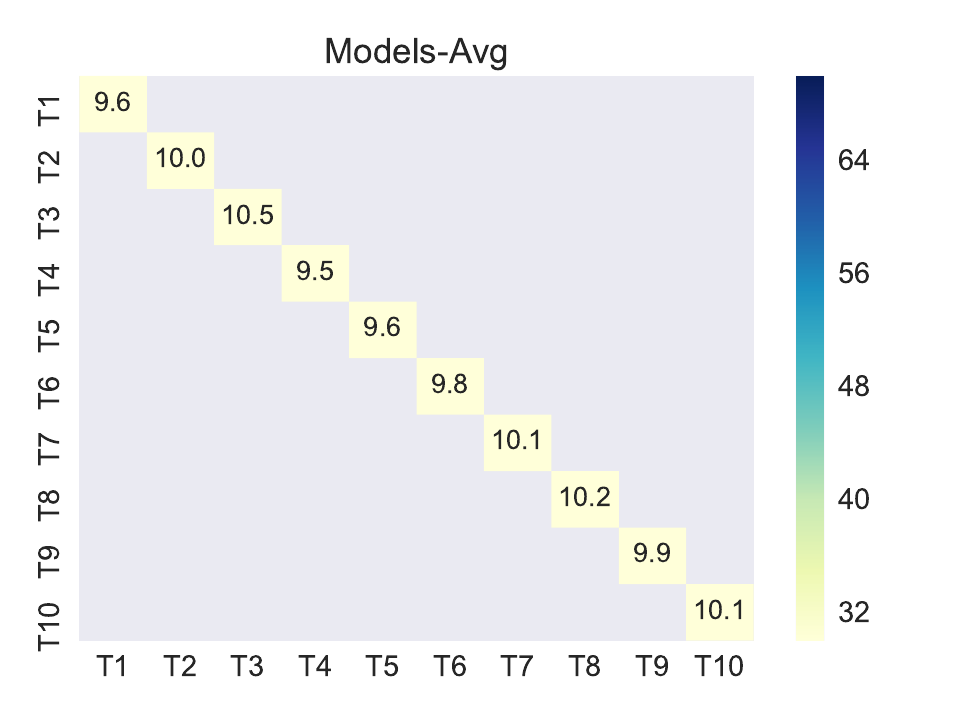}
\includegraphics[width=.33\textwidth]{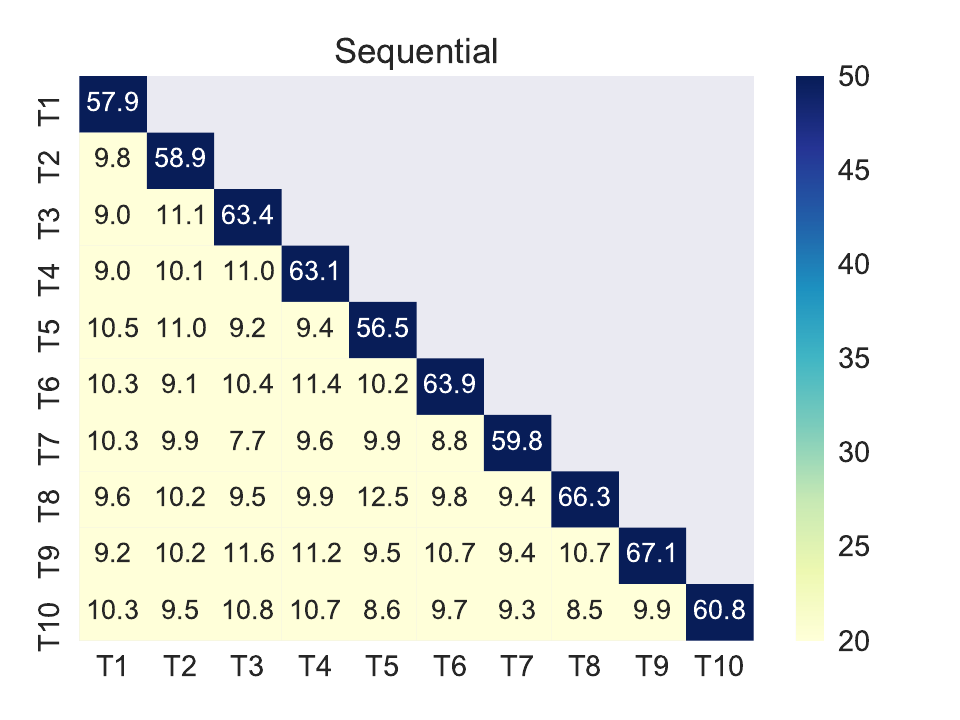}
\includegraphics[width=.33\textwidth]{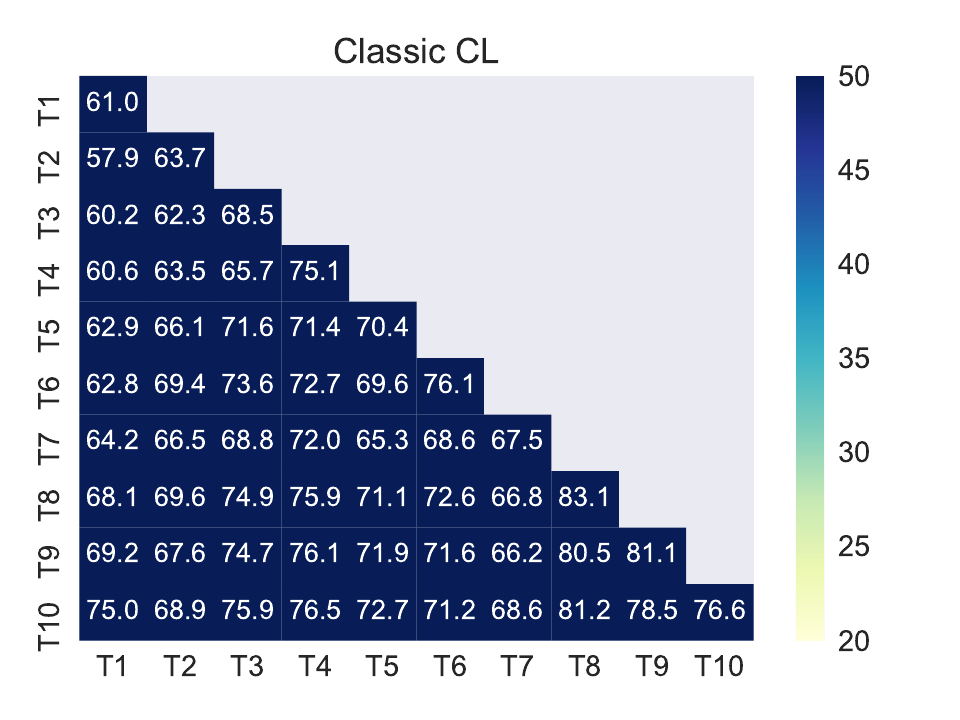}
\includegraphics[width=.33\textwidth]{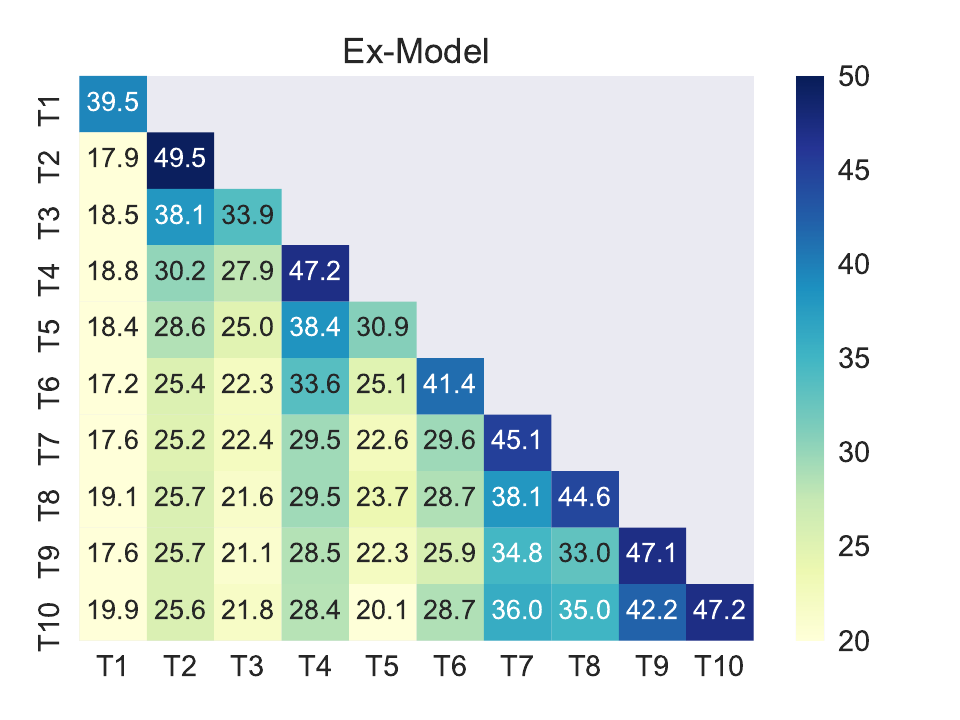}
\includegraphics[width=.33\textwidth]{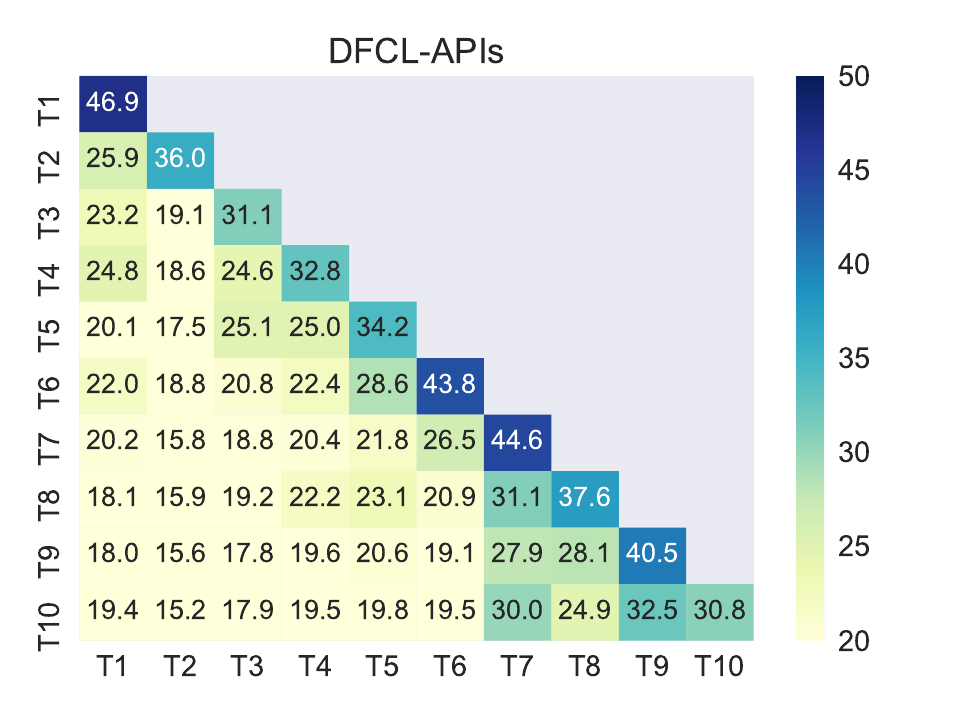}
\includegraphics[width=.33\textwidth]{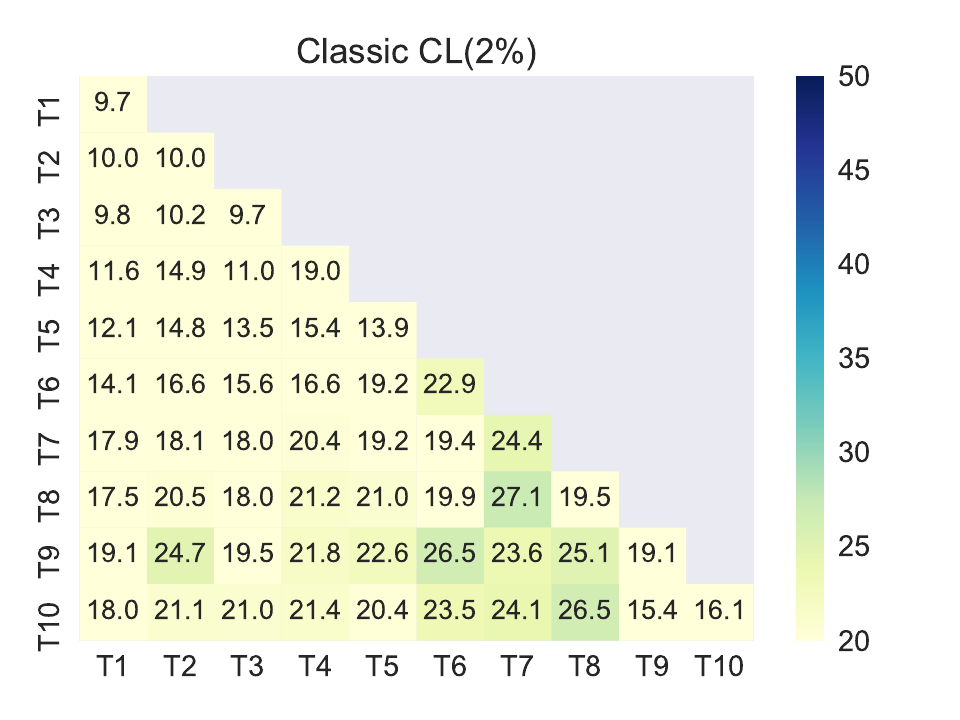}
\includegraphics[width=.33\textwidth]{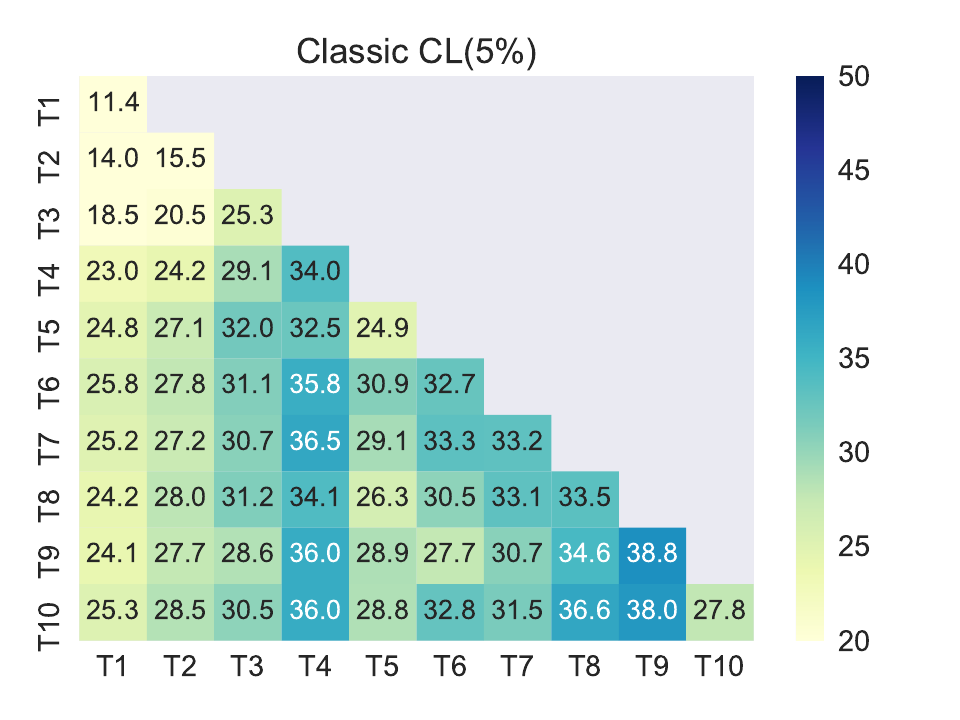}
\includegraphics[width=.33\textwidth]{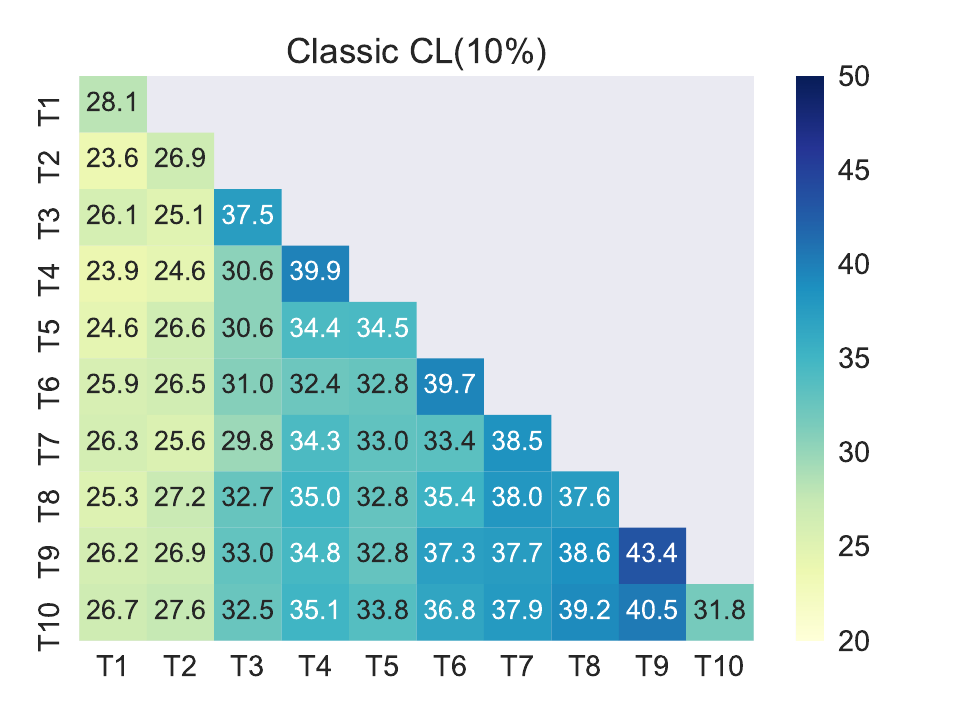}
\includegraphics[width=.33\textwidth]{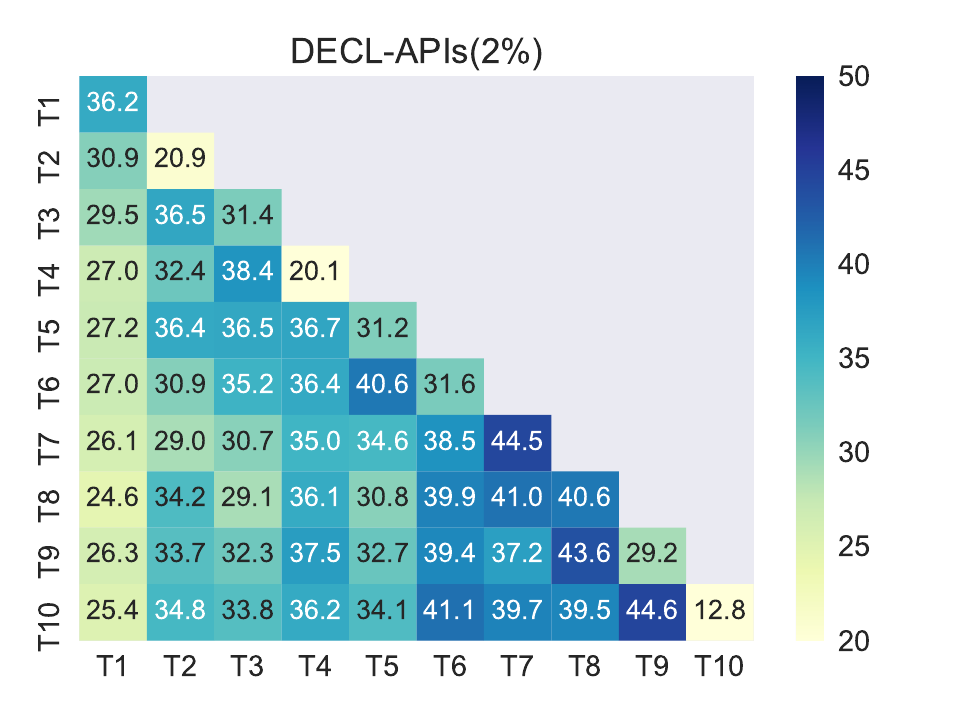}
\includegraphics[width=.33\textwidth]{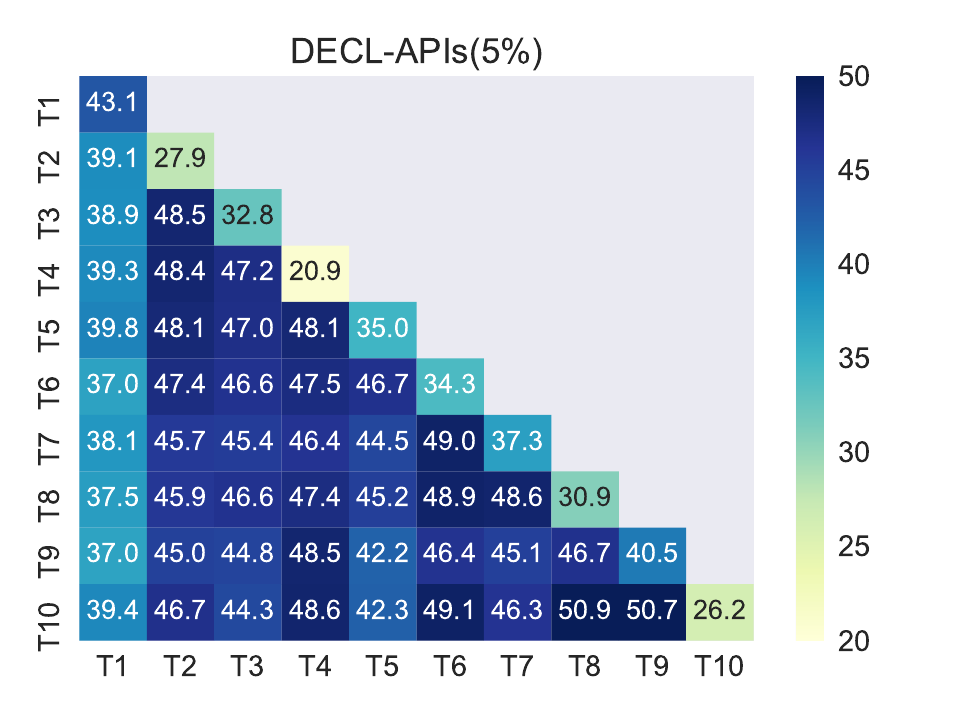}
\includegraphics[width=.33\textwidth]{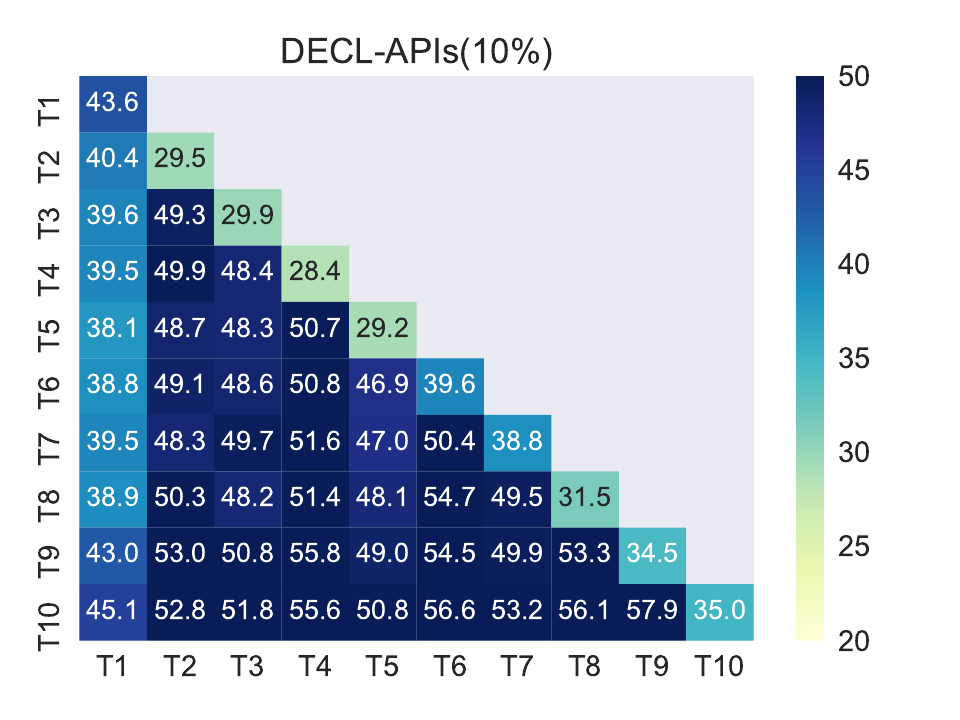}
\caption{The accuracy (Higher Better) on the \textbf{\revised{Split-}MiniImageNet} dataset. (1) Joint, (2) Models-Avg, (3) Sequential, (4) Classic CL, (5) Ex-Model, (6) DFCL-APIs(ours), (7) Classic CL-2\%, (8) Classic CL-5\%, (9) Classic CL-10\%, (10) DECL-APIs-2\%(ours), (11) DECL-APIs-5\%(ours), (12) DECL-APIs-10\%(ours). $t$-th row represents the accuracy of the network tested on tasks $1-t$ after task $t$ is learned.
}
\label{fig:acc_miniimagenet}
\end{figure*}

We provide the variation of the accuracy of each stage on the five datasets for all compared methods in Figures \ref{fig:acc_mnist}, \ref{fig:acc_svhn}, \ref{fig:acc_cifar10}, \ref{fig:acc_cifar100}, \ref{fig:acc_miniimagenet}. We have the following conclusions: (i) \textit{Joint} and \textit{Models-Avg} are not a sequential learning process, so there is no catastrophic forgetting problem. (ii) Naive \textit{Sequential} has a severe forgetting problem because there is no strategy to avoid forgetting. (iii) The \textit{Classic CL} obviously alleviates catastrophic forgetting due to the experience replay mechanism. (iv) Both \textit{Ex-Model} and \textit{DFCL-APIs} perform slightly worse than Classic CL due to lack of original training data. In addition, since Ex-Model is a white box, it is the upper bound of DFCL-APIs. (v) In the setting of \textit{Classic CL} and \textit{DECL-APIs} with small amounts of data (i.e., $\small 2\%$, $\small 5\%$ and $\small 10\%$), the performance increases continuously with the amount of available data. In addition, DECL-APIs is superior to Classic CL 
because of the additional use of generated data.

\end{document}